\documentclass{article}

\date{}

\usepackage{graphicx,float,wrapfig}
\usepackage{comment}
\usepackage{subfigure}
\usepackage{amsmath,amsfonts,amsthm,amssymb}
\newtheorem{theorem}{Theorem}
\newtheorem{lemma}{Lemma}

\newtheorem{corollary}{Corollary}

\newtheorem{assumption}{Assumption}

\def\hI{\widehat I}

\def\hH{\widehat H}
\def\Pr{\mathbb P}
\def\reals{\mathbb R}
\def\E{\mathbb E}
\def\V{{\rm Var}}
\def\Cov{{\rm Cov}}

\def\hf{\widehat{f}}
\def\cX{{\cal X}}
\def\cY{{\cal Y}}
\def\Ca{C_a}
\def\Cb{C_b}
\def\Cc{C_c}
\def\Cd{C_d}
\def\C0{C_0}
\def\tO{\widetilde{O}}

\title{Demystifying Fixed  $k$-Nearest Neighbor Information Estimators}

\author{ Weihao Gao\thanks{Department of Electrical and Computer  Engineering, Coordinated Science Laboratory, University of Illinois at Urbana-Champaign, email: \texttt{wgao9@illinois.edu}},  \;\;\; Sewoong Oh\thanks{Department of Industrial and Enterprise Systems Engineering, Coordinated Science Laboratory, University of Illinois at Urbana-Champaign, email: \texttt{swoh@illinois.edu}}, \;\;\; Pramod Viswanath\thanks{Department of Electrical and Computer  Engineering, Coordinated Science Laboratory, University of Illinois at Urbana-Champaign, email: \texttt{pramodv@illinois.edu}}
}

\begin{document}
% \nipsfinalcopy is no longer used

\maketitle

\begin{abstract}
  Estimating mutual information from i.i.d.\ samples drawn from an unknown joint density function is a basic statistical problem of broad interest with multitudinous applications. The most popular  estimator is one proposed by Kraskov and St{\"o}gbauer and Grassberger (KSG) in 2004, and is  nonparametric and based on the distances of each sample to its $k^{\rm th}$ nearest neighboring sample, where $k$ is a fixed small integer. Despite its widespread use (part of scientific software packages),  theoretical properties of this estimator have been largely unexplored. In this paper we demonstrate that the estimator is consistent  and also identify an upper bound on the rate of convergence of the $\ell_2$ error as a function of number of samples. We argue that the  performance benefits of the KSG estimator stems from a curious ``correlation boosting" effect and build on this intuition to modify the KSG estimator in novel ways to construct a superior estimator. As a byproduct of our investigations, we obtain nearly tight rates of convergence of the $\ell_2$ error of the well known fixed $k$ nearest neighbor estimator of differential entropy by Kozachenko and Leonenko.
\end{abstract}

\section{Introduction}

Information theoretic quantities such as  mutual information measure {\em relations} between random variables. A key property of these measures is that they are invariant to one-to-one transformations of the random variables and obey the data processing inequality \cite{cover1991information,jiao2015justification}. These properties combine to make information theoretic quantities attractive in several data science applications involving clustering \cite{muller2012information,ver2014maximally,chan2015multivariate}, classification \cite{peng2005feature} and more generally as a basic feature that can be used in  several downstream applications \cite{fleuret2004fast,battiti1994using,wells1996multi,turney2002thumbs}.
A canonical question in all these applications is to estimate the information theoretic quantities from {\em samples}, typically supposed to be drawn i.i.d.\ from an unknown distribution. This fundamental question has been of longstanding interest in the theoretical statistics community where it is a canonical question of estimating a functional of the (unknown) density  \cite{birge1995estimation}
 but also in the information theory \cite{wang2009divergence,paninski2004estimating,wu2014minimax,wang2005divergence}, machine learning \cite{gao2014efficient,kinney2014equitability} and theoretical computer science \cite{valiant2011estimating,batu2000testing,acharya2014estimating} communities, with significant renewed interest of late, summarized in detail in Section \ref{sec:priorwork}.
The most fundamental information theoretic quantity of interest is the mutual information between a pair of random variables, which is also the primary focus of this paper, in the context of real valued random variables (in potentially high dimensions).

The basic estimation question takes a different hue depending on whether the underlying distribution is discrete or continuous. In the discrete setting,  significant understanding of the minimax rate-optimal estimation of functionals, including entropy and mutual information, of an unknown probability mass function is attained via recent works \cite{paninski2004estimating,paninski2003estimation,valiant2011estimating,jiao2015minimax,wu2014minimax}.
 The continuous setting is significantly different, bringing to fore the interplay of geometry of the Euclidean space as well as the role of dimensionality of the domain in terms of estimating the information theoretic quantities; this setting is the focus of this paper.
Among the various estimation methods, of great theoretical interest and high practical relevance, are the {\em nearest neighbor} (NN) methods: the quantities of interest are estimated  based on distances (in an appropriate norm) of the samples to their $k$-nearest neighbors ($k$-NN). Of particular practical  interest is the situation when $k$ is a small  {\em fixed} integer -- typically in the range of 4$\sim$8 -- and the estimators based on fixed $k$-NN statistics typically perform significantly better than alternative approaches, discussed in detail in  Section \ref{sec:priorwork}, both in simulations and when tested in the wild; this is especially true when the random variables are in high dimensions.

 The exemplar fixed $k$-NN estimator is that of differential entropy from i.i.d.\ samples proposed in 1987 by Kozachenko and Leonenko \cite{KL87} which involved a novel bias correction term, and we refer to as the KL estimator (of differential entropy).
Since the mutual information between two random variables is the sum and difference of three differential entropy terms, any estimator of differential entropy naturally lends itself into an estimator of mutual information, which we christen as the 3KL estimator (of mutual information). In an inspired work in 2004, Kraskov and St{\"o}gbauer and Grassberger \cite{Kra04}, proposed a different fixed $k$-NN estimator of the mutual information, which we name the KSG estimator,  that involved subtle (sample dependent) alterations to the 3KL estimator. The authors of \cite{Kra04,khan2007relative} empirically demonstrated that the KSG estimator consistently improves over the 3KL estimator in a variety of settings. Indeed, the simplicity of the KSG estimator, combined with its superior performance, has made it a very popular estimator of mutual information in practice.

Despite its widespread use, even basic theoretical properties of the KSG estimator are unknown -- it is not even clear if the estimator has vanishing bias (i.e., consistent) as the number of samples grows, much less any understanding of the asymptotic behavior of the bias as a function of the number of samples. As observed elsewhere \cite{gao2015estimating}, characterizing the theoretical properties of the KSG estimator is of first order importance -- this study could  shed light on why the sample-dependent modifications lead to improved performance and perhaps this understanding could lead to the design of even better mutual information estimators. Such are the goals of this paper.

\bigskip\noindent
{\bf Main results.} We make the following contributions. \begin{itemize}
\item Our main result is to show that the KSG estimator is consistent. We also show upper bounds to the rate of convergence of the bias as a function of the dimensions of the two random variables involved: in the special case when the dimensions of the two random variables are equal and no more than one, the rate of convergence of the $\ell_2$ error is $1/\sqrt{N}$, which is the parametric rate of convergence.
\item We argue that the improvement of the KSG estimator over the 3KL estimator comes from a ``correlation boosting" effect, which can be further amplified by a suitable modification to the KSG estimator. This leads to a novel mutual information estimator, which we call the bias-improved-KSG estimator (BI-KSG). The asymptotic theoretical guarantees we show of the BI-KSG estimator are the same as the KSG estimator, but the improved performance can be seen empirically -- especially for moderate values of $N$.
\item We demonstrate sharp bounds on the $\ell_2$ rate of convergence of the KL estimator of (differential) entropy for arbitrary $k$ and arbitrary dimensions $d$, showing that the parametric rate of convergence of $1/\sqrt{N}$ is achievable when $d\leq 2$.
\end{itemize}
In the rest of the paper, we mathematically summarize these main results, following up with detailed empirical evidence. A key building block for our results is the asymptotic analysis of the theoretical properties of the KL estimator of differential entropy which we begin with below.

\subsection{KL Entropy Estimator and Convergence Rate}
Consider a random variable $X \in {\cal X}\subseteq {\mathbb R}^d$. Given $N$ i.i.d.\ samples $X_1,X_2,\ldots,X_N$ from the underlying probability density function $f_X(x)$, we want to estimate the differential entropy $H(X) = -{\mathbb E}[\log f_X(X)]$.  As mentioned earlier, a popular approach to estimate
the entropy from i.i.d.\ samples is to use $k$-NN statistics.
     Precisely, let $\rho_{k,i,p}$ denote the distance from $X_i$ to the $k^{\rm th}$ nearest neighbor as measured in $\ell_p$ distance, for some $p \geq 1$.
    Each $k$-NN distance  $\rho_{k,i,p} $ together with the choice of $k$, provides a {\em local} view of the underlying distribution around the $i^{\rm th}$ sample. Informally,
    considering the $\ell_{p}$-ball of radius $\rho_{k,i,p} $ centered at $X_i$ with sufficiently small radius, one can relate the distribution and the number of  samples within the ball via:
%       \begin{eqnarray}
   $    \hf_{X}(X_i) c_{d,p} (\rho_{k,i,p} )^d \simeq  \frac{k}{N}$,
%       \end{eqnarray}
       where $c_{d,p}$ is the volume of the unit $\ell_p$ ball in $d$ dimensions:  %\cite{wang2005volumes}:
       $(\Gamma(1+\frac{1}{p})^d/\Gamma(1+\frac{d}{p})) 2^d$.
    This simple intuition led Kozachenko and Leonenko to design a powerful and provably consistent {\em differential entropy estimator} in \cite{KL87}, which we have called the KL estimator. We begin with the resubstitution estimator $\hH(X) = - \frac{1}{N} \sum_{i=1}^N \log\hf_{X}(X_i) $ and combine it with the $k$-NN estimate of the density to get:
    \begin{eqnarray}
     \hH_{\rm KL}(X)   &=& \frac{1}{N} \sum_{i=1}^N  \log\left(\frac{N c_{d,p} (\rho_{k,i,p} )^{d}}{k}\right)  +  \log (k)- \psi(k) \; \;,
        \label{eq:KL2}
    \end{eqnarray}
    where
    $\psi(x)$ is the digamma function defined as  $\psi(x) = \Gamma^{-1}(x) d\Gamma(x)/dx$, and
    for large $x$, it is approximately equal to $\log(x)$ up to a correction of $O(1/x)$. Precisely, $\psi(x) = \log x - 1/2x + o(1/x)$.
        The correction term, introduced in  \cite{KL87}, is crucial for
    debiasing the estimator.
    Note that if we choose $k$ increasing with $N$, as commonly done in a significant part of the literature (and summarized in a later section),
    $\psi(k)$ converges to $\log(k)$ and no correction is necessary for consistency.
    However, in practice, $k$ is typically a small constant and the correction is crucial.
 %   The surprise here is that,
  %  not only were  Kozachenko and Leonenko  able to compute the exact correction term, but it turns out to be a relatively simple digamma function.
 %   This type of an estimate is known as the {\em resubstitution} type in the literature, e.g.,  \cite{BDG97},    and  refers to a family of estimators
  %  that first replaces the expectation by  the sample mean, i.e. $\E[\log f_{X}(X)] \simeq (1/N)\sum_{i=1}^N \log \hf_{X}(X_i)$, and then  substitutes
   % $\log \hf_{X}$  with an appropriately chosen estimate.
  %  Popular choices for resubstituting $\hf_{X}$ are based either on $k$-NN statistics or kernel density estimators.
%
    Consistency of the KL estimator has been established for $k=1$ by the original authors \cite{KL87} and for general $k$ by \cite{singh2003nearest} and the rate of convergence of the bias and variance has been established (for a certain large class of  smooth pdfs with unbounded support, including the Gaussian) only for one-dimensional random variables  \cite{tsybakov1996root}.

\bigskip\noindent
{\bf Main result.}  We show the following  result on the asymptotic rate of convergence of the KL estimator, over a class of pdfs with {\em bounded} support which includes the uniform and truncated Gaussian.  Below $d$ is the dimension of the random variable whose differential entropy is being estimated and the $\tO$-notation denotes the limiting behavior up to polylogarithmic factors in $N$.
\begin{theorem}
    \label{thm:bias_KL_short}
    The bias of the KL estimator is $\tO(N^{-\frac{1}{d}})$  and the variance is $\tO(1/N)$. Thus the $\ell_2$ error of the KL estimator is $\tO(\frac{1}{\sqrt{N}} + N^{-\frac{1}{d}})$.
\end{theorem}
We note that the parametric rate of convergence is obtained for $d \leq 2$. The result for $d=1$ is also new since our result holds for pdfs with bounded support, a class that were not included in the conditions for a similar result in \cite{tsybakov1996root}.
We briefly highlight the key ideas of the proof below and relegate the precise statement of the theorem in Section~\ref{sec:kl} and its proof to Section \ref{sec:proof_bias_KL}.
 \begin{enumerate}
\item We use the average pdf of a ball $B(x,r)$ centered at $x$ with small radius $r$ (usually the $k$-NN distance of $x$) to  approximate  $f(x)$, relying on the smoothness of  $f(\cdot)$. The error in this approximation is $O(r^2)$ if the Hessian of $f$ is bounded, and this error dominates the  convergence rate of bias. A similar idea was attempted in ~\cite{nilsson2007estimation}, but the authors mistakenly claimed that the error introduced by the approximation is $O(r^{2d})$, which is much smaller than $O(r^2)$ for $d>1$ and leads to an incorrect conclusion.
\item  If the density $f(x)$ is extremely small, the $k$-NN distance $r$ of $x$ will be large, which means that the $O(r^2)$ error is large. We truncate the $k$-NN distance by $a_N$ to solve this problem, but at the cost of additional bias. We need to control the total probability of the tail of $f$ to be small enough so that the additional bias introduced by truncation is not too large; this leads to the necessity of the assumption on the pdfs to be essentially uniformly lower bounded almost everywhere.
\end{enumerate}

%\textbf{TODO: Compare assumption with Tsybakov v.d.Meulen} This theorem  can be viewed as a generalization of the result %in \cite{tsybakov1996root}, under slightly more general conditions for the class of pdfs as considered in %\cite{tsybakov1996root} (we require pdfs to be essentially uniformly lower bounded almost everywhere).

\subsection{KSG Estimator: Consistency and Convergence Rate}
\label{sec:ksg}
Consider two random variables
 $X$ in $\cX\subseteq \reals^{d_x}$ and $Y$ in $\cY\subseteq\reals^{d_y}$.
     Given $N$ i.i.d.\ samples
      $(X_1,Y_1), \dots, (X_N,Y_N)$ from the underlying joint probability density function $f_{X,Y}(x, y)$, we want to estimate the mutual information $I(X;Y)$.
    Mutual information between two random variables $X$ and $Y$ is the sum and difference of differential entropy terms: $I(X;Y) = H(X) + H(Y) - H(X,Y)$. Thus given KL entropy estimator, there is a straightforward and consistent  estimation of the mutual information:
    \begin{eqnarray}
        \widehat{I}_{\rm 3KL}(X;Y) &=& \hH_{\rm KL}(X)+\hH_{\rm KL}(Y)-\hH_{\rm KL}(X,Y).
        \label{eq:3-KL}
    \end{eqnarray}
    While this estimator performs fairly well in practice, the authors of \cite{Kra04} introduced a simple, but inspired, modification of the 3KL estimator that does even better.
    Let
    $n_{x,i,p} \equiv \sum_{j \neq i} \mathbb{I}\{\|X_j - X_i\|_p \leq \rho_{k,i,p} \}$, which can be interpreted as the number of samples that are within a $X$-dimensions-only distance of $\rho_{k,i,p}$ with respect to sample $i$. Since $\rho_{k,i,p}$ is the $k$-NN distance (in terms of both the dimensions of $X$ and $Y$) of the sample $i$ it must be that $n_{x,i,p} \geq k$. Finally, $n_{y,i,p}$ is defined analogously. The KSG estimator measures distances using the $\ell_{\infty}$ norm, so $p = \infty$ in the notation above.

    The KSG mutual information estimator introduced in \cite{Kra04} is given by:
     \begin{eqnarray}
        \hI_{\rm KSG}(X;Y) &\equiv & \psi(k) + \log N - \frac1N \sum_{i=1}^N \left( \, \psi(n_{x,i,\infty}+1) + \psi(n_{y,i,\infty}+1)\,\right),
    \label{def:KSG}
    \end{eqnarray}
where $\psi(x) = \Gamma^{-1}(x) d\Gamma(x)/dx$ is the digamma function.
Observe that the estimate of the joint differential entropy $H(X,Y)$ is done exactly as in the KL estimator using fixed $k$-NN distances, but the KL estimates of $H(X)$ and $H(Y)$ are done using $n_{x,\cdot,\infty}$ and $n_{y,\cdot,\infty}$ NN distances, respectively, which are {\em sample dependent}. The point is that by this choice, the $k$-NN distance terms are canceled away exactly, although it is not clear why this would be a good idea. In fact, it is not even clear if the estimator is consistent. On the other hand, the authors of   \cite{Kra04} showed empirically that the KSG estimator is uniformly superior to the 3KL estimator in many synthetic experiments. A theoretical understanding of the KSG estimator, including a mathematical justification for the improved performance, has been  missing  in the literature.    Our main results fill this gap.

\bigskip\noindent
{\bf Main result.}
One of our main results is to show that the KSG estimator is indeed consistent. We prove this result by deriving a vanishingly small upper bound on the bias,  subject to regularity conditions on the Radon-Nikodym derivatives of $X$ and $Y$ and  standard smoothness conditions on the joint pdf which includes both bounded and unbounded supports. The formal statement of these assumptions is in Section~\ref{sec:KSG} and the proof of the theorem is moved to Sections \ref{sec:proof_KSG_bias} and \ref{sec:proof_KSG_variance}.

We show the following result on the asymptotic rate of convergence of the KSG estimator.
    \begin{theorem}
        \label{thm:KSG_short}
        The KSG estimator is consistent. The bias of the KSG estimator is $\tO(N^{-\frac{1}{d_x+d_y}})$  and the variance is $\tO(1/N)$. Thus the $\ell_2$ error of the KL estimator is $\tO(\frac{1}{\sqrt{N}} + N^{-\frac{1}{d_x+d_y}})$.
    \end{theorem}
    Observe that when $d_x = d_y$ and equal to 1, the rate of convergence is $\tO(\frac{1}{\sqrt{N}})$, the parametric rate of $\ell_2$ error, which cannot be improved upon.

    The correlation boosting explanation allows us to propose a new mutual information estimator, that
     we call the bias-improved KSG (BI-KSG) estimator. The new aspects include using the $\ell_2$ norm
    to measure distances and replacing the digamma function of $n_{x,2,\cdot,2}, n_{y,2,\cdot,2}$ by the logarithm -- and although the theoretical properties of the BI-KSG estimator we show are the same as that of the KSG
    estimator we empirically demonstrate its improved performance which is pronounced when $k$ is small
    and $N$ is moderate-valued.
    The formal definition of the BI-KSG estimator is the following.
    \begin{eqnarray}
        \hI_{BI-KSG}(X;Y) &\equiv& \psi(k) + \log N + \log \left(\, \frac{c_{d_x,2} c_{d_y,2}}{c_{d_x+d_y,2}} \,\right)- \frac{1}{N} \sum_{i=1}^N \left(\, \log(n_{x,i,2}) + \log(n_{y,i,2}) \,\right) \;,
    \end{eqnarray}
where $c_{d,2} = \pi^{d/2}/\Gamma(\frac{d}{2}+1)$ is the volume of $d$-dimensional unit $\ell_2$ ball.

%We note that these results only provide an understanding of the bias of the KSG estimator. Clearly the MSE improvement of the KSG estimator over the 3KL estimator must come from reduced variance, and  an important and interesting direction of future research is to have a theoretical understanding of the variance of the KSG estimator -- we conjecture that the variance is $\tO(1/N)$ regardless of the dimension.

\subsection{Outline of this paper}

In the next two sections we state our main results formally, also providing brief sketches of, and intuitions
behind, the corresponding proofs. Detailed proofs are relegated to the appendix. In Section~\ref{sec:CB} we discuss
the insights behind the KSG estimator: the correlation boosting effect and how this understanding leads to
the BI-KSG estimator with improved empirical performance. In Section~\ref{sec:MMI} we discuss generalization of KSG estimator to multivariate mutual information estimators. Section~\ref{sec:priorwork} puts our results in context
of the vast literature on entropy (and mutual information) estimators. Finally, the proofs of the main results are in
Sections \ref{sec:proof_bias_KL} through \ref{sec:proof_bino_tr}.

%----------------------------------------------------------------------------------------------------------------------------
\section{Convergence Rate of KL Entropy Estimator}
    \label{sec:kl}

 In this section we carefully analyze the performance of the KL estimator of differential entropy in terms of its $\ell_2$ error.  We show upper bounds to the rate of convergence of the bias and variance of the KL estimator separately which combine to provide an upper bound on the $\ell_2$ error. A  minimax lower bound on the $\ell_2$ error provides a baseline to understand how sharp our upper bound characterization is.  We start with the upper bound on the convergence rate of $\ell_2$ error.

\subsection{Upper Bounds}
\label{subsec:ach}
The starting point for our exploration is the pioneering work of
\cite{tsybakov1996root}, which established the  $\frac{1}{\sqrt{N}}$-consistency of the  one-dimensional KL  estimator.
 In particular,  \cite{tsybakov1996root}  proved that
the KL estimator  achieves $\sqrt{N}$-consistency in mean, i.e. $\E[\widehat{H}(X)]-H(X) = \tO(1/\sqrt{N})$,
and in variance,
i.e. $\E[(\widehat{H}(X)-\E[\widehat{H}(X)])^2] = \tO(1/N)$, under the assumption that
the $X$ is a one-dimensional random variable and the estimator uses only the nearest neighbor distance with $k=1$, along with a host of other assumptions on the class of pdfs under consideration (an important one is that the support be unbounded).
We prove a generalization of this rate of convergence
for general dimensions  $d$ and for a general $k$, but under  technical assumptions listed below; some of them  mirror the assumptions introduced in \cite{tsybakov1996root}, but the condition on the support is crucially different.
\begin{assumption}
\label{assumption_rate_ent}
We make the following assumptions: there exist finite constants $\Ca,\Cb,\Cc,\Cd$, and $\C0$ such that
\begin{itemize}
  \item[$(a)$] $f(x) \leq \Ca < \infty$ almost everywhere;
  \item[$(b)$] There exists $\gamma > 0$ such that $\int f(x) \left(\, \log f(x) \,\right)^{1+\gamma} dx \leq \Cb < \infty$;
  \item[$(c)$] $\int f(x) \exp\{-b f(x)\} dx \leq \Cc e^{-\C0 b}$ for all $b > 1$.
%  \item[$(d)$] The pdf $f(x)$ is smooth. We propose two different smoothness assumptions:
 %               \begin{itemize}
                    \item[$(d)$] $f(x)$ is twice continuously differentiable and the Hessian matrix  $H_f$ satisfy  $\|H_f(x)\|_2 < \Cd$ almost everywhere.
  %                  \item[$(d.2)$] $f$ belongs to the  H\"{o}lder Class $\Sigma(s, C_d)$, i.e., for any tuple $r = (r_1, \dots, r_d)$, define $D^r = \frac{\partial^{r_1 + \dots + r_d}}{\partial x_1^{r_1} \dots \partial x_d^{r_d}}$. Then for any $r$ such that $\sum_j r_j < s$. we have:
   %                 \begin{eqnarray}
    %                \|D^r(x) - D^r(y)\| \leq C_d \|x-y\|^{s-\sum_j r_j} \;.
     %               \label{eq:Holder_class}
      %              \end{eqnarray}
       %             for any $x, y$.
        %        \end{itemize}
    \item[$(e)$] The set of points which violates assumption (d) has finite $d-1$ dimensional Hausdorff measure, i.e. $H^{d-1} \left(\, \{x: \|H_f(x)\|_2 \geq C_d\} \,\right) < C_e$.

\end{itemize}
\end{assumption}

These  assumptions are slightly stronger than those in \cite{tsybakov1996root}, where
 assumption $(a)$ and $(e)$ are not required (and with some technical finesse can perhaps be eliminated here as well), assumption $(b)$ was mildly weaker requiring only $\int f(x)|\log f_x(x)| dx <\infty$, and assumption $(c)$ was weaker requiring only $\int f(x) \exp\{-b f(x)\} \leq O(1/b)$.
 The assumption $(c)$ is satisfied for any distribution with bounded support and pdf bounded away from zero.
 This assumption provides a sufficient condition to bound the average effect of the truncation.
 %is satisfied for pdfs with exponentially decaying tails, for example, the Gaussian pdf and is also immediately met by any distribution with bounded support.
 %It appears that this extra assumption on the tails is essential to handle general dimensions, especially when obtaining a tight upper bound on the bias.
Our analysis can be generalized to relax this assumption on the smoothness, requiring
only $\int f(x) \exp\{-b f(x)\} dx \leq \Cc b^{-\beta}$ for all $b > 1$, in which case the resulting guarantees will also depend on $\beta$.
This recovers the result of
\cite{tsybakov1996root} with $\beta=1$ which holds for $d=1$, and we
 assume stronger conditions here since we seek sharp convergence rates in higher dimensions.
 The assumption $(d)$ assumes that the pdf is reasonably smooth, and it is essential for NN-based methods.
  More general families of smoothness conditions have been assumed for other approaches,
 such as the H\"older condition,
 and we have made formal comparisons in Section \ref{sec:priorwork}.

 Note that there exist (families of) distributions,
  satisfying the assumptions $(a)$--$(d)$, where the convergence rates of $k$NN estimators can be made arbitrarily slow.
  Consider a family of distributions in two dimensional rectangle with uniform measure parametrized by $\ell$, such that one side has a length $\ell$ and the other $1/\ell$.
This family of distributions has differential entropy zero. However, for any sample size $N$,
there exists $\ell$ large enough such that the $k$-NN distances  are arbitrarily large and the estimated entropy is also large. To provide a sharp convergence rate for $k$NN estimators, we need to restrict the space of distributions by adding appropriate assumptions that captures this phenomenon.

The challenge in the above example has been addressed under the notion of {\em boundary bias}.
$K$NN distances are larger near the boundaries, which results in underestimating the density at boundaries. This effect is prominent for those distributions that $(i)$ have non-smooth boundaries such as a uniform distribution on a compact support,  and $(ii)$ have large surface area at the boundary. There are two solutions; either we strengthen Assumption 1.$(d)$ and require twice continuously differentiability {\em everywhere} including the boundaries or we can add another assumption on the surface area of the boundaries. In this paper, we take the second route. The reason is that the first option conflicts with the current Assumption 1.($c$) where the only examples we know have lower bounded densities, which implies non-smooth boundaries. It is an interesting future research direction to relax assumption $(c)$ as suggested above, and capture the tradeoff between the lightness of the tail in $\beta$ and also the smoothness in the boundaries.

Instead, we assume in 1.$e$ that the surface area of the boundaries is finite. Recall that the Hausdorff measure of a set $S$ is defined as
 \begin{eqnarray}
    H^{d-1}(S) = \lim_{\delta\to0} \,\inf_{\{U_i\}_{i=1}^\infty} \,\Big\{ \sum_{i=1}^\infty ({\rm diam}\, U_i)^{d-1} \,:\, \bigcup_{i=1}^\infty U_i \supseteq S \,,\, {\rm diam}\, U_i<\delta \big\}\;.
 \end{eqnarray}
 It is a measure of the surface area of the set $S$.
 Note that this could be unbounded for the boundary of a family of distributions, as is the case for the uniform rectangle example above.
 Assumption 1.$(e)$ restricts it to be finite, allowing us to limit the boundary bias to $\tilde{O}(N^{-1/d})$ as proved using  Lemma \ref{lem:F_nx}.
 Since in the (smooth) interior of the support, the bias is $\tilde{O}(N^{-2/d})$, the boundary bias dominates the error for the proposed $k$NN method.

% a standard feature in all works dealing with convergence rates.

We start with  a truncated version of the KL estimator,  similar in spirit to \cite{tsybakov1996root}. Consider $\rho_{k,i,p}$ be the distance to the $k^{\rm th}$ nearest neighbor of $X_i$ with respect to $\ell_p$ distance. Fix any $\delta > 0$, define the threshold $a_N$ as:
\begin{eqnarray}
a_N = \left(\, \frac{(\log N)^{1+\delta}}{N}\,\right)^{1/d} \;, \label{eq:a_N_ent}
\end{eqnarray}
for some $\delta > 0$. We define a local estimate $\xi_{k,i,p}(X)$ by:
\begin{eqnarray}
\xi_{k,i,p}(X) = \begin{cases}
                - \psi(k) + \log N + \log c_{d,p} + d \log \rho_{k,i,p} &\;, \textrm{ if } \rho_{k,i,p} \leq a_N \;,\\
                0 &\;, \textrm{ if } \rho_{k,i,p} > a_N.
                \end{cases}
\end{eqnarray}
Then the truncated KL estimator is:
\begin{eqnarray}
    \widehat{H}_{\rm tKL} (X) &\equiv &  \frac1N \sum_{i=1}^N \xi_{k,i,p}(X) \;.
    \label{def:truncated_KL}
\end{eqnarray}

The following theorem upper bounds  the bias of the truncated KL entropy estimator.
 Here  $\delta>0$ is arbitrarily small (and is from the truncation threshold cf.\ Equation\eqref{eq:a_N_ent}) and $d$ is the dimension of the random variable  $X$ and $k$ is any fixed finite integer and for any norm $p$.

%
% The main ideas of the proof are:
% \begin{enumerate}
%    \item We use the average pdf of a ball $B(x,r)$ centered at $x$ with small radius $r$ (usually the $k$-NN distance of $x$) to  approximation  $f(x)$, relying on the smoothness of  $f(\cdot)$. The error in this approximation is $O(r^2)$ if the Hessian of $f$ is bounded, and this error dominates the  convergence rate of bias. A similar idea was attempted in ~\cite{nilsson2007estimation}, but the authors mistakenly claimed that the error introduced by the approximation is $O(r^{2d})$, which is much smaller than $O(r^2)$ for $d>1$ and leads to an incorrect conclusion.
%    \item If the density $f(x)$ is extremely small, the $k$-NN distance $r$ of $x$ will be large, which means that the $O(r^2)$ error is large. We truncate the $k$-NN distance by $a_N$ to solve this problem, but at the cost of additional bias. We need to control the total probability of the tail of $f$ to be small enough so that the additional bias introduced by truncation is not too large; this leads to the necessity of  Assumption~\ref{assumption_rate_ent}.$(c)$.
%\end{enumerate}

 %
 %  Say 3-4 sentences on what goes into the proof here. Specifically point out the new ideas needed beyond the 1-dim case
 %  Point out the bug in the IT Trans paper which claimed to do something like this.
 %  Also worth pointing out intuitively, in words, where the tail bound really comes in.
 %
\begin{theorem}
    \label{thm:bias_KL}
    Under the Assumption \ref{assumption_rate_ent} and for finite $k=O(1)$ and $d=O(1)$,
    the bias of the truncated KL entropy estimator using $N$  i.i.d.\  samples is bounded by:
    \begin{eqnarray}
        \E \left[\, \widehat{H}_{\rm tKL}(X) \, \right] - H(X)  = O\left(\, \frac{\left(\, \log N \,\right)^{(1+\delta)(1+1/d)}}{N^{1/d}}  \,\right).
        \label{eq:bias_KL}
    \end{eqnarray}

 %   Under the Assumption \ref{assumption_rate_ent} $(a), (b), (c)$ and $(d.2)$
  %  the bias of the truncated KL entropy estimator using $N$ i.i.d.\ samples is bounded by:
  %  \begin{eqnarray}
  %      \E \left[\, \widehat{H}_{\rm tKL}(X) \, \right] - H(X)  = O\left(\, \frac{\left(\, \log N \,\right)^{(1+\delta)(1+s/d)}}{N^{s/d}} + \frac{\left(\, \log N\,\right)^{2(1+\delta)}}{N} \,\right).
  %      \label{eq:bias_KL_Holder}
  %  \end{eqnarray}

\end{theorem}

The following theorem establishes the upper bound for the variance of $\widehat{H}_{\rm tKL}(X)$, cf.\ \eqref{def:truncated_KL}, which we observe is independent of the dimension $d$ of the random variable $X$.  Again  $\delta>0$ is arbitrarily small (and is from the truncation threshold cf.\ Equation\eqref{eq:a_N_ent}) and $k$ is any fixed integer.

The main step of the proof is the observation that
\begin{eqnarray}
    \V \left[ \, \widehat{H}_{\rm tKL}(X) \, \right] &\leq& \frac{1}{N} \V \left[ \, \xi_{k,1,p} \, \right] + \Cov \left[\, \xi_{k,1,p}, \xi_{k,2,p}\,\right] .
\end{eqnarray}
The first term is bounded by $O((\log \log N)^2)$ due to the truncation of $k$-NN distances. The second term is actually the covariance of $k$-NN distances of a pair of  samples, which we show  to be $O(1/N)$ up to a polylogarithmic factor. Putting these two steps together completes the proof.
%
 %  Say 3-4 sentences on what goes into the proof here. Specifically point out the new ideas needed beyond the 1-dim case
 %  Which assumptions really matter here, etc. How come the bound doesn't depend on dimension, etc.
 %
 %
\begin{theorem}
    \label{thm:variance_KL}
    Under the Assumption~\ref{assumption_rate_ent} and for finite $k=O(1)$ and $d=O(1)$,
    the variance of the truncated KL entropy estimator using $N$ i.i.d. samples is bounded by:
    \begin{eqnarray}
        {\rm Var} \left[ \, \widehat{H}_{\rm tKL}(X) \, \right]   = O\left(\, \frac{ \left(\, \log \log N \,\right)^2 \, \left(\, \log N \,\right)^{(2k+2)(1+\delta)}}{N} \,\right).
        \label{eq:variance_rate_KL}
    \end{eqnarray}
\end{theorem}

The Mean Squared Error (MSE) of truncated KL estimator
\begin{eqnarray}
\E \left[\, \left(\, \widehat{H}_{\rm tKL}(X)  - H(X)  \,\right)^2 \,\right] &=&  \E \left[\,  \widehat{H}_{\rm tKL}(X)  - H(X)  \,\right]^2 + \V \left[ \, \widehat{H}_{\rm tKL}(X) \, \right],
\label{def:MSE_KL}
\end{eqnarray}
is the sum of the squared bias and variance. So
combining Theorems~\ref{thm:bias_KL} and ~\ref{thm:variance_KL}, we obtain the following upper bound on the MSE of truncated KL estimator. Again  $\delta>0$ is arbitrarily small (and is from the truncation threshold cf.\ Equation\eqref{eq:a_N_ent}) and $k$ is any fixed integer.

\begin{corollary}
    \label{cor:l2_KL}
    Under the Assumption \ref{assumption_rate_ent} and for finite $k=O(1)$ and $d=O(1)$,
    the MSE of the truncated KL entropy estimator using $N$ i.i.d.\ samples is bounded by:
    \begin{eqnarray}
        \E \left[\, \left(\, \widehat{H}_{\rm tKL}(X)  - H(X)  \,\right)^2 \,\right] = O\left(\, \frac{\left(\, \log N \,\right)^{(1+\delta)(2+2/d)}}{N^{2/d}} + \frac{\left(\, \log \log N \,\right)^2 \left(\, \log N \,\right)^{(2k+2)(1+\delta)}}{N} \,\right).
        \label{eq:l2_KL}
    \end{eqnarray}

%    Under the Assumption \ref{assumption_rate_ent} $(a), (b), (c)$ and $(d.2)$
%    the MSE of the truncated KL entropy estimator using $N$ i.i.d. samples is bounded by:
 %   \begin{eqnarray}
  %      \E \left[\, \left(\, \widehat{H}_{\rm tKL}(X)  - H(X)  \,\right)^2 \,\right] = O\left(\, \frac{\left(\, \log N \,\right)^{(1+\delta)(2+2s/d)}}{N^{2s/d}} + \frac{\left(\, \log \log N \,\right)^2 \left(\, \log N \,\right)^{(2k+2)(1+\delta)}}{N} \,\right).
 %       \label{eq:l2_KL_holder}
 %   \end{eqnarray}

\end{corollary}
To see how good this bound on rate of convergence is, we derive a worst case lower bound below.

\subsection{Minimax Lower Bound}
\label{subsec:lb}
We follow the standard techniques to lower bound estimator errors of functionals of a density -- Le Cam's method in general and \cite{birge1995estimation} in particular. Consider the class of smooth distributions:
\begin{eqnarray}
\mathcal{F}_d = \{f: \mathbb{R}^d \to \mathbb{R}^+: \int f(x) dx = 1, \|H_f(x)\| \leq C, a.e.\} \;,
\end{eqnarray}
where $H_f$ denotes the Hessian matrix of $f$.
%Also, we consider the density functions over $[0,1]^d$ in $s$-th  H\"{o}lder class:
%\begin{eqnarray}
%\mathcal{D}_{\Sigma(s,L)} = \{f: \mathbb{R}^d \to \mathbb{R}^+: \int f(x) dx = 1, f \in \Sigma(s,L)\} \;,
%\end{eqnarray}
We want to estimate the differential entropy of $f$ from $n$ i.i.d.\ samples $\{X_i\}_{i=1}^n$, where $X_i \in \mathbb{R}^d$.  We summarize  a minimax lower bound on the $\ell_2$ error rate in the following theorems.
%
%  2-3 sentences on what techniques went in here. Which parts are obvious, which ones are somewhat special.
%
Here  $\Omega(N^{-1})$ is the parametric minimax lower bound and  $\Omega(N^{-16/(d+8)})$ follows from the construction in \cite{birge1995estimation}.

\begin{theorem}
\label{thm:minimax_lower_bound}
The minimax error rate for estimating entropy from $N$ i.i.d. samples is lower bounded by
\begin{eqnarray}
\inf_{\widehat{H}_N} \; \sup_{f \in \mathcal{F}_d}\;  \E \left[\, \left(\, \widehat{H}_N(X) - H(X) \,\right)^2 \,\right] &\geq& \Omega(N^{-16/(d+8)} + N^{-1}) \;,
\label{minimax_lower_bound}
%\inf_{\widehat{H}_N} \sup_{f \in \mathcal{D}_{\Sigma(s, L)}} \E \left[\, \left(\, \widehat{H}_N(X) - H(X) \,\right)^2 \,\right] &\geq& \Omega(N^{-8s/(d+4s)} + N^{-1}) \;. \label{minimax_lower_bound_Holder}
\end{eqnarray}
where the infimum is taken over all measurable functions over the $N$  samples.
\end{theorem}

\subsection{Comparing the Bounds}

\begin{figure}
\begin{minipage}[b]{0.5\linewidth}
	\begin{center}
	\includegraphics[width=1.0\textwidth]{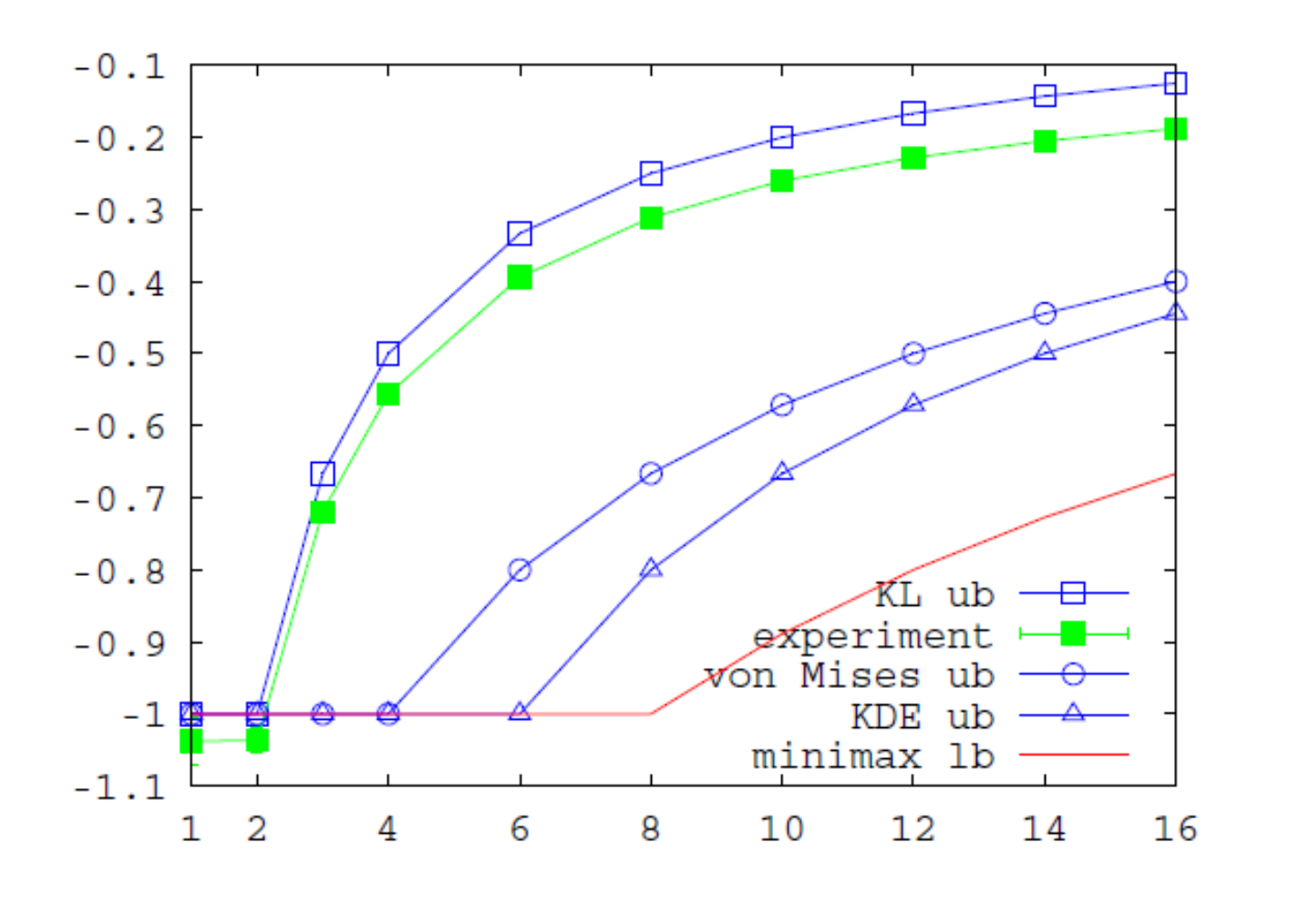}
	\put(-250,165){$\log (\E[(\widehat{H}(X)-H(X))^2])/\log N$}
	\put(-145,0){dimension $d$}
	\end{center}
	\caption{Exponents of the  convergence rate of $\ell_2$ error for various entropy estimators.}
		\label{fig:ub_vs_lb}
%\end{figure}
	\end{minipage}
	\quad \quad
%\begin{minipage}[b]{0.5\linewidth}
%	\begin{center}
%	\includegraphics[width=1.0\textwidth]{figure_6.pdf}
%	\end{center}
%	\caption{MSE versus sample size  in log scale.}
%	\label{fig:mse_vs_n}
%	\end{minipage}
%\begin{figure}
\begin{minipage}[b]{0.5\linewidth}
	\begin{center}
	\includegraphics[width=1.0\textwidth]{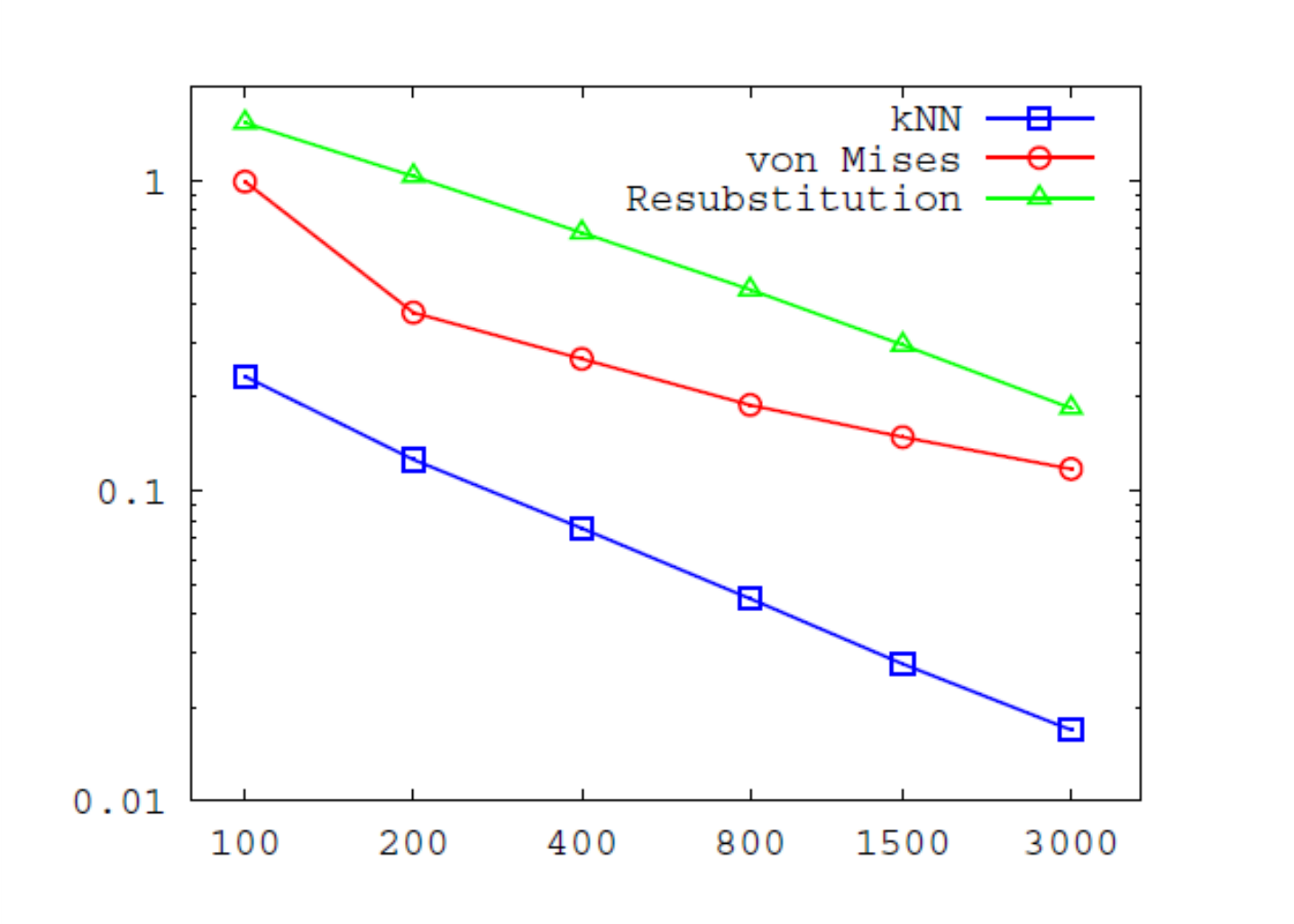}
	\put(-250,165){$\E[(\widehat{H}(X)-H(X))^2]$}
	\put(-145,0){sample size $N$}
	\end{center}
	\caption{Empirical performance of MSE vs sample size (log-log scale).}
		\label{fig:empiricalKL}
	\end{minipage}
\end{figure}

The minimax $\ell_2$ error of the KL estimator (over the class of functions with norm-bounded Hessian matrix) is lower bounded by $\tO(\frac{1}{\sqrt{N}} + N^{-\frac{8}{d+8}})$ (cf.\ this broadly follows from \cite{birge1995estimation}, but a detailed proof is also presented in  Section \ref{sec:proof_lowerbound} for completeness), we see that
the optimality gap of the exponent is characterized by  $\min\{1/2,8/(d+8)\}-\min\{0.5,1/d\}$, which is always non-negative.
    This characterizes the $\frac{1}{\sqrt{N}}$ rate of convergence for MSE for $d \leq 2$ (this is the parametric rate), while there is some gap in the upper and lower bounds for the rates when $d > 2$.  The upper and lower bounds of the MSE error of the KL estimator as a function of the number of samples is depicted in Figure~\ref{fig:ub_vs_lb} (along with the exponents for other entropy estimators: resubstitution \cite{joe1989estimation} and von Mises expansion estimators \cite{kandasamy2015nonparametric} with standard KDEs). We see that the upper and lower bounds match for $d \leq 2$ and in this regime the parametric rate of convergence of $\tO(\frac{1}{\sqrt{N}})$ is achieved. There is a gap when $d > 2$ and closing this gap is an interesting future direction of research.

\begin{figure}[h]
	\begin{center}
	\includegraphics[width=0.6\textwidth]{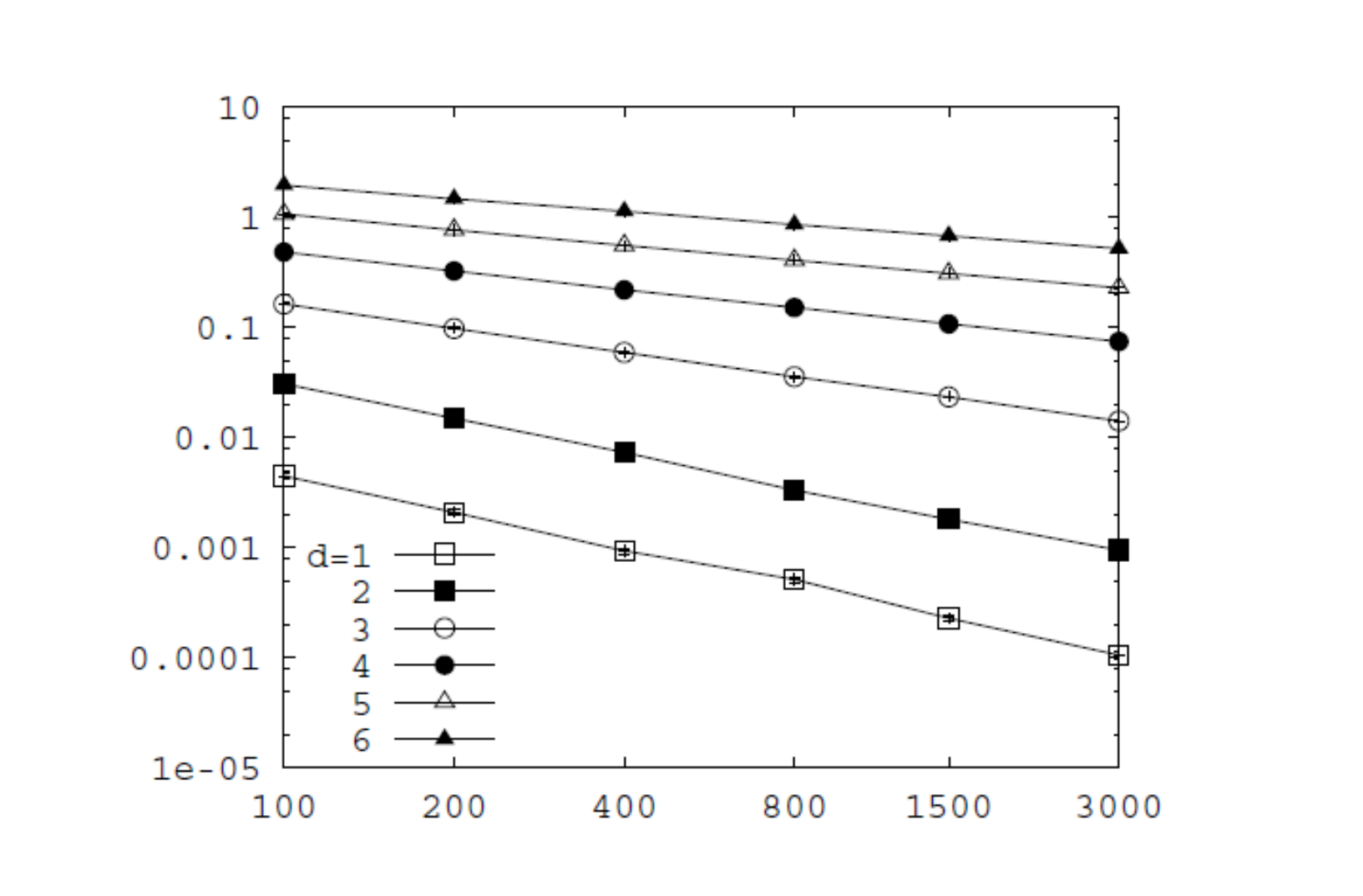}
	\put(-165,0){sample size $N$}
	\put(-250,175){$\E[(\widehat{H}(X)-H(X))^2]$}
	\end{center}
	\caption{MSE versus sample size  in log-log scale.}
	\label{fig:mse_vs_n}
\end{figure}

    To get a feel for whether the upper bound exponent should be improved or the lower bound (or both), it is instructive to plot sample MSE of the KL estimator for a specific pdf.
In this synthetic experiment, we choose $N$ i.i.d.\ samples $X_1, X_2, \dots, X_N$ from uniform distribution over $[0,1]^d$ and use the KL estimator to estimate entropy. Figure.~\ref{fig:mse_vs_n} plots the MSE vs the sample size for different dimensions in log scale; we observe that  $\log (\textrm{MSE})$ is linear in $\log N$. We can use standard linear regression to estimate the slope $\log (\textrm{MSE}) / \log N$ -- the experimental results are plotted  in Figure.~\ref{fig:ub_vs_lb} (using green color). We conclude that the simulation results are fairly close to the theoretical upper bounds on convergence rate -- which suggests that the improvements are to be most expected in lower bounds suited to $k$-NN estimation.

   It is interesting that the theoretical  rate of convergence is slowest for $k$-NN methods as compared to KDE (resubstitution or von-Mises expansions), while the empirical performance (for modest sample sizes) is {\em exactly the reverse} in many diverse settings; Figure~\ref{fig:empiricalKL} illustrates this phenomenon for a specific instance (independent Beta(2,2) in 6 dimensions, with sample sizes varying from 100 to 3000, averaged over 500 trials. The von-Mises estimator is implemented using the default parameters provided by~\cite{kandasamy2015nonparametric}). Clearly the difference in theoretical and empirical performance is to be explained by the constant terms (and not asymptotics in sample size $N$) -- a theoretical understanding of this phenomenon is another interesting direction for future research.

%-------------------------------------------------------------------------------------------------------------------------
\section{KSG Estimator: Consistency and Convergence Rate}
\label{sec:KSG}

A detailed understanding of the KL estimator sets the stage for the main results of this paper: deriving
theoretical properties of the KSG estimator of mutual information. Our main result is that the KSG estimator is consistent, as is our proposed modification, the so-called bias-improved KSG estimator (BI-KSG); these results are under some (fairly standard) assumptions on the joint pdf of $(X,Y)$.

%------------------------------------------
    \subsection{Consistency}
    \label{subsec:KSGconsistency}

We make the following assumptions on the joint pdf of $(X,Y)$. The first assumption is essentially needed to define the joint differential entropy of $(X,Y)$, the second assumption makes some regularity conditions on the Radon-Nikodym derivatives of $X$ and $Y$, and the third assumption is regarding standard smoothness conditions on the joint pdf. We note that these conditions are readily met by most popular pdfs, including multivariate Gaussians, and no assumption is made on the boundedness of the support.

\begin{assumption}
\label{assumption_cons_mi}

\begin{itemize}
  \item[$(a)$] $\int f(x, y) \left|\log f(x, y)\right| dxdy < \infty$.
   \item[$(b)$] There exists a finite constant $C'$ such that the conditional pdf
   %$f_{Y|X}(y|x) = \frac{f(x,y)}{f_X(x)}$ is bounded, i.e.,
   $f_{Y|X}(y|x) < C'$ and $f_{X|Y}(x|y)<C'$ almost everywhere.
   %\item[$(c)$] The pdf $f(x,y)$ is smooth. We propose two different smoothness assumptions:
    %            \begin{itemize}
    \item[$(c)$] $f(x,y)$ is twice continuously differentiable and the Hessian matrix  $H_f$ satisfy  $\|H_f(x,y)\|_2 < C$ almost everywhere.
    %                \item[$(c.2)$] $f$ belongs to the  H\"{o}lder Class $\Sigma(s, C)$.
     %           \end{itemize}
\end{itemize}
\end{assumption}

Under these assumptions, the KSG and the BI-KSG estimators are both consistent, in probability. This is a formal version of Theorems~2 and 4 of the main text.

\begin{theorem}
    Under the Assumption \ref{assumption_cons_mi} and for finite  $k>\max\{d_x/d_y,d_y/d_x\}$, $d_x,d_y=O(1)$, and
    for all $\varepsilon>0$,
    \begin{eqnarray}
        \lim_{N\to\infty} \Pr\left(\, \left|\hI_{\rm KSG}(X;Y)-I(X;Y)\right|>\varepsilon\,\right) = 0\;, \; \text{ and }
        \label{eq:consistent_KSG}
    \end{eqnarray}
    \begin{eqnarray}
        \lim_{N\to\infty} \Pr\left(\, \left|\hI_{\rm BI-KSG}(X;Y)-I(X;Y)\right|>\varepsilon\,\right) = 0\;.
        \label{eq:consistent_BI_KSG}
    \end{eqnarray}
    \label{thm:consistent_KSG}
\end{theorem}

%\begin{theorem}
 %   Under the Assumption \ref{assumption_cons_mi},
  %  if $k>\max\{d_y/d_x,d_x/d_y\}$,
   % then the mutual information estimator $\hI_{k,N}(X;Y)$ converges to
%    the true value in probability, i.e. for all $\varepsilon>0$,
 %   \begin{eqnarray}
  %      \lim_{N\to\infty} \Pr\left(\, \left|\hI_{k,N}(X;Y)-I(X;Y)\right|>\varepsilon\,\right) = 0\;.
   %     \label{eq:convergence}
%    \end{eqnarray}
 %   \label{thm:convergence}
%\end{theorem}

\subsection{Convergence rate}
    \label{subsec:KSGrate}
The KSG and BI-KSG mutual information estimators are  reintroduced here for ease of reference:
     \begin{eqnarray}
        \hI_{\rm KSG}(X;Y) &\equiv & \psi(k) + \log N - \frac1N \sum_{i=1}^N \left( \, \psi(n_{x,i,\infty}+1) + \psi(n_{y,i,\infty}+1)\,\right),
    \label{def:KSG} \\
    \hI_{\rm BI-KSG} (X;Y) &\equiv & \psi(k) + \log N  + \log \Big(\frac{c_{d_x,2}\; \; c_{d_y,2}}{c_{d_x+d_y,2}}\Big) - \frac1N \sum_{i=1}^N \left(\, \log(n_{x,i,2}) + \log (n_{y,i,2}) \,\right)  \;,
    \label{def:BI-KSG}
    \end{eqnarray}
To understand the rate of convergence of the bias of the KSG and BI-KSG estimators, we first truncate  the $k$-NN distance $\rho_{k,\cdot,\cdot}$, similar to the undertaking in Section~\ref{subsec:ach}.   For any $\delta > 0$, let the truncation threshold be:
\begin{eqnarray}
a_N = \left(\, \frac{(\,\log N\,)^{1+\delta}}{N}\,\right)^{1/(d_x+d_y)} \;, \label{eq:a_N}
\end{eqnarray}
where $d_x$ and $d_y$ are the dimensions of the random variables $X$ and $Y$ respectively.
We define  local information estimates $\iota_{k,i,\infty}$ and $\iota_{k,i,2}$ by:
\begin{eqnarray}
\iota_{k,i,\infty} = \begin{cases}
                \psi(k) + \log N - \psi(n_{x,i,\infty} + 1) - \psi(n_{y,i,\infty} + 1) & \textrm{ if } \rho_{k,i,\infty} \leq a_N, \\
                0 & \textrm{ if } \rho_{k,i,\infty} > a_N,
                \end{cases}
\end{eqnarray}
and \begin{eqnarray}
\iota_{k,i,2} = \begin{cases}
                \psi(k) + \log N + \log (\frac{c_{d_x,2}c_{d_y,2}}{c_{d_x+d_y,2}}) - \log(n_{x,i,2}) - \log(n_{y,i,2}) & \textrm{ if } \rho_{k,i,2} \leq a_N, \\
                0 & \textrm{ if } \rho_{k,i,2} > a_N.
                \end{cases}
\end{eqnarray}
The modified (via truncation) KSG and  BI-KSG  estimators (compare with \eqref{def:KSG} and \eqref{def:BI-KSG}) are:
\begin{eqnarray}
\hI_{tKSG} (X;Y) &\equiv &  \frac1N \sum_{i=1}^N \iota_{k,i,\infty}.\\
    \hI_{tBI-KSG} (X;Y) &\equiv &  \frac1N \sum_{i=1}^N \iota_{k,i,2}.
    \label{def:truncated_mi}
\end{eqnarray}

%While retaining Assumption \ref{assumption_rate_ent}, we require an additional condition on the conditional pdfs:
%\begin{assumption}
%\label{assumption_rate_mi}
%\begin{itemize}
%      \item[$(e)$] The conditional pdf $f_{Y|X}(y|x) \leq C_e$ and $f_{X|Y} \leq C_e$ almost everywhere, for some $C_e < \infty$.
%\end{itemize}
%\end{assumption}
The following theorem (a formal version of Theorems~2 and 4 of the main text) provides an upper bound on the rate of convergence of the bias and variance, under the conditions in Assumption \ref{assumption_rate_mi} below, and holds for any $k$ and $\delta > 0$ (parameter in the truncation threshold, cf.\ \eqref{eq:a_N}).

\begin{assumption}
\label{assumption_rate_mi}

We make the following assumptions: there exist finite constants $C_a$,$C_b$,$C_c$,$C_d$,$C_e$,$C_f$,$C_g$,$C_h$ and $C_0$ such that
\begin{itemize}
    \item[$(a)$] $f(x,y) \leq C_a < \infty$ almost everywhere.
    \item[$(b)$] There exists $\gamma > 0$ such that $\int f(x, y) \left(\, \log f(x, y) \,\right)^{1+\gamma} dxdy \leq C_b < \infty$.
    \item [$(c)$] $\int f(x,y) \exp\{-b f(x,y)\} dx dy \leq C_c e^{-C_0 b}$ for all $b > 1$.
  % \item[$(d)$] The pdf $f(x,y)$ is smooth. We propose two different smoothness assumptions:
   %             \begin{itemize}
                    \item[$(d)$] $f(x,y)$ is twice continuously differentiable and the Hessian matrix  $H_f$ satisfy  $\|H_f(x,y)\|_2 < \Cd$ almost everywhere.
    %                \item[$(d.2)$] $f$ belongs to the  H\"{o}lder Class $\Sigma(s, C)$.
     %           \end{itemize}
    \item[$(e)$] The conditional pdf $f_{Y|X}(y|x) < C_e$ and $f_{X|Y}(x|y)<C_e$ almost everywhere.
    \item[$(f)$] The marginal pdf $f_X(x) < C_f$ and $f_Y(y) < C_f$ almost everywhere.
    \item[$(g)$] The set of points violating $(d)$ has finite $d_x+d_y-1$-dimensional Hausdorff measure, i.e., $H^{d_x+d_y-1} \left(\,\{(x,y): \|H_f(x,y)\| \geq C_d\} \,\right) \leq C_g$.
    \item[$(h)$] The set of points such that $H_{f_X}(x)$ or $H_{f_Y}(y)$ is larger than $C_d$ also has finite $d_x-1$ (or $d_y-1$)-dimensional Hausdorff measure, i.e., $H^{d_x-1} \left(\,\{x: \|H_{f_X}(x)\| \geq C_d\} \,\right) \leq C_h$ and $H^{d_y-1} \left(\,\{y: \|H_{f_Y}(y)\| \geq C_d\} \,\right) \leq C_h$.
\end{itemize}
\end{assumption}

Here Assumption~\ref{assumption_rate_mi}.$(a) - (d)$ are the same as in  Assumption~\ref{assumption_rate_ent} (which were introduced in the context of characterizing the convergence rate of the KL estimator). Assumption~\ref{assumption_rate_mi}.$(e)$ makes sure that the marginal entropy estimator converges at  certain rate. Compared to Assumption~\ref{assumption_cons_mi},  we need an upper bound for the joint entropy $(a)$. The condition $(b)$ is slightly stronger than Assumption~\ref{assumption_cons_mi} by changing the power from $1$ to $1+\gamma$. The condition $(c)$ is the tail bound which ensures the convergence rate of truncated KL joint entropy estimator. The conditions Assumption~\ref{assumption_rate_ent}.$(g)$ and $(h)$ are natural generalizations of Assumption~\ref{assumption_rate_ent}.$(e)$. We note that truncated multivariate Gaussians and uniform random variables meet these constraints. 

\begin{theorem}
    Under Assumption~\ref{assumption_rate_mi}, and for finite  $k>\max\{d_x/d_y,d_y/d_x\}$, $d_x,d_y=O(1)$,
    \begin{eqnarray}
      \E \left[\hI_{\rm tKSG}(X;Y) \right] - I(X;Y)  = O\left(\, \frac{\left(\, \log N \,\right)^{(1+\delta)(1+\frac{1}{d_x+d_y})}}{N^{\frac{1}{d_x+d_y}}}
     \,\right)\;.
        \label{eq:bias_KSG}\\
     \E \left[\hI_{\rm tBI-KSG}(X;Y) \right] - I(X;Y)  = O\left(\,  \frac{\left(\, \log N \,\right)^{(1+\delta)(1+\frac{1}{d_x+d_y})}}{N^{\frac{1}{d_x+d_y}}}
     \,\right)\;.
        \label{eq:bias_BI_KSG}
    \end{eqnarray}
    \label{thm:bias_BI_KSG}
\end{theorem}

The following theorem establishes an upper bound for the variance of truncated KSG and BI-KSG estimators.

\begin{theorem}
    Under Assumption~\ref{assumption_rate_mi}, %and for finite  $k>\max\{d_x/d_y,d_y/d_x\}$, $d_x,d_y=O(1)$,
    \begin{eqnarray}
      \V \left[\hI_{\rm tKSG}(X;Y) \right] = O\left(\, \frac{ \left(\, \log \log N \,\right)^2 \left(\, \log N \,\right)^{(2k+2)(1+\delta)}}{N}
     \,\right)\;.
        \label{eq:variance_KSG}\\
     \V \left[\hI_{\rm tBI-KSG}(X;Y) \right] = O\left(\, \frac{ \left(\, \log \log N \,\right)^2 \left(\, \log N \,\right)^{(2k+2)(1+\delta)}}{N}
     \,\right)\;.
        \label{eq:variance_BI_KSG}
    \end{eqnarray}
    \label{thm:variance_BI_KSG}
\end{theorem}

Combining Theorem~\ref{thm:bias_BI_KSG} and Theorem~\ref{thm:variance_BI_KSG}, we obtain the following upper bound on the MSE of truncated KSG or BI-KSG estimator.

\begin{corollary}
    \label{cor:l2_BI_KSG}
    Under the Assumption \ref{assumption_rate_mi} and for finite $k=O(1)$ and $d=O(1)$,
    the MSE of the truncated KSG or BI-KSG mutual information estimator using $N$ i.i.d.\ samples is bounded by:
    \begin{eqnarray}
        \E \left[ \left( \hI_{\rm tKSG}(X;Y)  - I(X;Y)  \right)^2 \right] = O\left( \frac{\left( \log N \right)^{2(1+\delta)(1+\frac{1}{d_x+d_y})}}{N^{\frac{2}{d_x+d_y}}} + \frac{ \left( \log \log N \right)^2 \left( \log N \right)^{(2k+2)(1+\delta)}}{N} \right)\;.
        \label{eq:l2_KSG}\\
        \E \left[ \left( \hI_{\rm tBI-KSG}(X;Y)  - I(X;Y)  \right)^2 \right] = O\left( \frac{\left( \log N \right)^{2(1+\delta)(1+\frac{1}{d_x+d_y})}}{N^{\frac{2}{d_x+d_y}}} + \frac{ \left( \log \log N \right)^2 \left( \log N \right)^{(2k+2)(1+\delta)}}{N} \right)\;.
        \label{eq:l2_BI_KSG}
    \end{eqnarray}

%    Under the Assumption \ref{assumption_rate_ent} $(a), (b), (c)$ and $(d.2)$
%    the MSE of the truncated KL entropy estimator using $N$ i.i.d. samples is bounded by:
 %   \begin{eqnarray}
  %      \E \left[\, \left(\, \widehat{H}_{\rm tKL}(X)  - H(X)  \,\right)^2 \,\right] = O\left(\, \frac{\left(\, \log N \,\right)^{(1+\delta)(2+2s/d)}}{N^{2s/d}} + \frac{\left(\, \log \log N \,\right)^2 \left(\, \log N \,\right)^{(2k+2)(1+\delta)}}{N} \,\right).
 %       \label{eq:l2_KL_holder}
 %   \end{eqnarray}

\end{corollary}

{It is instructive to compare these upper bounds on mean squared error to that of the 3KL estimator, which can be derived directly from Corollary~\ref{cor:l2_KL}.
%\begin{eqnarray}
%\E \left[\hI_{tKSG}(X;Y) \right] - I(X;Y)  = \widetilde{O}\left(N^{-\frac{2}{d_x + d_y}} + N^{-1}\right),
%\end{eqnarray}
%where $\widetilde{O}$-notation denotes that we are neglecting polylogarithmic terms.
We see that the rates of convergence of the mean squared error (at least viewed through the upper bounds on their rates  of convergence) have the same scaling for 3KL, KSG, and BI-KSG.
}
%is  only worse for the KSG (and BI-KSG) estimator as compared to the 3KL estimator, and equal in the important special case below.
%}
\begin{corollary}
    If $d_x = d_y = 1$, we obtain:
    \begin{eqnarray}
          \E \left[\, \left(\, \hI_{\rm tKSG}(X;Y)  - I(X;Y)  \,\right)^2 \,\right]  \; =\;  O\left(\, \frac{(\log N)^{(2k+2)(1+\delta)}}{N} \,\right)\;. \\
        \E \left[\, \left(\, \hI_{\rm tBI-KSG}(X;Y)  - I(X;Y)  \,\right)^2 \,\right]  \; =\;  O\left(\, \frac{(\log N)^{(2k+2)(1+\delta)}}{N} \,\right)\;.
        \label{eq:convergence_rate_cor}
    \end{eqnarray}
\end{corollary}
This establishes the $1/N$ convergence rate of the MSE of the  KSG and BI-KSG and 3KL  estimators up to a poly-logarithmic factor; this (parametric) convergence rate cannot be improved upon.
%
%  need a proper citation for why the parametric rate of 1/sqrt(N) cannot be improved on.
%

%-------------------------------------------------------------------------------------------------------------------------
 \section{Correlation Boosting}
 \label{sec:CB}
  Perhaps to build an  intuition towards a deeper theoretical understanding of the KSG estimator,   we ask for the key features that make it perform better than the 3KL one. This is the focus of the present section, where we see a curious correlation boosting effect which explains the superior performance of the KSG estimator and allows us to derive an even better estimator of mutual information. A related intuitive explanation is provided in \cite{zhu2014bias}. 

\bigskip
\noindent{\bf Correlation Boosting Effect.}
We begin by rewriting the KSG  estimator, cf.\ \eqref{def:KSG},   as:
\begin{eqnarray}
\widehat{I}_{KSG}(X;Y) = \frac{1}{N}\sum_{i=1}^N \iota_{k,i,\infty} = \frac{1}{N} \sum_{i=1}^N \left(\, \xi_{k,i,\infty}(X) + \xi_{k,i,\infty}(Y) - \xi_{k,i,\infty}(X,Y) \,\right)
\end{eqnarray}
where
\begin{eqnarray}
\xi_{k,i,\infty}(X,Y) &\equiv& -\psi(k) + \log N + \log c_{d_x,\infty} c_{d_y,\infty} + (d_x+d_y) \log \rho_{k,i,\infty} \,\notag\\
\xi_{k,i,\infty}(X) &\equiv& -\psi(n_{x,i,\infty}+1 ) + \log N + \log c_{{d_x},\infty} + d_x \log \rho_{k,i,\infty} \,\notag\\
\xi_{k,i,\infty}(Y) &\equiv& -\psi(n_{y,i,\infty}+1 ) + \log N + \log c_{{d_y},\infty} + d_y \log \rho_{k,i,\infty}. \label{eq:xi}
\end{eqnarray}
Here $\xi_{k,i,\infty}(X,Y),\xi_{k,i,\infty}(X)$ and $\xi_{k,i,\infty}(Y)$ are local estimates of the  differential entropies $H(X,Y), H(X)$ and $H(Y)$, respectively, at the $i^{\rm th}$ sample. %One may consider that we can just use the KL entropy estimator~\cite{KL87} to estimate the joint entropy and the marginal entropy separately and estimate the mutual information by $\widehat{I}(X;Y) = \widehat{H}(X) + \widehat{H}(Y) - \widehat{H}(X,Y)$ (which is called a 3-KL mutual information estimator).
We will show that the bias of joint entropy estimate
 $b_{k,i,\infty}(X,Y) = \xi_{k,i,\infty}(X,Y) - H(X,Y)$
is positively correlated to the bias of marginal entropy estimates
$b_{k,i,\infty}(X) = \xi_{k,i,\infty}(X) - H(X)|$
and
$b_{k,i,\infty}(Y) = \xi_{k,i,\infty}(Y) - H(Y).$
Since the bias of the KSG estimator is simply equal to $\frac{1}{N} \sum_{i=1}^N b_{k,i,\infty}(X,Y) - b_{k,i,\infty}(X) - b_{k,i,\infty}(Y)$ the bias is reduced if $b_{k,i,\infty}(X,Y)$ is {\em positively} correlated with $b_{k,i,\infty}(X)$ and $b_{k,i,\infty}(Y)$. The same effect is true for the 3KL estimator, which is already based on estimating the three differential entropy terms separately.  We tabulate the Pearson correlation coefficients of the biases in Table \ref{tab:correlation_boosting} for two exemplar pdfs (independent uniforms and Gaussians). The main empirical observation is that the correlation is positive even for the 3KL estimator but is significantly  higher for the KSG estimator (and at times even higher for the BI-KSG estimator which we introduce below).

\begin{table}[h]
\centering
\begin{tabular}{|c|ccc|ccc|}
 \hline
  & \multicolumn{3}{c|}{$(X,Y) \sim \textrm{Unif}([0,1]^2)$} & \multicolumn{3}{c|}{$(X,Y) \sim \mathcal{N}(0,I_2)$} \\
 \hline
 $N$ & 1024 & 2048 & 4096 & 1024 & 2048 & 4096 \\
 \hline
 3KL & 0.1276 & 0.1259 & 0.0930 & 0.4602 & 0.4471 & 0.3717 \\
 KSG & \bf 0.9312 & \bf 0.9328 & \bf 0.9085 & 0.6750 & 0.7151 & 0.6687 \\
 BI-KSG & 0.9253 & 0.9251 & 0.8880 & \bf 0.6823 & \bf 0.7330 & \bf 0.6939 \\
 \hline
\end{tabular}
\caption{Pearson Correlation Coefficient $\rho\left(\, b(X,Y), b(X) \right)$
%= \Cov\left( b(X,Y), b(X) \right) / \sqrt{\V [b(X,Y)]} \sqrt{\V [b(X)]}$
 for different mutual information estimators.}
\label{tab:correlation_boosting}
\end{table}

% We hypothesize that this correlation {\em boosting} effect is the main reason for the KSG estimator having smaller  mean-square error than the 3KL one and  formally see this effect in the context of an example next.

% \begin{theorem}
%     $X$ and $Y$ be independently and uniformly distributed in $[0,1]$. Then:
%         \begin{eqnarray}
%         \E \left[ b_{k,i,\infty}(X) | b_{k,i,\infty}(X,Y)\right] = \sqrt{\frac{1}{N}} \left( a_k^{(1)} + b_{k,i,\infty}(X,Y) a_k^{(2)} + O(b_{k,i,\infty}(X,Y)^2) \right) + O(1/N)
%     \end{eqnarray}
%     where $a_k^{(1)}$ and $a_k^{(2)} > 0$ are constants that only depend on $k$.
%     \label{thm:correlation_boosting}
% \end{theorem}

\begin{figure}[h]
	\begin{center}
	\includegraphics[width=0.9\textwidth]{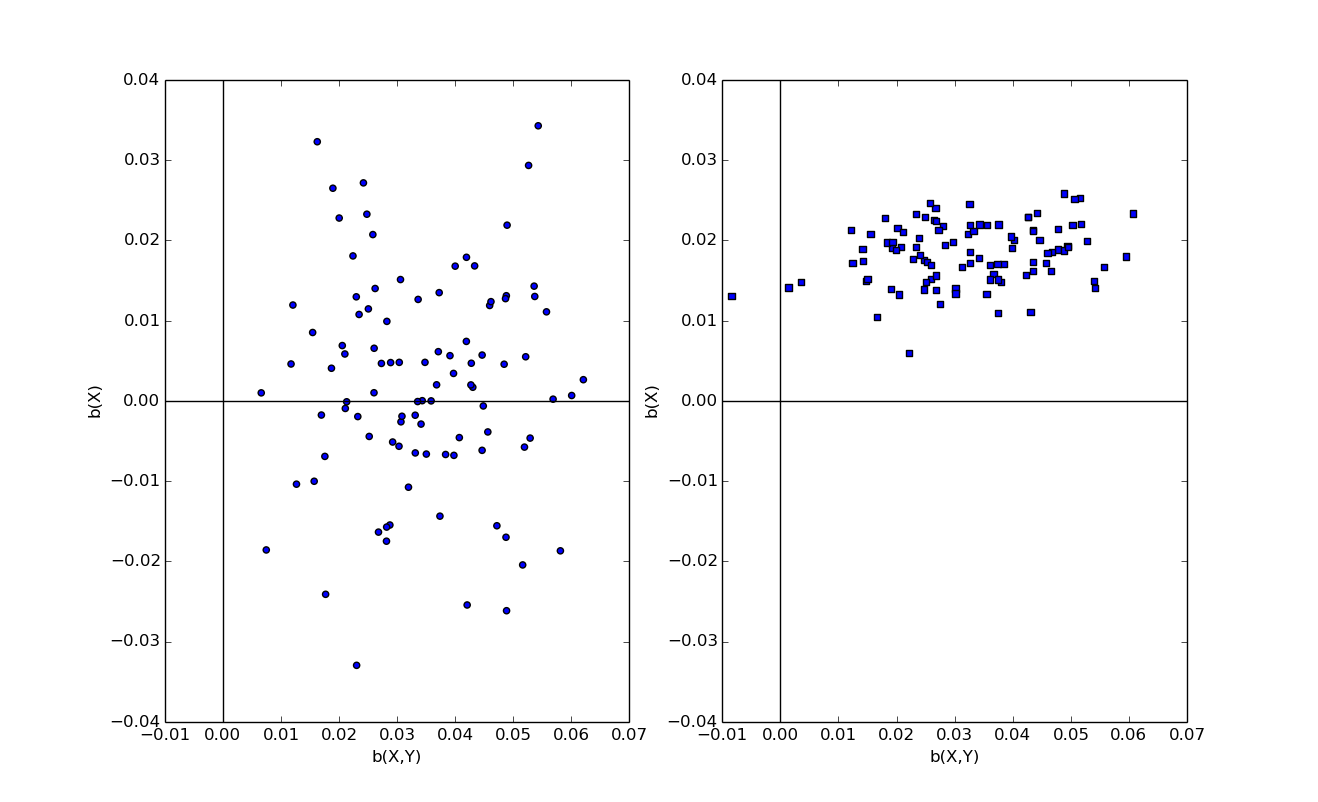}
	\end{center}
	\caption{Scatter plot of the biases $b(X,Y)$ and $b(X)$ to illustrate the correlation boosting effect. Left: 3KL. Right: KSG.}
	\label{fig:correlation_boosting}
\end{figure}

We hypothesize that this correlation {\em boosting} effect is the main reason for the KSG estimator having smaller  mean-square error than the 3KL one.
We  simulate 100 i.i.d.\  samples uniformly from $[0,1]^2$ and map the scatter-plot of the biases $b(X,Y)$ and $b(X)$ in Figure ~\ref{fig:correlation_boosting}, where the boosted correlation for the KSG estimator is visibly significant.

\bigskip\noindent
{\bf New Estimator of Mutual Information.}
 Given the understanding of the correlation boosting effect, it is natural to ask if this can lead to a new estimator that  furthers the  improvement in MSE. This goal is achieved below, where we  discuss potential areas of improvement of the KSG estimator and conclude with our proposal:  Bias Improved KSG (BI-KSG) estimator of mutual information. One of the key differences comes from using  $\ell_2$ norm  to measure $k$-NN distances,  while KSG uses $\ell_{\infty}$ distance. Next,  BI-KSG uses $\log(n_{x,i,2})$ and $\log(n_{y,i,2})$ instead of $\psi(n_{x,1,\infty}+1)$ and $\psi(n_{y,i,\infty} + 1)$, respectively. We briefly discuss the intuitions behind these changes below.
 We begin by noting  that the KSG estimator can be written as:
\begin{eqnarray}
\widehat{I}_{\rm KSG}(X;Y) =  \widehat{H}_{\rm KSG}(X) + \widehat{H}_{\rm KSG}(Y) - \widehat{H}_{\rm KL}(X,Y),
\end{eqnarray}
where $\widehat{H}_{\rm KL}(X;Y)$ is the KL entropy estimator (and already known to be consistent). The marginal entropy estimator is  \begin{eqnarray}
\widehat{H}_{\rm KSG}(X) = \frac{1}{N} \sum_{i=1}^N \left(\, -\psi(n_{x,i,\infty} + 1) + \psi(N) + \log c_{d_x,\infty} + d_x \log \rho_{k,i,\infty}\,\right),
\label{eq:ksg-entropy}
\end{eqnarray}
and we note that this has a form similar to that of the KL entropy estimator, except that $k$ is replaced by $n_{x,i,\infty}+1$, which is sample dependent. Suppose $(X^{(k)}_i, Y^{(k)}_i)$ be the $k$-NN of $(X_i,Y_i)$ with distance $\rho_{k,i,\infty}$, then the ``KSG entropy estimator" in \eqref{eq:ksg-entropy} implicitly assumes that $\rho_{k,i,\infty}$ is {\em both} the $(n_{x,i,\infty}+1)$-NN distance of $X_i$ on $X$-space {\em and} the $(n_{y,i,\infty}+1)$-NN of $Y_i$ on $Y$-space. But since $\ell_{\infty}$-distance is used, $(X^{(k)}_i, Y^{(k)}_i)$ either lies on the $X$-boundary of the hypercube $S_{(X,Y,\rho_{k,i,\infty})} = \left\{\, (x,y): \max\left\{ \|x - X_i\|_{\infty}, \|y - y_i\|_{\infty} \right\} \leq \rho_{k,i,\infty}\,\right\}$, or on the $Y$-boundary of $S_{(X,Y,\rho_{k,i,\infty})}$ (the chance of lying on a corner, and thus on both the boundaries, has zero probability).  If the $k$-NN lies on the $X$-boundary, i.e. $\|X^{(k)}_i - X_i\| = \rho_{k,i,\infty}$ and $\|Y^{(k)}_i - Y_i\|_{\infty} < \rho_{k,i,\infty}$, then $\rho_{k,i,\infty}$ is the $(n_{x,i,\infty}+1)$-NN distance of $X_i$, but {\em not} the $(n_{y,i,\infty}+1)$-NN distance of $Y_i$. Thus, while the estimate of entropy of $X$ is correct, the entropy of $Y$ is over-estimated. Since $\rho_{k,i,\infty}$ is between the $n_{y,i,\infty}$-th and $(n_{y,i,\infty}+1)$-th NN distance,  the  ``KSG entropy estimator" in \eqref{eq:ksg-entropy}  introduces a bias of order $1/n_{y,i,\infty}$. Similarly,  a $1/n_{x,i,\infty}$-bias if $(X^{(k)}_i, Y^{(k)}_i)$ is introduced if the $k$-NN sample lies on the $Y$-boundary.

\begin{figure}[h]
	\begin{center}
	\includegraphics[width=1.0\textwidth]{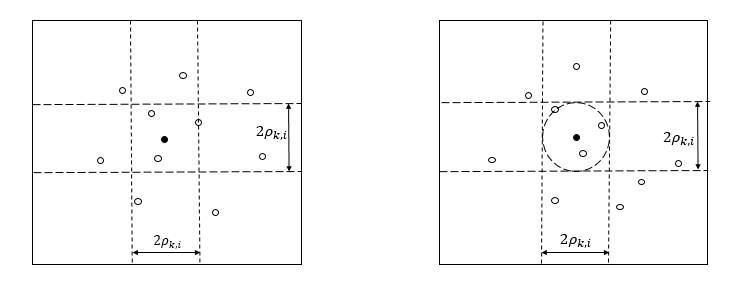}
	\end{center}
	\caption{Illustration of choice of $\rho_{k,i}$ for $k=3$. Left: use $\ell_{\infty}$-distance. Right: use $\ell_2$-distance}
	\label{fig:l_infty_vs_l_2}
\end{figure}

This discussion suggests that we use an $\ell_2$ ball, instead of an $\ell_{\infty}$ ball  to find the $k$-NN. % (See Figure~\ref{fig:l_infty_vs_l_2}). 
This would ensure  that $\rho_{k,i,2}$ is neither the $(n_{x,i,2}+1)$-NN distance of $X_i$ on $X$-space nor the $(n_{y,i,2}+1)$-NN distance of $Y_i$ on $Y$-space. But then, we are unable to directly use the KL estimator for $H(X)$ and $H(Y)$ with this distance. The following theorem sheds some light on this conundrum, with the proof relegated to the appendix.

\begin{theorem}~\label{thm_bino_tr}
Given $(X_i, Y_i) = (x,y)$ such that the density $f$ is twice continuously differentiable at (x,y) and $\rho_{k,i,2} = r < r_N$ for some deterministic sequence of $r_N$ such that $\lim_{N \to \infty} r_N = 0$, the number of neighbors $n_{x,i,2}-k$ is distributed as $\sum_{l=k+1}^{N-1} U_l$, where $U_l$ are i.i.d.\  Bernoulli random variables with mean $p$, and there exists a positive constant $C_1$ such that for sufficiently large $N$.
%\begin{eqnarray}
$r^{-d_x} \left|\, p - f_X(x) c_{{d_x},2} r^{d_x} \,\right| \leq C_1 \, \left(\, r^{2} + r^{d_y} \,\right). $% \label{eq:thm_bino_tr}
%\end{eqnarray}

\end{theorem}

Intuitively, the theorem says that $E[n_{x,i,2}] \approx  N f_X(x) c_{{d_x},2} \rho_{k,i,2}^{d_x}$. This suggests that  we estimate the log of the density ($\log \widehat{f_X}(x)$) by $\log (n_{x,i,2}) - \log N - \log c_{{d_x},2} - d_x \log \rho_{k,i,2}$. The resubstitution estimate of the marginal entropy $H(X)$ is now:
\begin{eqnarray}
\widehat{H}_{\rm BI-KSG}(X) = \frac{1}{N} \sum_{i=1}^N \left(\, - \log (n_{x,i,2}) + \log N + \log c_{{d_x},2} + d_x \log \rho_{k,i,2} \,\right)
\end{eqnarray}
which is different from the KL estimate only via replacing the digamma function by the logarithm. This technique kills the $O(1/n_{x,i,2} + 1/n_{y,i,2})$ bias of the  ``KSG entropy estimator" and leads to the new estimator of mutual information that we christen {\em bias-improved KSG estimator}:
\begin{eqnarray}
    \hI_{\rm BI-KSG} (X;Y) &\equiv & \psi(k) + \log N  + \log \Big(\frac{c_{d_x,2}\; \; c_{d_y,2}}{c_{d_x+d_y,2}}\Big) - \frac1N \sum_{i=1}^N  \log(n_{x,i,2}) + \log (n_{y,i,2}),
    \label{def:BI-KSG_new}
\end{eqnarray}
where $c_{d,2} = \pi^{\frac{d}{2}}/\Gamma(\frac{d}{2}+1)$ be the volume of $d$-dimensional unit $\ell_2$ ball. We show the following result on the theoretical performance of this new estimator, which mimics our result on the KSG estimator.
\begin{theorem}
    The BI-KSG estimator is consistent. The bias of the BI-KSG estimator is $\tO(N^{-\frac{1}{d_x+d_y}})$  and the variance is $\tO(1/N)$. Thus the $\ell_2$ error of the BI-KSG estimator is $\tO(\frac{1}{\sqrt{N}} + N^{-\frac{1}{d_x+d_y}})$.
\end{theorem}
Indeed,
when $N$ gets large, so do $n_{x,i,2}$ and $n_{y,i,2}$, and hence the  KSG and BI-KSG estimators asymptotically perform similarly.  But when $k$ is small and $N$ is moderate and  $X$ and $Y$ are not independent, then $n_{x,i,2}$ and $n_{y,i,2}$ are {\em expected} to be small. In such cases, BI-KSG should outperform KSG. We demonstrate this empirically in Table~\ref{tab:l_2_correlation} where we choose $k=1$ and $X$ and $Y$ are joint Gaussian with mean 0 and covariance $\Sigma =  [1, 0.9; 0.9, 1]$. %$\Sigma = \begin{pmatrix} 1 & 0.9 \\ 0.9 & 1 \end{pmatrix}$.
We can see that all the estimators converge to the ground truth as $N$ goes to infinity, but BI-KSG has the best sample complexity for moderate values of   $N$. Overall,  the empirical gains of correlation boosting are most seen in moderate sample sizes. 
 
  Our current theoretical understanding leads to the same upper bounds on the asymptotic rates of convergence for the KSG and BI-KSG estimators, and fails to explain the  correlation boosting effects. We suspect that the gains of correlation boosting are not in the first order terms in the rates of convergence (of bias and variance) but in the multiplicative constants. A theoretical understanding of these constant terms is an interesting future direction; such an effort has been successfully conducted for entropy estimators based on kernel density estimators \cite{joe1989estimation}. 

\begin{table}[h]
\centering
\begin{tabular}{|c|cccccc|}
 \hline
  N & 100 & 200 & 400 & 800 & 1600 & 3200 \\
 \hline
 3KL & 0.0590 & 0.1025 & 0.0313 & 0.0053 & 0.0097 & 0.0079 \\
 KSG & 0.0240 & 0.0100 & 0.0217 & 0.0024 & 0.0087 & 0.0046 \\
 BI-KSG & \bf 0.0096 & \bf -0.0035 & \bf 0.0133 & \bf -0.0012 & \bf 0.0071 & \bf 0.0032 \\
 \hline
\end{tabular}
\caption{Comparison of bias
 for different mutual information estimators.}
\label{tab:l_2_correlation}
\end{table}

  \section{Multivariate Mutual Information}
  \label{sec:MMI}
\begin{figure}[h]
\begin{minipage}[b]{0.5\linewidth}
    \begin{center}
	\includegraphics[width=0.8\textwidth]{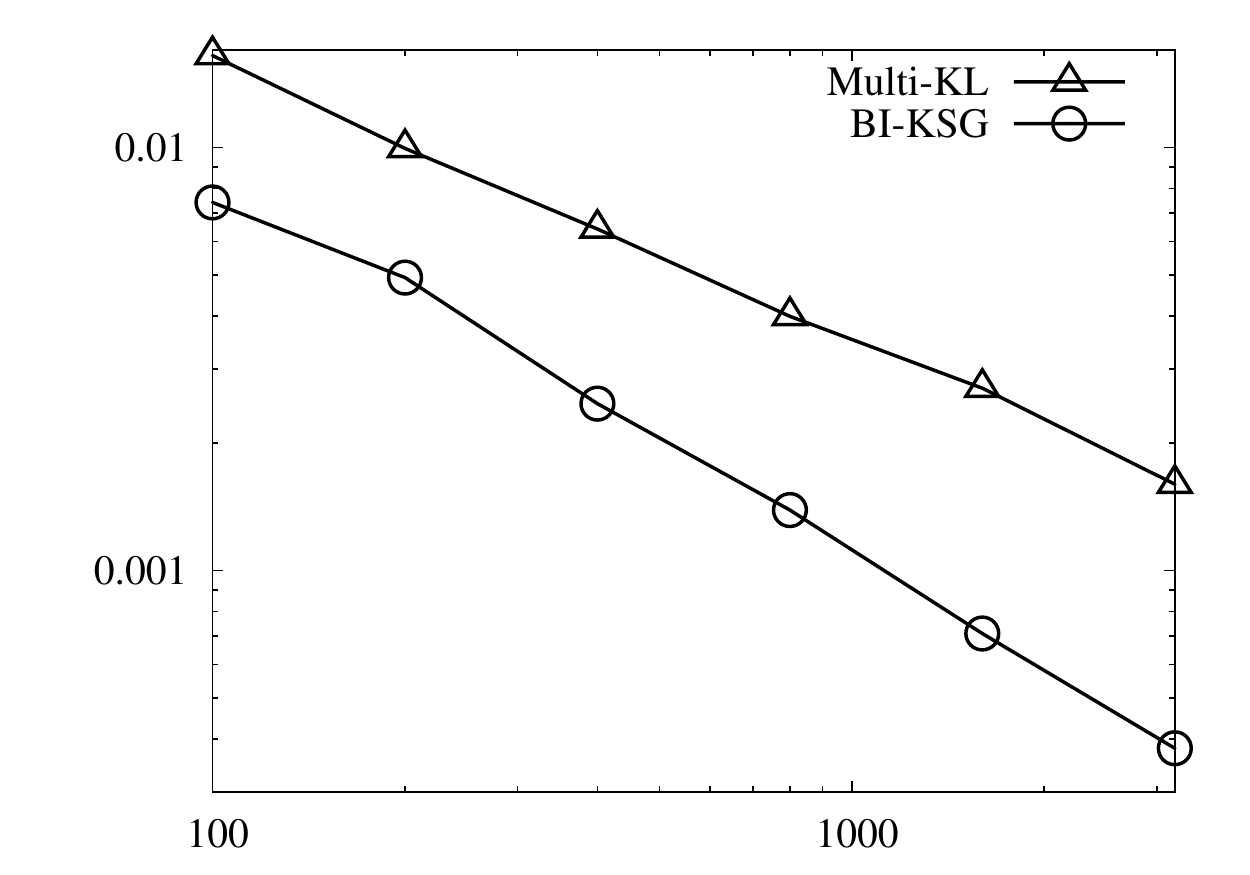}
	\end{center}
	\caption{Plot of MSE with sample size. BI-KSG performs marginally better than  KSG. } \label{fig:mmi}
    \end{minipage} \quad \quad
   \begin{minipage}[b]{0.5\linewidth}
    \vspace{-1.7cm}
    \begin{center}
    \includegraphics[width=0.9\textwidth]{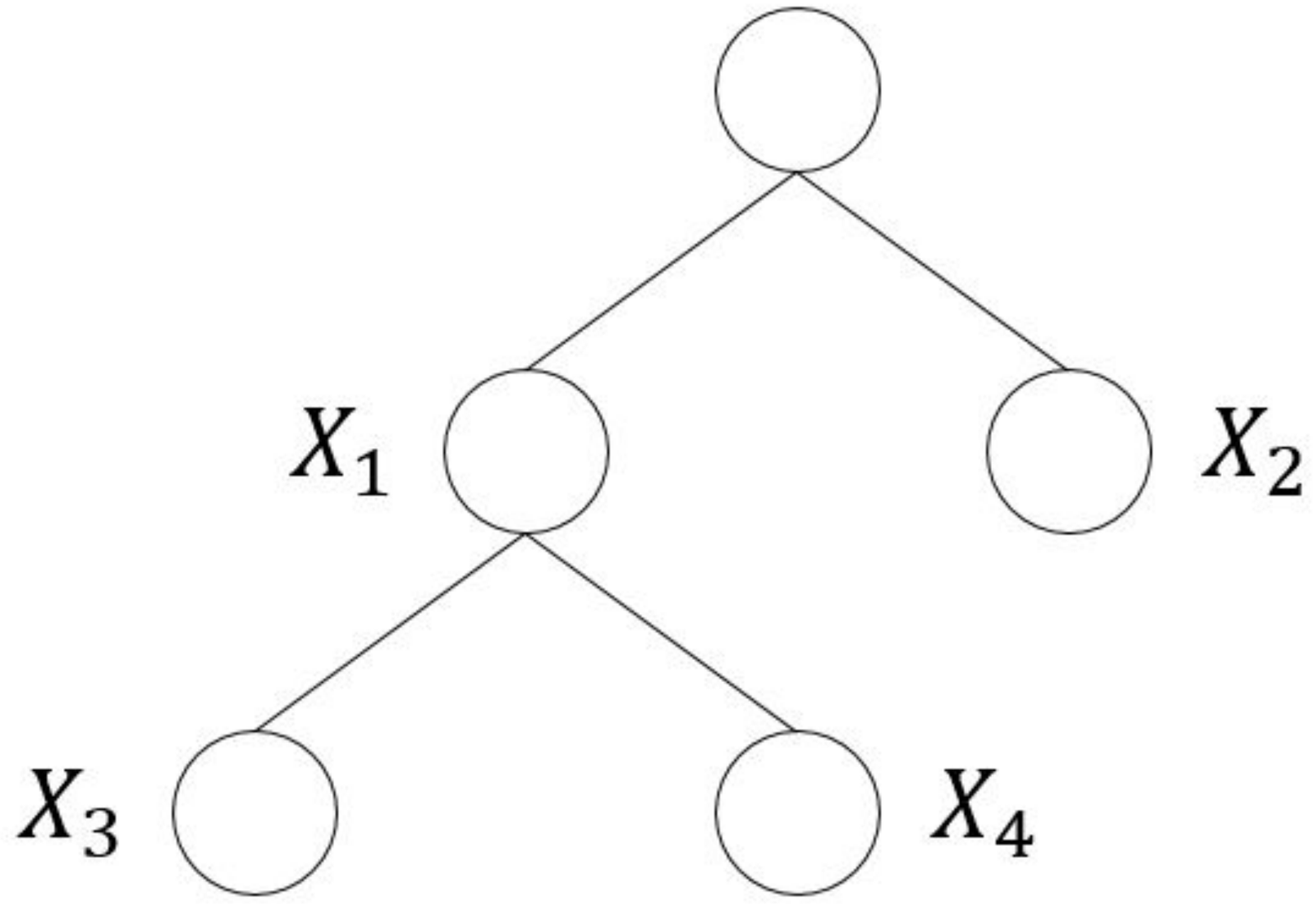}
    \vspace{-0.3cm}
	\end{center}
	\caption{Causal influence on a specific graphical model.  }
    \end{minipage}
	\label{fig:mmi_graph}
\end{figure}

 Generalizations of the standard mutual information that measure the relation {\em among a sequence} of random variables
are routinely used in various applications of machine learning. We discuss two such multivariate versions of mutual information below and show how the correlation boosting ideas from the previous section can be used to construct sample-efficient estimators. The first version is a straightforward generalization and routinely used in unsupervised clustering and correlation extraction, cf.\  \cite{ver2014discovering,ver2014maximally,chanclustering,steeg2015information} for a few recent applications:
\begin{eqnarray}
I(X_1; X_2 ; X_3; \cdots ;X_L) = \sum_{\ell =1}^L H(X_\ell) - H(X_1,X_2,\ldots ,X_L).
\label{eq:mmi_def}
\end{eqnarray}
One natural way to estimate this multivariate mutual information (MMI) is to use the sum and differences of the basic entropy estimators. In particular,  one can use the fixed $k$-NN based KL entropy estimator to estimate MMI from i.i.d.\ samples (we can christen such a method as the $L+1$-KL estimator, generalizing from the 3KL estimator). Alternatively, one can use the correlation boosting ideas of KSG and BI-KSG to construct superior MMI estimators. Generalizing from Equations~\eqref{def:KSG} and~\eqref{def:BI-KSG} we construct the estimators:
\begin{eqnarray}
I_{\rm KSG}(X_1; X_2 ; X_3; \cdots ;X_L) &=&   \psi(k) + \log N - \frac1N \sum_{i=1}^N \sum_{\ell=1}^L \psi(n_{x_{\ell},i,\infty}),  \\
I_{\rm BI-KSG}(X_1; X_2 ; X_3; \cdots ;X_L) &=&   \psi(k) + \log N  + \log \Big(\frac{\prod_{\ell=1}^Lc_{d_\ell,2}}{c_{\sum_{\ell=1}^L d_\ell,2}}\Big) - \frac1N \sum_{i=1}^N \sum_{\ell=1}^L  \log(n_{x_{\ell},i,2}). \nonumber
\end{eqnarray}
Here $d_\ell$ is the dimension of $X_\ell$. The key property we used in constructing these estimators is that the definition of MMI is {\em balanced} with respect to each of the $L$ random variables: for every entropy term with a positive coefficient featuring a random variable $X_\ell$ there is a corresponding entropy term with a negative coefficient featuring the same random variable $X_{\ell}$. From a theoretical perspective, the balance property ensures that the theoretical properties (including consistency) proved in the (pairwise) mutual information setting in Section~\ref{sec:ksg} carry over to this MMI setting as well. From an empirical perspective, we see  that the correlation boosting estimators perform significantly better than the simpler $(L+1)$-KL estimator defined as $\hI_{(L+1)-\textrm{KL}} = \sum_{j=1}^L \hH_{\rm KL}(X_j) - \hH_{\rm KL} (X_1, \dots, X_L)$ in Figure~\ref{fig:mmi} where $N = 100 \sim 3000$ and $L=3$ and the random variables are jointly Gaussian with covariance matrix [1 1/2 1/4; 1/2 1 1/2; 1/4 1/2 1].

As another application of our ideas, we consider a more general form of multivariate mutual information:
\begin{eqnarray}
{\rm MMI}(X_1; X_2 ; X_3; \cdots ;X_L) = \sum_{S\subset \{ 1,\ldots ,L\}} a_{S}\; \cdot \;  H(X_S),
\end{eqnarray}
for some {\em balanced} real valued set function $a_S$, i.e., for every $\ell = 1\ldots L$ we have $\sum_{S \ni \ell \in S} a_S = 0$. Such a metric was posited recently in the context of causal influence measurement on probabilistic graphical models (cf.\ Equation (9) in  \cite{Janzing13}) and widely studied in the information theory community due to its invariance to scaling (cf.\ \cite{DBLP:journals/corr/MakkuvaW16} for a recent example). The definition in Equation~\eqref{eq:mmi_def} is a special case with the set function equal to 1 for singletons and -1 for the whole set and 0, otherwise (and can be viewed as arising out of a graphical model with a single latent variable). Such MMI can be estimated from samples using the correlation boosting ideas presented in this paper: we briefly describe the procedure in the context of an example (which can be viewed as a certain causal strength measurement \cite{Janzing13} with respect to the graphical model in Figure~\ref{fig:mmi_graph}): ${\rm MMI}(X_1,X_2,X_3,X_4) = H(X_1 X_3) + H(X_1 X_4) - H(X_1) + H(X_2) - H(X_1 X_2 X_3 X_4)$. For each sample $i$, we first find the $k$-NN distance $\rho$ in the joint space (of four random variables) and use it estimate the joint entropy using the KL estimator. Then we use this distance to calculate the number of neighbors in each of the other subset of random variables (in this case two pairwise ones ($(X_1X_3)$ and $(X_1X_4)$), and two marginal ones ($X_1$ and $X_2$), and use these to estimate the corresponding entropies. The balanced nature of the metric ensures that the actual distance $\rho$ is {\em precisely canceled out} when all the entropy estimators are put together. In this case, the full estimator (in the spirit of the KSG estimator) is the following, and directly inherits the theoretical and empirical flavor of results from those in Section~\ref{sec:ksg}:
\begin{eqnarray*}
{\rm MMI}_{\rm KSG} &=&   \psi(k) + \log N - \frac1N \sum_{i=1}^N \psi(n_{x_{1}x_{3},i,\infty}) + \psi(n_{x_{1}x_{4},i,\infty}) - \psi(n_{x_{1},i,\infty}) + \psi(n_{x_{2},i,\infty}).
\end{eqnarray*}

\section{Related Work}
\label{sec:priorwork}

The basic estimation question studied in this paper takes a different hue depending on whether the underlying distribution is discrete or continuous. In the discrete setting,  significant understanding of the minimax rate-optimal estimation of functionals, including entropy and mutual information, of an unknown probability mass function is attained via recent works \cite{paninski2004estimating,paninski2003estimation,valiant2011estimating,jiao2015minimax,wu2014minimax}.
 The continuous setting is significantly different, bringing to fore the interplay of geometry of the Euclidean space as well as the role of dimensionality of the domain in terms of estimating the information theoretic quantities; this setting is the focus of this paper.  This fundamental question has been of longstanding interest in the theoretical statistics community where it is a canonical question of estimating a functional of the (unknown) density  \cite{birge1995estimation} but also in the machine learning \cite{gao2014efficient,kinney2014equitability}, information theory \cite{wang2009divergence,paninski2004estimating,wu2014minimax,wang2005divergence},  and theoretical computer science \cite{valiant2011estimating,batu2000testing,acharya2014estimating} communities. The popularity of mutual information and other information theoretic quantities comes from  their wide use as  basic features in several downstream applications \cite{fleuret2004fast,battiti1994using,wells1996multi,turney2002thumbs}.

A conceptually straightforward way to estimate the  differential entropy and mutual information is to use a kernel density estimator (KDE) \cite{silverman1986density,joe1989estimation,ahmad1976nonparametric,eggermont1999best,paninski2008undersmoothed,hall1993estimation}:   the densities $f_{X,Y}, f_X, f_Y$ are separately estimated from samples and the estimated densities are then used to calculate the entropy and mutual information  via the resubstitution estimator. A typical approach to avoid overfitting is to conduct data splitting (DS): split the samples and use one part for KDE and the other for the resubstitution.

In some cases, the parametric rate of convergence of $\sqrt{N}$ of $\ell_2$ error  is achieved: of particular interest is the result of \cite{joe1989estimation} where the
parametric rate is achieved for differential entropy estimation via KDE of density followed by the resubstitution estimator when the dimension is no more than 6. Numerical evidence %(cf.\ Figure \ref{fig:ub_vs_lb} and Section~\ref{sec:comparing})
 suggests the hypothesis that the lower bounds derived in Theorem~\ref{thm:minimax_lower_bound} below could perhaps be improved when the dimension is more than 4 and estimators constrained to only use fixed $k$-NN distances. Under certain very strong conditions on the density class (that are relevant in certain applications on graphical model selection \cite{liu2012exponential}), exponential rate of convergence can be demonstrated \cite{singh2014exponential,singh2014generalized}.
Recent works \cite{krishnamurthy2014nonparametric,kandasamy2015nonparametric} have studied the performance of the leave-one-out (LOO) approach where all but the sample of resubstitution are used for KDE, involving  techniques such as von Mises expansion methods.

Alternative methods involve estimation of the
entropies using spacings \cite{vasicek1976test,van1992estimating},  the Edgeworth expansion \cite{van2005edgeworth}, and convex optimization \cite{nguyen2010estimating}.  Among the $k$-NN methods, there are two variants: either $k$ is chosen to grow with the sample size $N$ or $k$ is fixed. There is a large literature on the former, where the classical result is the possibility of consistent estimation of the density from $k$-NN distances \cite{loftsgaarden1965nonparametric,fix1951discriminatory}, including recent sharper consistency characterizations \cite{biau2011weighted,kpotufe2011pruning}. Several works have applied this basic insight towards the estimation of the specific case of information theoretic quantities \cite{dasgupta2014optimal,vincent2003locally} and extensions to generalized NN graphs~\cite{pal2010estimation}. For fixed $k$-NN methods, apart from the works referred to in the main text,  detailed experimental comparisons are in \cite{perez2009estimation} and local Gaussian approaches studied in  \cite{gao2014efficient,gao2015estimating,lombardi2016nonparametric} bringing together local likelihood density estimation methods  \cite{loader1996local,hjort1996locally} with $k$-NN driven choices of kernel bandwidth.

In this paper we have considered the smoothness of the class of pdfs studied via  bounded Hessians. In nonparametric estimation, a standard feature is to consider whole families of smooth pdfs as defined by how the differences of derivatives relate to the differences of the samples \cite{bickel2009springer}. Of specific interest is the H\"{o}lder family:
$\Sigma(s, C)$, i.e., for any tuple $r = (r_1, \dots, r_d)$, define $D^r = \frac{\partial^{r_1 + \dots + r_d}}{\partial x_1^{r_1} \dots \partial x_d^{r_d}}$. Then for any $r$ such that $\sum_j r_j =  \lfloor s \rfloor$, where $\lfloor s \rfloor$ is the largest integer smaller than $s$, we have:
                   \begin{eqnarray}
                    \|D^r f(x) - D^r f(y)\| \leq C \|x-y\|^{s - \sum_j r_j} \;.
                    \label{eq:Holder_class}
                    \end{eqnarray}
                    for any $x, y$.
The rate of convergence of various nonparametric estimators depends on the parameter $s$ of the H\"{o}lder family under consideration, cf.\ \cite{krishnamurthy2014nonparametric,kandasamy2015nonparametric} for recent work on convergence rate characterization of information theoretic quantities via KDE and resubstitution estimators as a function of the smoothness parameter $s$.  It is natural to ask if such smoothness considerations could lead to a refined understanding of the rates of convergence of the fixed $k$-NN KL and KSG estimators studied here.

In the context of the KL estimator, the only place where smoothness plays a critical role is in the statement (and proof) of
Lemma~\ref{lemma_p_xr}. For small enough $r$, defining  $P(x,r)(u) = \Pr \{\|X-x\| < r\}$, we seek to understand how this probability can be approximated by the density at $x$. With bounded Hessian norms, Lemma~\ref{lemma_p_xr} asserts the following:
\begin{eqnarray}
\left|\, P(x,r) - f(x) c_d r^d \,\right| &\leq& C r^{d+2}\;,
\end{eqnarray}
which is crucial in deriving the rate of convergence upper bounds on the KL estimator. A fairly straightforward calculation shows that this condition does not change even if we allow for smoother class of families of pdfs, as defined via the H\"{o}lder class -- we conclude that refined rates of convergence for fixed $k$-NN estimators do not materialize by standard approaches such as the
H\"{o}lder class.

% \textbf{!`!`!`TODOS!!!} KSG is only functionals of $n_{x,i}$, $n_{y,i}$. Talk about how to use this to domain adaption problems.

Although our analysis technique is inspired by that of \cite{tsybakov1996root}, while generalizing it to higher dimensions,
several subtle differences emerge and \cite{tsybakov1996root} does not imply our result even for $d=1$: 
hence, we complement the understanding of $K$NN methods even for univariate random variables.
For example, random variables with strictly positive densities over a bounded support are covered by our analysis,
whereas random variables with unbounded support that are smooth everywhere are covered by the results of \cite{tsybakov1996root}.
The reason is that non-smooth boundaries are not handled in \cite{tsybakov1996root} and
densities approaching zero are not handled by our analysis. We believe it is possible to extend our analysis to have a theorem that includes both types of random variables, which is an interesting future research direction.

%\textbf{!`!`!`TODOS!!!} cite Al Hero and Samworth

In this paper, $k$ is assumed to be a finite constant, and we do not keep track of how the convergence rate depends on $k$.
Analyses on fixed $\rho$ estimators \cite{sricharan2013ensemble}, where instead of fixing $k$ and using the distance $\rho_k$, one
fixes the distance $\rho$ and uses the number of neighbors $k_\rho$ within that distance,
we expect the convergence rate of the variance to be independent of $k$, and the convergence rate of bias to be of order
$O( (k/N)^{1/d})$.
Recently, the idea of using an ensemble of $k$NN entropy estimators to achieve a faster convergence rate has been introduced in \cite{sricharan2013ensemble,moon2014ensemble}. If the first-order terms in the convergence rate is known, then it is possible to achieve the parametric rate of $O(1/N)$ by taking a (weighted) linear combination of multiple estimators with varying $k$, whose weight depends on the convergence rate.
Applying this idea together with KSG (and KL) estimators have the potential to improve the convergence rate we provide in this paper.
The main challenge is in identifying the exact constants in the first-order terms in the convergence rate, and estimating it from samples in the case when
the constant depends on the underlying distribution.

%\clearpage
%%%%%%%%%%%%%%%%%%%%%%%%%%%%%%%%%%%%%%%
%%   Appendix
%%%%%%%%%%%%%%%%%%%%%%%%%%%%%%%%%%%%%%%
%\section*{Appendix}
%\appendix

% bias of joint entropy estimation
\section{Proof of Theorem \ref{thm:bias_KL}}
\label{sec:proof_bias_KL}

We follow closely the proof from \cite{tsybakov1996root} of the $\sqrt{N}$-consistency of the  one-dimensional entropy estimator introduced in \cite{KL87}.
 It was proved in \cite{tsybakov1996root} that
the KL entropy estimator %in Equation \eqref{eq:KL2}
achieves $\sqrt{N}$-consistency in mean, i.e. $\E[\widehat{H}(X)]-H(X) =O(1/\sqrt{N})$,
and in variance,
i.e. $\E[(\widehat{H}(X)-\E[\widehat{H}(X)])^2] =O(1/N)$, under the assumption that
the $X$ is a one-dimensional random variable and the estimator uses only the nearest neighbor distance with $k=1$.
In the process of proving our main result, we prove a generalization of this rate of convergence of the KL entropy estimator for general $d$-dimensional space and for a general $k$.
Also notice that our proof works for any choice of $\ell_p$ distance for $1 \leq p \leq \infty$, so we will drop the subscribe $p$ in the proof of Theorem \ref{thm:bias_KL} and Theorem \ref{thm:variance_KL}.

Firstly, we notice that $\xi_{k,i}(X)$ are identically distributed and $\xi_{k,i} = 0$ if $\rho_{k,i} > a_N$, so we have:
\begin{eqnarray}
\E \left[\, \widehat{H}_{\rm tKL}(X) \,\right] &=& \frac{1}{N} \sum_{i=1}^N \E \left[\, \xi_{k,i}(X) \,\right] = \E \left[\, \xi_{k,1}(X)\,\right] \,\notag\\
&=& \E \left[\, \xi_{k,1}(X) \cdot \mathbb{I}\{\rho_{k,i} \leq a_N\} \,\right]. \label{eq:expectedentropy}
\end{eqnarray}
We introduce the following notations. Let $b_N = e^{-\psi(k)} N c_d a_N^d = e^{-\psi(k)} c_d (\log N)^{1+\delta}$ and for every $u>0$ define
\begin{eqnarray}
r_N(u) = \left(\, \frac{u e^{\psi(k)}}{c_d N} \,\right)^{1/d} \;,
\end{eqnarray}
such that $r_N(e^{\xi_{k,1}(X)}) = \rho_{k,1}$ for $\rho_{k,1}\leq a_N$ and $r_N(b_N) = a_N$. It is easy to check that $\frac{d r_N(u)}{du} = \frac{r_N(u)}{ud}$.
These definitions provides a new representation of the expectation in \eqref{eq:expectedentropy} using a change of variables $u=r_N^{-1}(\rho_{k,1})$:
\begin{eqnarray}
\E \left[\, \xi_{k,1}(X) \cdot \mathbb{I}\{\rho_{k,1} \leq a_N\} \, \right] = \E \left[\,\log u \cdot \mathbb{I}\{u \leq b_N\}\,\right] =\int \left( \int_0^{b_N} \log u \, dF_{N,x}(u)\right) f(x) dx \;,
\end{eqnarray}
where
we define the following distribution:
\begin{eqnarray}
F_{N,x}(u) &=& \Pr \left(\, e^{\xi_{k,1}(X)} < u \,\big|\, X_1 = x \,\right) = \Pr \left(\, \rho_{k,1} < r_N(u) \,\big|\, X_1 = x \,\right) \;.\label{eq:F_Nx}
\end{eqnarray}
Similar change of variables holds for the actual entropy as follows.

\begin{lemma}\label{lemma_h_x}
\begin{eqnarray}
H(X) &=& \int \left( \int_0^{\infty} \log u dF_x(u)\right) f(x) dx \;,
\end{eqnarray}
where
\begin{eqnarray}
F_x(u) &=& 1 - \exp\{ - u e^{\psi(k)} f(x)\} \sum_{j=0}^{k-1} \frac{(u e^{\psi(k)} f(x))^j}{j!}\;.\label{eq:F_x}
\end{eqnarray}
\end{lemma}
This allows us to decompose the bias into three terms, each of which can be bounded separately.
\begin{eqnarray}
\Big| \E \left[\, \widehat{H}_{\rm tKL}(X) \,\right] - H(X) \Big| &=& \Big| \E \left[\, \xi_{k,1}(X) \cdot \mathbb{I}\{\rho_{k,1} \leq a_N\}  - H(X) \, \right] \Big|  \\
&\leq& \int \left(I_1(x) + I_2(x) + I_3(x)\right) f(x) dx \;,
\end{eqnarray}
where
\begin{eqnarray}
I_1(x) &=& \left| \int_{b_N}^{\infty} \log u \, dF_x(u) \right| \;,\notag\\
I_2(x) &=& \left| \int_0^1 \log u \, dF_{N,x}(u) - \int_0^1 \log u \, dF_x(u)\right| \;,\notag\\
I_3(x) &=& \left| \int_1^{b_N} \log u \, dF_{N,x}(u) - \int_1^{b_N} \log u \, dF_x(u)\right| \;.
\end{eqnarray}
 We will bound the three terms separately.
The main idea is that $I_1(x)$ is small when $b_N$ is sufficiently large, and $I_2(x)$ and $I_3(x)$ are small when $f_x(u)$ and $f_{N,x}(u)$ are close.

$I_1(x)$: We upper bound the tail probability that the $k$-NN distance is truncated. By plugging in the cdf~\eqref{eq:F_x} of $F_x(u)$, we get:
\begin{eqnarray}
I_1(x) &=& \left| \int_{b_N}^{\infty} \log u dF_x(u) \right| \notag\\
&=& \left| \int_{b_N}^{\infty} \log u \frac{d F_x(u)}{du} du \right|\,\notag\\
&=& \frac{1}{(k-1)!} \left| \int_{b_N}^{\infty} (\log u) \, e^{\psi(k)} f(x) \exp\{-u e^{\psi(k)} f(x)\} (u e^{\psi(k)} f(x))^{k-1} du\right| \,\notag\\
&=& \frac{1}{(k-1)!} \left| \int_{b_N e^{\psi(k)} f(x)}^{\infty} \left( \log t - \psi(k) - \log f(x) \right) e^{-t} t^{k-1} dt \right| \;, \label{eq:ub_i1}
\end{eqnarray}
where the third equality is from Equation \eqref{eq:f_x} and the last equality comes from changing of variable $t = ue^{\psi(k)}f(x)$.  Now we consider two cases:
\begin{enumerate}
    \item $b_N e^{\psi(k)} f(x) < 1$. Then \eqref{eq:ub_i1} is upper bounded by:
    \begin{eqnarray}
        &&\frac{1}{(k-1)!} \left| \int_{b_N e^{\psi(k)} f(x)}^{\infty} \left( \log t - \psi(k) - \log f(x) \right) e^{-t} t^{k-1} dt \right| \,\notag\\
        &\leq& \frac{1}{(k-1)!} \left(\, \left| \int_{b_N e^{\psi(k)} f(x)}^{\infty} \log t \, e^{-t} t^{k-1} dt \right| + |\psi(k) + \log f(x)| \, \left| \int_{b_N e^{\psi(k)} f(x)}^{\infty} e^{-t} t^{k-1} dt \right| \,\right) \,\notag\\
        &\leq& \frac{1}{(k-1)!} \left(\, \int_0^{\infty} |\log t| \, e^{-t} t^{k-1} dt + |\psi(k) + \log f(x)| \int_0^{\infty} e^{-t} t^{k-1} dt \,\right)\,\notag\\
        &\leq& C_1(1+|\psi(k) + \log f(x)|).
    \end{eqnarray}
    where $C_1 = \max \left\{\, \frac{1}{(k-1)!}\int_0^{\infty} |\log t| \, e^{-t} t^{k-1} dt, \frac{1}{(k-1)!} \int_0^{\infty} e^{-t} t^{k-1} dt \,\right\}$.
    \item $b_N e^{\psi(k)} f(x) \geq 1$. Then \eqref{eq:ub_i1} is upper bounded by:
    \begin{eqnarray}
        && \frac{1}{(k-1)!} \left| \int_{b_N e^{\psi(k)} f(x)}^{\infty} \left(\, \log t - \psi(k) - \log f(x) \,\right)  \, e^{-t} t^{k-1} dt \right| \,\notag\\
        &\leq& \frac{1}{(k-1)!} \left(\, \left| \int_{b_N e^{\psi(k)} f(x)}^{\infty} \log t \, e^{-t} t^{k-1} dt \right| + |\psi(k) + \log f(x)| \, \left| \int_{b_N e^{\psi(k)} f(x)}^{\infty} e^{-t} t^{k-1} dt \right| \,\right) \,\notag\\
        &\leq& C_2 (1 + |\psi(k) + \log f(x)|) \int_{b_N e^{\psi(k)} f(x)}^{\infty} e^{-t/2} dt \,\notag\\
        &\leq& 2 C_2 (1 + |\psi(k) + \log f(x)|) \exp\{-b_N e^{\psi(k)} f(x)\} \;,
    \end{eqnarray}
    where $C_2$ is a constant satisfying $ \log t  \cdot t^{k-1}/(k-1)! < C_2 e^{t/2}$ and $t^{k-1}/(k-1)! < C_2 e^{t/2}$ for all $t > 1$.
\end{enumerate}
Now combining the two cases, $I_1(x)$ is bounded by:
\begin{eqnarray}
I_1(x) &\leq&  (1+|\psi(k) + \log f(x)|) \left(\, C_1 \, \mathbb{I}\{b_N e^{\psi(k)}f(x) < 1\} + 2 C_2 \exp\{-b_N e^{\psi(k)} f(x)\} \,\right) \,\notag \\
&\leq&  C_3 (1 + |\log f(x)|)\, \exp\{-b_N e^{\psi(k)} f(x) \} \;,\label{ub_i1}
\end{eqnarray}
where we use the fact that $\mathbb{I}\{b_N e^{\psi(k)}f(x) < 1\} \leq \exp\{1 - b_N e^{\psi(k)}f(x)\}$. Here $C_3 = (C_1 e + 2C_2) (1+|\psi(k)|)$.
\\

$I_2(x)$: $I_2(x)$ can be bounded by:
\begin{eqnarray}
I_2(x) &=& \left| \int_0^1 \log u \, dF_{N,x}(u) - \int_0^1 \log u \, dF_x(u)\right| \leq \int_0^1 |\log u| \, \left| f_{N,x}(u) - f_x(u)\right| du
\end{eqnarray}
where $f_{N,x}(u)$ and $f_x(u)$ are the corresponding pdfs of $F_{N,x}(u)$ and $F_x(u)$, respectively.  Here we partition the support into two parts. Let
\begin{eqnarray}
S_1 &=& \{x: \|H_f(y)\|_2 < C_d, \forall y \in B(x, a_N) \} \,\notag\\
S_2 &=& \{x: \|H_f(y)\|_2 \geq C_d \textrm{ for some } y \in B(x, a_N) \} = S_1^C
\end{eqnarray}
From Assumption~\ref{assumption_rate_ent}, $(d-1)$-dimensional Hausdorff measure of the set that $\|H_f(x)\| \geq C_d$ is finite, so the Lebegue measure of $S_2$ is bounded by $2 a_N C_e$ for sufficiently large $N$. For points in $S_1$ and $S_2$, In the following lemma we give an upper bound for the difference of $f_{N,x}(u)$ and $f_x(u)$ for $x$ in $S_1$ and $S_2$ separately.

\begin{lemma}
    \label{lem:f_nx}
    Under the Assumption~\ref{assumption_rate_ent}, for any $x \in S_1$,
    \begin{eqnarray}
        \left|  f_{N,x}(u) - f_x(u) \right| \leq C_4 \left(\, N^{-2/d} + N^{-1} \,\right)\;,
        \label{eq:upper_bound_f_nx}
    \end{eqnarray}
    for $u \leq 1$. For $x \in S_2$, we have
    \begin{eqnarray}
        \left|  f_{N,x}(u) - f_x(u) \right| \leq C_4\;,
        \label{eq:upper_bound_f_nx_s2}
    \end{eqnarray}
    for $u \leq 1$.
    %If we replace Assumption ~\ref{assumption_rate_ent}$(d.1)$ by $(d.2)$, we have,
    %\begin{eqnarray}
     %   \left|  f_{N,x}(u) - f_x(u) \right| \leq C_4 \left(\, N^{-s/d} + N^{-1} \,\right)\;.
      %  \label{eq:upper_bound_f_nx_Holder}
    %\end{eqnarray}
    %for $u \leq 1$.
\end{lemma}

Using Lemma \ref{lem:f_nx} and the fact that $\int_0^1 |\log u| du = 1$, $I_2(x)$ is upper bounded by:
\begin{eqnarray}
I_2(x) &\leq& C_4 (N^{-2/d} + N^{-1}) \int_0^1 |\log u| du \leq C_4 (N^{-2/d} + N^{-1}) \;.\label{ub_i2}
\end{eqnarray}
for $x \in S_1$ and
\begin{eqnarray}
I_2(x) &\leq& C_4  \int_0^1 |\log u| du \leq C_4  \;.\label{ub_i2_s2}
\end{eqnarray}
for $x \in S_2$.
%or $I_2(x) \leq C_4 (N^{-s/d} + N^{-1})$ under Assumption \ref{assumption_rate_ent}.$(d.2)$.

$I_3(x)$: $I_3(x)$ can bounded by:
\begin{eqnarray}
I_3(x) &=& \left| \int_1^{b_N} \log u \, dF_{N,x}(u) - \int_1^{b_N} \log u \, dF_x(u)\right| \,\notag\\
&=& \left| \int_1^{b_N} \frac{1}{u} (1-F_{N,x}(u)) du - \int_1^{b_N} \frac{1}{u} (1-F_{x}(u)) du \right| \,\notag\\
&\leq& \int_1^{b_N} \frac{1}{u} |F_{N,x}(u) - F_x(u)| du\;.
\end{eqnarray}
 In the following lemma we give an upper bound for the difference of $F_{N,x}(u)$ and $F_x(u)$ for $x $ in $S_1$ and $S_2$ separately.

\begin{lemma}
    \label{lem:F_nx}
    Under the Assumption~\ref{assumption_rate_ent},
    \begin{eqnarray}
        \left| F_{N,x}(u) - F_x(u) \right| \leq C_5 \left(\, u^{1+2/d} N^{-2/d} +u^2/N \,\right)\;, \label{eq:upper_bound_F_nx}
    \end{eqnarray}
    for $x \in S_1$ and
    \begin{eqnarray}
        \left| F_{N,x}(u) - F_x(u) \right| \leq C_5 \left(\, u +u^2/N \,\right)\;, \label{eq:upper_bound_F_nx_s2}
    \end{eqnarray}
    for $x \in S_2$.
%    for $u \leq b_N = e^{-\psi(k)}c_d (\log N)^{1+\delta}$.%(\log N)^{1+\delta}$. %If we replace Assumption ~\ref{assumption_rate_ent}$(d.1)$ by $(d.2)$, we have,
    %\begin{eqnarray}
     %   \left|  F_{N,x}(u) - F_x(u) \right| \leq C_5 \left(\, u^{1+s/d} N^{-s/d} + u^2/N\,\right)\;.
      %  \label{eq:upper_bound_F_nx_Holder}
    %\end{eqnarray}
\end{lemma}

Using Lemma \ref{lem:F_nx}, $I_3(x)$ is upper bounded by:
\begin{eqnarray}
I_3(x) &\leq& C_5 \int_1^{b_N} \Big( (u/N)^{2/d} + u/N \Big) du \,\notag\\
&\leq& C_5 \left(\,  b_N^{1+2/d} N^{-2/d} + b^2_N N^{-1} \,\right)\;,\label{ub_i3}
\end{eqnarray}
for $x \in S_1$ and
\begin{eqnarray}
I_3(x) &\leq& C_5 \int_1^{b_N} \Big( 1 + u/N \Big) du \,\notag\\
&\leq& C_5 \left(\,  b_N + b^2_N N^{-1} \,\right)\;,\label{ub_i3_s2}
\end{eqnarray}
for $x \in S_2$
%or $I_3(x) \leq C_5 \left(\,  b_N^{1+s/d} N^{-s/d} + b^2_N/N \,\right)$ under Assumption \ref{assumption_rate_ent}.$(d.2)$.

Combining the upper bounds of $I_1(x)$, $I_2(x)$ and $I_3(x)$ and defining $C_6 = \max\{C_3, C_4, C_5\}$, the bias is bounded by:
\begin{eqnarray}
&& \E\left[\widehat{H}_{\rm tKL}(X) \,\right] - H(X) \,\notag\\
&\leq& \int \left(I_1(x) + I_2(x) + I_3(x)\right) f(x) dx \,\notag\\
&\leq& \int I_1(x) f(x) dx + \int_{S_1} \left(\, I_2(x) + I_3(x) \,\right) f(x) dx + \int_{S_2} \left(\, I_2(x) + I_3(x) \,\right) f(x) dx \,\notag\\
&\leq& C_6 \int \left( |1+\log f(x)| \exp\{-b_N e^{\psi(k)} f(x)\} \right) f(x) dx \,\notag\\
&& + \int_{S_1} \left(\, N^{-2/d} + N^{-1} + b_N^{1+2/d}N^{-2/d} + b_N^2 N^{-1} \right) f(x) dx + \int_{S_2} \left(\, 1 + b_N + b_N^2 N^{-1} \right) f(x) dx \,\notag\\
&\leq& C_6 \Big(\, \int f(x) \exp\{-b_N e^{\psi(k)} f(x)\} + \int f(x) |\log f(x)| \exp\{-b_N e^{\psi(k)} f(x)\} \,\notag\\
&&+ b_N^{1+2/d}N^{-2/d} + b_N^2 N^{-1} + b_N \left(\, \int_{S_2} f(x) dx\,\right) \,\Big) \;,
\end{eqnarray}
%under assumption $(d.2)$ (or replace $s$ by $2$ under assumption $(d.1)$).
By Assumption~\ref{assumption_rate_ent}.$(c)$, the first term is bounded by: $\int f(x) \exp\{-b_N e^{\psi(k)} f(x)\} \leq C_d e^{-b_N C_0}$. The second term is bounded by H\"{o}lder  inequality as:
\begin{eqnarray}
&&\int f(x) |\log f(x)| \exp\{-b_N e^{\psi(k)} f(x)\} \,\notag\\
&\leq& \left(\, \int f(x) (\log f(x))^{1+\gamma} dx \,\right)^{1/(1+\gamma)} \, \left(\, \int f(x) \exp\{-\frac{1+\gamma}{\gamma} b_N e^{\psi(k)} f(x)\} dx \,\right)^{\gamma/(1+\gamma)} \,\notag\\
&\leq& C_b^{1/(1+\gamma)} \left(\, C_d e^{-\frac{1+\gamma}{\gamma} C_0 b_N} \,\right)^{\gamma/(1+\gamma)}.
\end{eqnarray}
By choosing $b_N = e^{-\psi(k)}c_d (\log N)^{1+\delta}$ for some $\delta > 0$, we know that $e^{-C_0 b_N}$ decays faster than $N^{-\alpha}$ for any $\alpha$.

The last term is bounded by $b_N (\, \int_{S_2} f(x) dx\,) \leq b_N C_a m(S_2) \leq 2b_N a_N C_a C_e$, where $m(S_2)$ is the Lebesgue measure of $S_2$. Recall that we choose $a_N = ((\log N)^{1+\delta}/N)^{1/d}$, so the proof is complete.

% F_{n,x} integrates to H(X)
\subsection{Proof of Lemma \ref{lemma_h_x}}

Since $F_x(u)$ is a continuous CDF, the corresponding pdf is given by:
\begin{eqnarray}
f_x(u) = \frac{d F_x(u)}{du} &=& -\exp\{-u e^{\psi(k)} f(x)\} \sum_{j=1}^{k-1} \frac{(u e^{\psi(k)} f(x))^{j-1}}{(j-1)!} e^{\psi(k)} f(x) \,\notag\\
&+& e^{\psi(k)} f(x) \exp\{-u e^{\psi(k)} f(x)\} \sum_{j=0}^{k-1} \frac{(u e^{\psi(k)} f(x))^j}{j!} \,\notag\\
&=& \frac{1}{(k-1)!} e^{\psi(k)} f(x) \exp\{-u e^{\psi(k)} f(x)\} (u e^{\psi(k)} f(x))^{k-1}. \label{eq:f_x}
\end{eqnarray}
Therefore,
\begin{eqnarray}
\int_0^{\infty} \log u dF_x(u) &=& \frac{1}{(k-1)!} \int_0^{\infty} \log u \, e^{\psi(k)} f(x) \exp\{-u e^{\psi(k)} f(x)\} (u e^{\psi(k)} f(x))^{k-1} du \,\notag\\
&=& \frac{1}{(k-1)!} \int_{0}^{\infty} \left(\log t - \psi(k) - \log f(x)\right) e^{-t} t^{k-1} dt \,\notag\\
&=& \psi(k) - \psi(k) - \log f(x) \,\notag\\
&=& -\log f(x) \;,
\end{eqnarray}
where the third to last equation comes from change of variable $t = u e^{\psi(k)} f(x)$. The penultimate equation comes from the fact that $\psi(k) = \frac{1}{(k-1)!} \int_{0}^{\infty} (\log t) \, t^{k-1} e^{-t} dt$ and $1 = \frac{1}{(k-1)!} \int_{0}^{\infty} t^{k-1} e^{-t} dt$. Therefore,
\begin{eqnarray}
\int \left( \int_0^{\infty} \log u dF_x(u)\right) f(x) dx = \int \left( -\log f(x) \right) f(x) dx = H(X).
\end{eqnarray}

% Bound on difference of $f_{n,x}$ and $f_x$
\subsection{Proof of Lemma ~\ref{lem:f_nx}}

Recall that
\begin{eqnarray}
f_x(u) = \frac{1}{(k-1)!} e^{\psi(k)} f(x) \exp\{-u e^{\psi(k)} f(x)\} (u e^{\psi(k)} f(x))^{k-1}.
\end{eqnarray}
Notice that $r_N(u)$ is the $k^{\rm th}$ order statistic of $\big\{\, \|X_1 - x\|, \|X_2 - x\|, \dots, \|X_{N-1} - x\| \,\big\}$. Therefore the density  $f_{N,x}(u)$ is given by:
\begin{eqnarray}
f_{N,x}(u) &=& f_{r_N(u)} \, \frac{dr_N(u)}{du} \,\notag\\
&=& \frac{(N-1)!}{(k-1)!(N-k-1)!} \left( P(x, r_N(u)) \right)^{k-1} \left( 1-P(x, r_N(u)) \right)^{N-k-1} \frac{dP(x, r_N(u))}{dr_N(u)} \frac{dr_N(u)}{du} \,\notag\\
&=& \frac{(N-1)!}{(k-1)!(N-k-1)!} \left( P(x, r_N(u)) \right)^{k-1} \left( 1-P(x, r_N(u)) \right)^{N-k-1}\frac{dP(x, r_N(u))}{du}.
\end{eqnarray}
Here $P(x,r)(u) = \Pr \{\|X-x\| < r\} = \int_{ t\in B(x,r)} f(t) dt$. Since $f$ is twice differentiable and $r_N(u)$ goes to 0 as $N$ goes to infinity,  we can use $f(z) Vol(B(z,r_N(u))$ to estimate $P(z,r_N(u))$. The following lemma bounds the error of this estimation for $x \in S_1$ and $x \in S_2$ separately:

\begin{lemma}~\label{lemma_p_xr}
Under Assumption~\ref{assumption_rate_ent}, there exists a constant $C$ such that for sufficiently small $r$, we have
\begin{eqnarray}
\left|\, P(x,r) - f(x) c_d r^d \,\right| &\leq& C r^{d+2}\;,
\end{eqnarray}
and
\begin{eqnarray}
\left|\, \frac{dP(x,r)}{dr} - f(x) d c_d r^{d-1} \,\right| &\leq& C r^{d+1}\;,
\end{eqnarray}
for $x \in S_1$. For $x \in S_2$, we have
\begin{eqnarray}
\left|\, P(x,r) - f(x) c_d r^d \,\right| &\leq& C r^{d}\;,
\end{eqnarray}
and
\begin{eqnarray}
\left|\, \frac{dP(x,r)}{dr} - f(x) d c_d r^{d-1} \,\right| &\leq& C r^{d-1}\;,
\end{eqnarray}

%Under Assumption \ref{assumption_rate_ent}.$(d.2)$, for sufficiently small $r$, we have
%\begin{eqnarray}
%\left|\, P(x,r) - f(x) c_d r^d \,\right| &\leq& C r^{d+s}\;,
%\end{eqnarray}
%and
%\begin{eqnarray}
%\left|\, \frac{dP(x,r)}{dr} - f(x) d c_d r^{d-1} \,\right| &\leq& C r^{d+s-1}\;.
%\end{eqnarray}
\end{lemma}

%From now on, we will focus on the case that $f \in \Sigma(s, C_d)$, i.e. Assumption ~\ref{assumption_rate_ent}. We can replace $s$ by $2$ for $f$ satisfying $\|H_f(x)\| \leq C_d$.
Using Lemma~\ref{lemma_p_xr} and substituting  $r = r_N(u) = (u e^{\psi(k)}/(c_d N))^{1/d}$, we have:
\begin{eqnarray}
\left| P(x, r_N(u)) - \frac{u e^{\psi(k)} f(x)}{N}\right| = \left| P(x, r_N(u)) - f(x) c_d (r_N(u))^d\right| \leq C_1 (r_N(u))^{d+2}. \label{eq:ub_volume}
\end{eqnarray}
for $x \in S_1$. Similarly, $\left| \frac{d}{du} P(x, r_N(u)) - \frac{e^{\psi(k)} f(x)}{N}\right|$ can be bounded by:
\begin{eqnarray}
&&\left| \frac{d}{du} P(x, r_N(u)) - \frac{e^{\psi(k)} f(x)}{N}\right| \,\notag\\
&=& \frac{dr_N(u)}{du} \left| \frac{d}{dr_N(u)} P(x, r_N(u)) - (\frac{dr_N(u)}{du})^{-1}\frac{e^{\psi(k)} f(x)}{N}\right| \,\notag\\
&=& \frac{r_N(u)}{u\,d} \left| \frac{d}{dr_N(u)} P(x, r_N(u)) -  f(x) d c_d (r_N(u))^{d-1}\right| \,\notag\\
&\leq& \frac{C_1 (r_N(u))^{d+2}}{u}. \label{eq:ub_surface}
\end{eqnarray}
for $x \in S_1$. Analogously we have $\left| P(x, r_N(u)) - \frac{u e^{\psi(k)} f(x)}{N}\right| \leq C_1 (r_N(u))^d$ and $\left| \frac{d}{du} P(x, r_N(u)) - \frac{e^{\psi(k)} f(x)}{N}\right| \leq C_1 (r_N(u))^d/u$ for $x \in S_2$. Now we can write the difference of $f_{N,x}(u)$ and $f_x(u)$ via two terms:
\begin{eqnarray}
|f_{N,x}(u) - f_x(u)| &\leq& |f_{N,x}(u) - f^{(1)}_{N,x}(u)| \,+\, |f^{(1)}_{N,x}(u) - f_x(u)|\;, \label{eq:tri1}
\end{eqnarray}
where $f^{(1)}_{N,x}(u)$ defined as:
\begin{eqnarray}
f^{(1)}_{N,x}(u) &=& \frac{(N-1)!}{(k-1)!(N-k-1)!} (\frac{u e^{\psi(k)}f(x)}{N})^{k-1} (1 - \frac{u e^{\psi(k)} f(x)}{N})^{N-k-1} \frac{e^{\psi(k)} f(x)}{N}.
\end{eqnarray}
Consider the function $g(p) = \frac{(N-1)!}{(k-1)!(N-k-1)!} p^{k-1} (1-p)^{N-k-1}$ for $p \in (0,1)$. By basic calculus, we can see that $g(p) \leq C_2 N$ and $|g'(p)| \leq C_3 N^2$ for $p \in (0,1)$. Therefore, the first term in~\eqref{eq:tri1} can be bounded as:
\begin{eqnarray}
&&|f_{N,x}(u) - f^{(1)}_{N,x}(u)| \,\notag\\
&=& \left|\, g\left( P(x,r_N(u)) \right) \frac{dP(x,r_N(u))}{du} - g \left( \frac{ue^{\psi(k)}f(x)}{N}\right) \frac{e^{\psi(k)}}f(x){N} \,\right| \,\notag\\
&\leq& g\left( P(x,r_N(u)) \right) \left|\, \frac{dP(x,r_N(u))}{du} - \frac{e^{\psi(k)}f(x)}{N} \,\right| + \left|\, g\left( P(x,r_N(u)) \right) - g\left( \frac{ue^{\psi(k)}f(x)}{N} \right) \,\right| \frac{e^{\psi(k)}f(x)}{N} \,\notag\\
&\leq& g\left( P(x,r_N(u)) \right) \left|\, \frac{dP(x,r_N(u))}{du} - \frac{e^{\psi(k)}f(x)}{N} \,\right| + \max_{p\in(0,1)} |g'(p)| \, \left|\,  P(x,r_N(u)) - \frac{ue^{\psi(k)}f(x)}{N} \,\right| \frac{e^{\psi(k)}f(x)}{N} \,\notag\\
&\leq& C_1 C_2 N  (r_N(u))^{d+2}/u + C_1 C_3 N^2 (r_N(u))^{d+2} \frac{e^{\psi(k)}f(x)}{N} \,\notag\\
&\leq& C_4 \frac{u^{1+2/d}}{N^{2/d}}(1+\frac{1}{u}) \,\notag\\
&\leq& C_4 N^{-2/d} \;,\label{eq:ub1_fnx}
\end{eqnarray}
for $u \leq 1$ and $x \in S_1$. Here $C_4 = \max\{C_1 C_2 \left(\, e^{\psi(k)} C_a\,\right)^{1+2/d} , C_1 C_3 \left(\, e^{\psi(k)} C_a\,\right)^{2+2/d}\}$, where $C_a = \sup_{x} f(x)$ by Assumption~\ref{assumption_rate_ent}.$(a)$. Similarly, we have $|f_{N,x}(u) - f_{N,x}^{(1)}(u)| \leq C_4$ for $u \leq 1$ and $x \in S_2$. For the second term, we denote $q = ue^{\psi(k)}f(x)$ for short. Then the second term in~\eqref{eq:tri1} can be bounded as:
\begin{eqnarray}
&& |f^{(1)}_{N,x}(u) - f_x(u)| \,\notag\\
&=& \frac{1}{u} \, \left|\, \frac{(N-1)!}{(k-1)!(N-k-1)!} \Big(\frac{q}{N}\Big)^k \Big(1-\frac{q}{N}\Big)^{N-k-1} - \frac{1}{(k-1)!}q^k e^{-q} \,\right| \,\notag\\
&=& \frac{k}{u} \, \left|\,{N-1 \choose k} \Big(\frac{q}{N}\Big)^k \Big(1-\frac{q}{N}\Big)^{N-k-1} - \frac{q^k e^{-q}}{k!}\,\right|.
\end{eqnarray}
Notice that the difference inside the absolute value is just the difference of $P(X=k)$ under Bino$(N-1,q/N)$ and Poisson$(q)$. The difference is bounded by:

\begin{lemma}\label{lemma:bino_vs_poisson}
For $q < C \sqrt{N}$, we have:
\begin{eqnarray}
\left|\, {N-1 \choose k} \Big(\frac{q}{N}\Big)^k \Big(1-\frac{q}{N}\Big)^{N-k-1} - \frac{q^k e^{-q}}{k!}\,\right| \;\; \leq\;\; C_5 q^{k+2} e^{-q}N^{-1}  \;,
\end{eqnarray}
for some $C_5 > 0$.
\end{lemma}

Therefore, by lemma ~\ref{lemma:bino_vs_poisson}, we have:
\begin{eqnarray}
|f^{(1)}_{N,x}(u) - f_x(u)| &\leq& C_5 \, \frac{k q^{k+2} e^{-q}}{u N} \;\leq\; C_5 \, \frac{k ({e^{\psi(k)} f(x)})^{k+2} u^{k+1} }{N} \; \leq \; C_6 N^{-1} \;,\label{eq:ub2_fnx}
\end{eqnarray}
for $u \leq 1$, here $C_6 = C_5 k (e^{\psi(k)} C_a)^{k+1}$. Therefore, combining \eqref{eq:ub1_fnx} and ~\eqref{eq:ub2_fnx}, we have the desired statement.

% Bound on difference of $F_{n,x}$ and $F_x$
\subsection{Proof of Lemma ~\ref{lem:F_nx}}

Recall that
\begin{eqnarray}
F_x(u) = 1 - \exp\{-u e^{\psi(k)} f(x)\} \sum_{j=0}^{k-1} \frac{u e^{\psi(k)} f(x))^{j}}{j!}.
\end{eqnarray}
The cdf $F_{N,x}(u) = \Pr \left(\,\rho_{k,i} < r_N(u)|X_i = x\,\right)$ is just the probability that at least $k$ samples are inside the ball $B(x,r_N(u))$ and hence
\begin{eqnarray}
F_{N,x}(u) &=& 1 - \sum_{j=0}^{k-1} \frac{(N-1)!}{j!(N-j-1)!} \left( P(x, r_N(u)) \right)^j \left( 1-P(x, r_N(u)) \right)^{N-j-1} .
\end{eqnarray}
So we have:
\begin{eqnarray}
&&|F_{N,x}(u) - F_x(u)| \,\notag\\
&=& \left|\, \sum_{j=0}^{k-1} \frac{(N-1)!}{j!(N-j-1)!} \left( P(x, r_N(u)) \right)^j \left( 1-P(x, r_N(u)) \right)^{N-j-1} - \exp\{-u e^{\psi(k)} f(x)\} \sum_{j=0}^{k-1} \frac{u e^{\psi(k)} f(x))^{j}}{j!}\,\right| \,\notag\\
&\leq& \sum_{j=0}^{k-1} \frac{1}{j!}\left|\, \frac{(N-1)!}{(N-j-1)!} \left( P(x, r_N(u)) \right)^j \left( 1-P(x, r_N(u)) \right)^{N-j-1} - \exp\{-u e^{\psi(k)} f(x)\} (u e^{\psi(k)} f(x))^j\,\right|.
\end{eqnarray}
Let
\begin{eqnarray}
h_{N,x,j}(u) = \frac{(N-1)!}{j!(N-j-1)!} \left( P(x, r_N(u)) \right)^j \left( 1-P(x, r_N(u)) \right)^{N-j-1} \;,
\end{eqnarray}
and
\begin{eqnarray}
h_{x,j}(u) = \frac{1}{j!} \exp\{-u e^{\psi(k)} f(x)\} (u e^{\psi(k)} f(x))^j.
\end{eqnarray}
Consider
\begin{eqnarray}
h^{(1)}_{N,x,j}(u) = \frac{(N-1)!}{j!(N-j-1)!} \left( \frac{ue^{\psi(k)}f(x)}{N} \right)^j \left( 1-\frac{ue^{\psi(k)}f(x)}{N} \right)^{N-j-1}.
\end{eqnarray}
We will bound $|h_{N,x,j}(u) - h_{x,j}(u)|$ by $|h_{N,x,j}(u) - h_{N,x,j}^{(1)}(u)| + |h_{N,x,j}^{(1)}(u) - h_{x,j}(u)|$. For the first term, consider function $g_j(p) = \frac{(N-1)!}{j!(N-j-1)!} p^j (1-p)^{N-j-1}$. It is easy to see that $|g'_j(p)| \leq C_1 N$ for any $p \in (0,1)$. Therefore, by Lemma~\ref{lemma_p_xr}, we obtain:
\begin{eqnarray}
|h_{N,x,j}(u) - h_{N,x,j}^{(1)}(u)| &=& \left|\, g(P(x, r_N(u))) - g(\frac{ue^{\psi(k)}f(x)}{N})\,\right| \,\notag\\
&\leq& \max_{p \in (0,1)} |g'(p)| \,\left|\, P(x, r_N(u))) - \frac{ue^{\psi(k)}f(x)}{N}\,\right| \,\notag\\
&\leq& C_1 N (r_N(u))^{d+2} \,\notag\\
&\leq& C_2 u^{1+2/d} N^{-2/d} \;,\label{eq:ub1_Fnx}
\end{eqnarray}
for $x \in S_1$ and $|h_{N,x,j}(u) - h_{N,x,j}^{(1)}(u)|  \leq C_2 u$ for $x \in S_2$, where $C_2 = \max\{ C_1 (e^{\psi(k)} C_a)^{1+2/d}, C_1 e^{\psi(k)} C_a\}$.
%Similar as the proof in Lemma~\ref{lem:f_nx}, we focus on the case that $f$ satisfies Assumption~\ref{assumption_rate_ent} and let $s = 2$ if $f$ satisfies Assumption~\ref{assumption_rate_ent}.
For the second term, let $q = ue^{\psi(k)}f(x)$, and using a  similar analysis as~\eqref{eq:ub2_fnx}, we obtain:
\begin{eqnarray}
|h_{N,x,j}^{(1)}(u) - h_{x,j}(u)| &=& \left|\,{N-1 \choose j} (\frac{q}{N})^j (1-\frac{q}{N})^{N-j-1} - \frac{q^j e^{-q}}{j!}\,\right| \,\notag\\
&\leq& C_3 \frac{q^{j+2} e^{-q}}{N}.  \label{eq:ub2_Fnx}
\end{eqnarray}
Combine~\eqref{eq:ub1_Fnx} and ~\eqref{eq:ub2_Fnx}, and we obtain:
\begin{eqnarray}
|F_{N,x}(u) - F_x(u)| &\leq & \sum_{j=0}^{k-1} |h_{N,x,j}(u) - h_{x,j}(u)| \,\notag\\
&\leq& \sum_{j=0}^{k-1} \left(\, |h_{N,x,j}(u) - h_{N,x,j}^{(1)}(u)| + |h_{N,x,j}^{(1)}(u) - h_{x,j}(u)| \,\right) \,\notag\\
&\leq& k C_2 u^{1+2/d} N^{-2/d} + C_3 \sum_{j=0}^{k-1} \frac{q^{j+2} e^{-q}}{N} \,\notag\\
&\leq& k C_2 u^{1+2/d}(N)^{-2/d} + (k-1)! C_3 q^2/N \,\notag\\
&\leq& k C_2 u^{1+2/d}(N)^{-2/d} + (k-1)! C_3 (e^{\psi(k)} C_a)^2 u^2/N \;,
\end{eqnarray}
for $x \in S_1$. Here we used the fact that $\sum_{j=1}^{k-1} q^j e^{-q} \leq (k-1)!\sum_{j=1}^{k-1} \frac{q^j e^{-q}}{(k-1)!} \leq (k-1)!$. Analogously, we have $|F_{N,x}(u) - F_x(u)| \leq kC_2 u + (k-1)!C_3(e^{\psi(k)}C_a)^2 u^2/N$ for $x \in S_2$. Therefore, we have the desired statement by $C_5 = \max\{kC_2, (k-1)! C_3 (e^{\psi(k)} C_a)^2\}$.

\subsection{Proof of Lemma ~\ref{lemma_p_xr}}
We will prove the lemma for $x \in S_1$ and $x \in S_2$ separately. For $x \in S_1$,  we have $\|H_f(x)\| \leq C_d$ for every $y \in B(x,r)$ as long as $r \leq a_N$. Hence, there exists a $y=at+(1-a)x$ for some $a\in[0,1]$ such that
% \begin{eqnarray}
% \left|\, P(x,r) - f(x) c_d r^d\,\right| &=& \left|\, \int_{t \in B(x,r)} \left(\, f(t) - f(x) \,\right) dt \,\right| \,\notag\\
% &=& \left|\, \int_{t \in B(x,r)} \left(\, f(x) + \left(\, \nabla f(x) \,\right)^T (t-x) + (t-x)^T H_f(x) (t-x) + o(\|t-x\|^2) - f(x) \,\right) dt \,\right| \,\notag\\
% &=& \left|\, \int_{t \in B(x,r)} \left(\, (t-x)^T H_f(x) (t-x) + o(\|t-x\|^2)  \,\right) dt \,\right| \,\notag\\
% &\leq& C_d \int_{t \in B(x,r)}  \|t-x\|^2 dt \,\notag\\
% &\leq& C_d Vol(B(x,r)) \cdot d \cdot r^2 \leq C_1 r^{d+2} \;,
% \end{eqnarray}
\begin{eqnarray}
\left|\, P(x,r) - f(x) c_d r^d\,\right| &=& \left|\, \int_{t \in B(x,r)} \left(\, f(t) - f(x) \,\right) dt \,\right| \,\notag\\
&=& \left|\, \int_{t \in B(x,r)} \left(\, f(x) + \left(\, \nabla f(x) \,\right)^T (t-x) + (t-x)^T H_f(y) (t-x)  - f(x) \,\right) dt \,\right| \,\notag\\
&=& \left|\, \int_{t \in B(x,r)} \left(\, (t-x)^T H_f(y) (t-x)   \,\right) dt \,\right| \,\notag\\
&\leq& C_d \int_{t \in B(x,r)}  \|t-x\|^2 dt \,\notag\\
&\leq& C_d Vol(B(x,r)) \cdot d \cdot r^2 \leq C_1 r^{d+2} \;,
\end{eqnarray}
where $Vol(B(x,r))$ is the volume of $B(x,r)$. $\|t-x\|^2 \leq d \cdot r^2$ for all $t \in B(x,r)$ (here $B(x,r)$ can be any $p$-norm ball with $1 \leq p \leq \infty$). For the second part, Let $S(B(x,r))$ be the surface of $B(x,r)$. Consider $m^{d-1}$ be the Lebesgue measure on $\mathbb{R}^{d-1}$, so $m^{d-1}\left(\, S(B(x,r)) \,\right) = d c_d r^{d-1}$. Similarly we have:
\begin{eqnarray}
\left|\, \frac{d P(x,r)}{dr} - f(x) d c_d r^{d-1}\,\right| &=& \left|\, \int_{t \in S(B(x,r))} \left(\, f(t) - f(x) \,\right) d m^{d-1}(t) \,\right| \,\notag\\
&\leq& C_d \int_{t \in S(B(x,r))}  \|t-x\|^2 d m^{d-1}(t) \,\notag\\
&\leq& C_2 r^{d+1}.
\end{eqnarray}

For $x \in S_2$, we simply bound the difference by:
\begin{eqnarray}
\left|\, P(x,r) - f(x) c_d r^d \,\right| \leq f(x) c_d r^d \leq C_a c_d r^d
\end{eqnarray}
and
\begin{eqnarray}
\left|\, \frac{d P(x,r)}{dr} - f(x) d c_d r^{d-1}\,\right| \leq f(x) d c_d r^{d-1} \leq C_a d c_d r^{d-1}
\end{eqnarray}
since $f(x) \leq C_a$ by Assumption~\ref{assumption_rate_ent}.(a).
%Under Assumption~\ref{assumption_rate_ent}.$(d.2)$, $f \in \Sigma(s, C_d)$, namely $|f(x) - f(y)| \leq \|x-y\|^s$. Therefore we have:
%\begin{eqnarray}
%\left|\, P(x,r) - f(x) c_d r^d\,\right| &=& \left|\, \int_{t \in B(x,r)} \left(\, f(t) - f(x) \,\right) dt \,\right| \,\notag\\
%&\leq& \int_{t \in B(x,r)} |f(t) - f(x)| dt \,\notag\\
%&\leq& \int_{t \in B(x,r)} \|t-x\|^s dt \,\notag\\
%&\leq& C_d d^{s/2} r^s \cdot Vol(B(x,r)) \leq C_1 r^{d+s}.
%\end{eqnarray}
%Similarly, we obtain that $\left|\, \frac{d P(x,r)}{dr} - f(x) d c_d r^{d-1}\,\right| \leq C_2 r^{d+s-1}.$

\subsection{Proof of Lemma ~\ref{lemma:bino_vs_poisson}}
We will prove that:
\begin{eqnarray}
\left|\, \log \left(\, {N \choose k} \Big(\frac{q}{N} \Big)^k \Big(1-\frac{q}{N} \Big)^{N-k} \,\right) - \log \left(\, \frac{q^k e^{-q}}{k!} \,\right)\,\right| \; \leq \; Cq^2/N. \label{eq:diff}
\end{eqnarray}
Then for sufficiently small $q$ such that $\exp\{Cq^2/N\} \leq 2Cq^2/N$, we obtain our desired statement by the fact that $|x-y| \leq |\log x - \log y| \cdot \frac{y}{2}$ for small enough $|\log x - \log y|$. Using Stirling's formula: $\log(N!) = N \log N - N + \frac{1}{2} \log(2\pi N) + O(1/N)$, the difference~\eqref{eq:diff} is given by:
\begin{eqnarray}
&& \left|\, \log \left(\, {N \choose k} \Big(\frac{q}{N}\Big)^k \Big(1-\frac{q}{N}\Big)^{N-k} \,\right) - \log \left(\, \frac{q^k e^{-q}}{k!} \,\right)\,\right| \,\notag\\
&=& \left|\, \log N! - \log (N-k)! - \log k! + k\log q + (N-k) \log (N-q) - N \log N - k \log q + q + \log (k!) \,\right|\,\notag\\
&=& \left|\, \log N! - \log (N-k)!  + (N-k)\log(N-q) - N \log N + q \,\right| \,\notag\\
&\leq&\big|\, N \log N - N + \frac{1}{2} \log(2\pi N) - (N-k) \log(N-k) + (N-k) \,\notag\\
&&\,- \frac{1}{2} \log(2\pi (N-k)) +  (N-k)\log(N-q) - N \log N + q \,\big| + C/N\,\notag\\
&=&\left|\, -k + \frac{1}{2} \log \frac{N}{N-k} + (N-k) \log \frac{N-q}{N-k} + q \,\right| + C/N \,\notag\\
&=&\left|\, -k+q + (N-k) \left(\, \frac{k-q}{N-k} - \frac{(k-q)^2}{2(N-k)^2} + O(\frac{(k-q)^3}{(N-k)^3}))\,\right) \,\right| + C/N \,\notag\\
&\leq& \frac{(k-q)^2}{2(N-k)} + Cq^3/N^2 + C/N \,\notag\\
&\leq& Cq^2/N \;,
\end{eqnarray}
where we used the assumption that $q<C\sqrt{N}$ for sufficiently small constant $C>0$.

% variance of joint entropy estimation
\section{Proof of Theorem \ref{thm:variance_KL}}

Recall that $\widehat{H}_{\rm tKL}(X) = \frac{1}{N} \sum_{i=1}^N \xi_{k,i}(X)$ and $\xi_{k,i}(X)$ are identically distributed, therefore, we obtain
\begin{eqnarray}
\V \left[\, \widehat{H}_{\rm tKL}(X) \,\right] &=& \frac{1}{N^2} \sum_{i=1}^N \V \left[\, \xi_{k,i}(X) \,\right] + \frac{1}{N^2} \sum_{i=1}^N \sum_{j \neq i} \Cov \left[\, \xi_{k,i}(X), \xi_{k,j}(X)\,\right] \,\notag\\
&=& \frac{1}{N} \V \left[\, \xi_{k,1}(X) \,\right] + \frac{N(N-1)}{N^2}  \Cov \left[\, \xi_{k,1}(X), \xi_{k,2}(X)\,\right] \,\notag\\
&\leq& \frac{1}{N} \V \left[\, \xi_{k,1}(X) \,\right] +  \Cov \left[\, \xi_{k,1}(X), \xi_{k,2}(X)\,\right].
\end{eqnarray}

We claim the following two lemmas:

\begin{lemma}
\label{lem_var_xi}
Under the Assumption \ref{assumption_rate_ent},
\begin{eqnarray}
{\rm Var} \left[\, \xi_{k,1}(X) \,\right] = O\left(\, (\log \log N)^2 \,\right)\;,
\end{eqnarray}
\end{lemma}

\begin{lemma}
\label{lem_cov_xi}
Under the Assumption \ref{assumption_rate_ent},
\begin{eqnarray}
\Cov \left[\, \xi_{k,1}(X), \xi_{k,2}(X) \,\right] = O\left(\, (\log \log N)^2 (\log N)^{(2k+2)(1+\delta)} \, N^{-1} \,\right)\;.
\end{eqnarray}
\end{lemma}
Combining the two lemmas, we obtain the desired statement.

\subsection{Proof of Lemma \ref{lem_var_xi}}
Recall that in the proof of Theorem \ref{thm:bias_KL}, we have defined the following distributions:
\begin{eqnarray}
F_{N,x}(u) = \Pr \left(\, e^{\xi_{k,1}(X)} < u \,|\, X_1 = x \,\right) = \Pr \left(\, \rho_{k,1} < r_N(u) \,|\, X_1 = x\,\right);
\end{eqnarray}
\begin{eqnarray}
F_x(u) = 1 - \exp\{-u e^{\psi(k)} f(x)\} \sum_{j=0}^{k-1} \frac{(ue^{\psi(k)f(x)})^j}{j!},
\end{eqnarray}
and their corresponding pdfs $f_{N,x}(u)$ and $f_x(u)$. The variance of $\xi_{k,i}(X)$ is upper bounded by:
\begin{eqnarray}
\V \left[\, \xi_{k,1}(X) \,\right] &\leq& \E \left[\, \left(\, \xi_{k,1}(X) \,\right)^2 \,\right] = \E_X \left[\, \E \left[\,\left(\, \xi_{k,1}(X) \,\right)^2 \,|\, X_1 = x\,\right] \,\right] \,\notag\\
&=& \int \left(\, \int_0^{b_N} (\log u)^2 f_{N,x}(u) du\,\right) f(x) dx \,\notag\\
&=& \int \left(\, \int_0^1 (\log u)^2 f_{N,x}(u) du + \int_1^{b_N} (\log u)^2 f_{N,x}(u) du\,\right) f(x) dx. \label{eq:var_xi}
\end{eqnarray}

For $u < 1$, Lemma~\ref{lem:f_nx} told us that there exists some $C_1 > 0$ such that
\begin{eqnarray}
|f_{N,x}(u) - f_x(u)| \leq C_1 \;,
\end{eqnarray}
holds for $x \in S_1$ or $S_2$. The closed form of $f_x(u)$ is given by:
\begin{eqnarray}
f_x(u) = \frac{1}{(k-1)!} e^{\psi(k)} f(x) \exp\{-u e^{\psi(k)} f(x)\} (ue^{\psi(k)} f(x))^{k-1}.
\end{eqnarray}
Since $\frac{1}{(k-1)!} t^{k-1} e^{-t} < 1$ for all $t > 0$, we know that $f_x(u) < e^{\psi(k)}f(x)$. Therefore, $f_{N,x}(u) \leq C_1 + e^{\psi(k)} f(x)$ by triangle inequality. Therefore,
\begin{eqnarray}
\int_0^1 \log^2(u) f_{N,x}(u) du \leq \left(\, C_1 + e^{\psi(k)} f(x)\,\right) \int_0^1 (\log u)^2 du =  2 \left(\, C_1 + e^{\psi(k)} f(x)\,\right).
\end{eqnarray}
For $1 \leq u \leq b_N$, we have $(\log u)^2 \leq (\log  b_N)^2 = \log^2((\log N)^{1+\delta}) = (1+\delta)^2 (\log \log N)^2$ for sufficiently large $N$. Therefore,
\begin{eqnarray}
\int_1^{b_N} (\log u)^2 f_{N,x}(u) du \leq (1+\delta)^2 (\log \log N)^2 \int_1^{b_N} f_{N,x}(u) du \leq (1+\delta)^2 (\log \log N)^2 .
\end{eqnarray}
Combine these two results into \eqref{eq:var_xi}, and we obtain:
\begin{eqnarray}
\V \left[\, \xi_{k,1}(X) \,\right] &\leq& \int \left(\, \int_0^1 (\log u)^2 f_{N,x}(u) du + \int_1^{b_N} (\log u)^2 f_{N,x}(u) du\,\right) f(x) dx \,\notag\\
&\leq& \int \left(\, 2 \left(\, C_1 + e^{\psi(k)} f(x)\,\right) + (1+\delta)^2 (\log \log N)^2\,\right) f(x) dx \,\notag\\
&\leq& 2C_1 + 2e^{\psi(k)}C_a + (1+\delta)^2 (\log \log N)^2 \,\notag\\
&=& O\left(\, (\log \log N)^2 \,\right),
\end{eqnarray}
where we used the assumption that $f(x) \leq C_a$.

\subsection{Proof of Lemma \ref{lem_cov_xi}}

The covariance can be rewritten as:
\begin{eqnarray}
&&\Cov \left[\, \xi_{k,1}(X), \xi_{k,2}(X)\,\right] \,\notag\\
&=& \E \left[\, \left(\, \xi_{k,1}(X) - \E\left[\,\xi_{k,1}(X)\,\right]\,\right) \left(\, \xi_{k,2}(X) - \E\left[\,\xi_{k,2}(X)\,\right]\,\right) \,\right] \,\notag\\
&=& \int_{x,y} \E \left[\, \left(\, \xi_{k,1}(X) - \E\left[\,\xi_{k,1}(X)\,\right]\,\right) \left(\, \xi_{k,2}(X) - \E\left[\,\xi_{k,2}(X)\,\right]\,\right) \,|\, X_1 = x, X_2 = y\,\right] f(x) f(y) dx dy \,\notag\\
&=& \int_{\|x-y\| \leq 2 a_N} \E \left[\, \left(\, \xi_{k,1}(X) - \E\left[\,\xi_{k,1}(X)\,\right]\,\right) \left(\, \xi_{k,2}(X) - \E\left[\,\xi_{k,2}(X)\,\right]\,\right) \,|\, X_1 = x, X_2 = y\,\right] f(x) f(y) dx dy \,\notag\\
&+& \int_{\|x-y\| > 2 a_N} \E \left[\, \left(\, \xi_{k,1}(X) - \E\left[\,\xi_{k,1}(X)\,\right]\,\right) \left(\, \xi_{k,2}(X) - \E\left[\,\xi_{k,2}(X)\,\right]\,\right) \,|\, X_1 = x, X_2 = y\,\right] f(x) f(y) dx dy. \label{eq:cov_xi}
\end{eqnarray}
We split the covariance into  two separate cases: If $\|x-y\| \leq 2a_N$, the first term of \eqref{eq:cov_xi} can be bounded by Cauchy-Schwarz inequality as:
\begin{eqnarray}
&& \int_{\|x-y\| \leq 2 a_N} \E \left[\, \left(\, \xi_{k,1}(X) - \E\left[\,\xi_{k,1}(X)\,\right]\,\right) \left(\, \xi_{k,2}(X) - \E\left[\,\xi_{k,2}(X)\,\right]\,\right) \,|\, X_1 = x, X_2 = y\,\right] f(x) f(y) dx dy \,\notag\\
&\leq& \int_{\|x-y\| \leq 2 a_N} \V \left[\, \xi_{k,1}(X) \,|\, X_1 = x, X_2 = y\,\right]^{1/2}  \V \left[\, \xi_{k,2}(X) \,|\, X_1 = x,X_2 = y\,\right]^{1/2} f(x) f(y) dx dy.
\end{eqnarray}
Consider the following CDF:
\begin{eqnarray}
F_{N,x}(u) = \Pr \left(\, e^{\xi_{k,1}(X)} < u \,|\, X_1 = x,X_2 = y \,\right) = \Pr \left(\, \rho_{k,1} < r_N(u) \,|\, X_1 = x,X_2 = y\,\right);
\end{eqnarray}
and the corresponding pdf $f_{N,x,y}(u)$, which is given by order statistic~\cite{mcbook}:
\begin{eqnarray}
f_{N,x,y}(u) &=& \begin{cases} \frac{(N-2)!}{(k-2)!(N-k-1)!} p^{k-2} (1-p)^{N-k-1} \frac{dp}{du} &\;, \|x-y\| \leq u \;,\notag\\ \frac{(N-2)!}{(k-1)!(N-k-2)!} p^{k-1} (1-p)^{N-k-2} \frac{dp}{du} &\;, \|x-y\| > u \end{cases}
\end{eqnarray}
where $p = P(x, r_N(u)) = \int_{t \in B(x,r_N(u))} f(t) dt$. Since $f(x) \leq C_a$ almost everywhere, we have:
\begin{eqnarray}
p \leq Vol(B(x,r_N(u))) \left(\, \sup_{t \in B(x,r_N(u))}  f(t) \,\right) = c_d r_N(u)^d \cdot C_a \leq \frac{2u C_a e^{\psi(k)}}{N}.
\end{eqnarray}
\begin{eqnarray}
\frac{dp}{du} &=& \frac{dp}{d r_N(u)} \frac{d r_N(u)}{du} \leq S(B(x,r_N(u))) \left(\, \sup_{t \in B(x,r_N(u))} f(t) \,\right) \frac{r_N(u)}{ud} \,\notag\\
&\leq& d c_d r_N(u)^{d-1} \cdot C_a \cdot \frac{r_N(u)}{ud} = \frac{c_d r_N(u)^d C_a}{u} \leq \frac{2 C_a e^{\psi(k)}}{N}.
\end{eqnarray}
Therefore, for any $u \leq 1$, we have:
\begin{eqnarray}
f_{N,x,y}(u) &\leq& \begin{cases} \frac{1}{(k-1)!} N^{k-1} p^{k-2} \frac{dp}{du} \leq \frac{1}{(k-1)!} (2u C_a e^{\psi(k)})^{k-2} (2 C_a e^{\psi(k) }) \leq \frac{1}{(k-1)!}(2 C_a e^{\psi(k)})^{k-1} &\;, \|x-y\| \leq u \,\notag\\
\frac{1}{(k-1)!} N^{k} p^{k-1} \frac{dp}{du} \leq \frac{1}{(k-1)!} (2u C_a e^{\psi(k)})^{k-1} (2 C_a e^{\psi(k) }) \leq \frac{1}{(k-1)!}(2 C_a e^{\psi(k)})^{k} &\;, \|x-y\| > u \end{cases}
\end{eqnarray}
So there exists some $C_2$ not depend on $N$ such that $f_{N,x,y}(u) \leq C_2$ for all $u \leq 1$. Therefore, we can bound $\V \left[\, \xi_{k,1}(X) \,|\, X_1 = x, X_2 = y\,\right]$ as:

\begin{eqnarray}
\V \left[\, \xi_{k,1}(X) \,|\, X_1 = x, X_2 = y\,\right] &\leq& \E \left[\, \xi_{k,1}^2(X) \,|\, X_1 = x, X_2 = y \,\right] \,\notag\\
&=& \int_0^{b_N} (\log u)^2 f_{N,x,y}(u) du \,\notag\\
&=& \int_0^1 (\log u)^2 f_{N,x,y}(u) du + \int_1^{b_N} (\log u)^2 f_{N,x,y}(u) du \,\notag\\
&=& C_2 \int_0^2 (\log u)^2 du + (\log b_N)^2 \,\notag\\
&=& C_2 + (1+\delta)(\log \log N)^2 \leq C'_2 (\log \log N)^2
\end{eqnarray}

for some $C'_2 > 0$. Similarly, we know that $\V \left[\, \xi_{k,2}(X) \,|\, X_1 = x, X_2 = y \,\right] \leq C'_2 (\log \log N)^2 $ . Therefore,
\begin{eqnarray}
&& \int_{\|x-y\| \leq 2 a_N} \V \left[\, \xi_{k,1}(X) \,|\, X_1 = x, X_2 = y \,\right]^{1/2}  \V \left[\, \xi_{k,2}(X) \,|\, X_1 = x,X_2 = y\,\right]^{1/2} f(x) f(y) dx dy  \,\notag\\
&\leq& C'_2 (\log \log N)^2 \int_{\|x-y\| \leq 2 a_N} f(x) f(y) dx dy \,\notag\\
&\leq& C'_2 (\log \log N)^2 \Pr \left[\, \|x-y\| \leq 2a_N \,\right] \,\notag\\
&=& C'_2 (\log \log N)^2 \int \left(\, \int_{y \in B(x,2a_N)} f(y) dy\,\right) f(x) dx.
\end{eqnarray}
Notice that by Lemma~\ref{lemma_p_xr}, we know that
\begin{eqnarray}
| \int_{y \in B(x,2a_N)} f(y) dy - f(x) c_d (2a_N)^d | \leq C_3 a_N^{d+2} \leq C_3 a_N^d \;,
\end{eqnarray}
for some constant $C_3 > 0$. So we have $\int_{y \in B(x,2a_N)} f(y) dy \leq f(x) c_d (2a_N)^d + C_3 a_N^d$. Therefore, by plugging in $a_N = (\log(N)^{1+\delta}/N)^{1/d}$, we obtain that:
\begin{eqnarray}
\int \left(\, \int_{y \in B(x,2a_N)} f(y) dy\,\right) f(x) dx &\leq& \int \left(\, f(x) c_d (2a_N)^d + C_3 a_N^d\,\right) f(x) dx \,\notag\\
&\leq& (C_a c_d 2^d + C_3)  \frac{(\log N)^{1+\delta}}{N}.
\end{eqnarray}
Therefore, we know that the  first term of \eqref{eq:cov_xi} is upper bounded by $C_4 (\log\log N)^2 (\log N)^{1+\delta}/N$ for some $C_4$.
\\

Now consider the case that $\|x-y\| > 2a_N$. Then the two balls $B(x,\rho_{k,1})$ and $B(y,\rho_{k,2})$ are disjoint since $\rho_{k,i} \leq a_N$. Therefore, consider the following joint distribution:
\begin{eqnarray}
F_{N,x,y}(u,v) = \Pr \left(\, e^{\xi_{k,1}(X)} < u, e^{\xi_{k,2}(X)} < v \,|\, X_1 = x, X_2 = y\,\right) = \Pr \left(\, \rho_{k,1} < r_N(u), \rho_{k,2} < r_N(v) \,|\, X_1 = x, X_2 = y \,\right).
\end{eqnarray}
Therefore, the covariance can be written as:
\begin{eqnarray}
&& \int_{\|x-y\| > 2 a_N} \E \left[\, \left(\, \xi_{k,1}(X) - \E\left[\,\xi_{k,1}(X)\,\right]\,\right) \left(\, \xi_{k,2}(X) - \E\left[\,\xi_{k,2}(X)\,\right]\,\right) \,|\, X_1 = x, X_2 = y\,\right] f(x) f(y) dx dy \,\notag\\
&=& \int_{\|x-y\| > 2 a_N} \left(\, \int_0^{b_N} \int_0^{b_N} \log u \log v f_{N,x,y}(u,v) du dv - \left(\, \int_0^{b_N} \log u f_{N,x}(u) du \,\right) \left(\, \int_0^{b_N} \log v f_{N,y}(v) dv \,\right) \,\right) f(x) f(y) dx dy \,\notag\\
&=& \int_{\|x-y\| > 2 a_N} \left(\, \int_0^{b_N} \int_0^{b_N} \log u \log v \left(\, f_{N,x,y}(u,v) - f_{N,x}(u) f_{N,y}(v) \,\right) du dv \,\right) f(x) f(y) dx dy \,\notag\\
&\leq& \int_{\|x-y\| > 2 a_N} \left(\, \int_0^{b_N} \int_0^{b_N} \left|\, \log u \log v \,\right| \, \left|\, f_{N,x,y}(u,v) - f_{N,x}(u) f_{N,y}(v) \,\right|\, du dv \,\right) f(x) f(y) dx dy.  \label{eq:cov_xi_geqan}
\end{eqnarray}
Here by the pdf of order statistic~\cite{mcbook}, the pdf of $f_{N,x,y}(u,v)$ and $f_{N,x}(u)$ and $f_{N,y}(v)$ is given by:
\begin{eqnarray}
f_{N,x,y}(u,v) &=& \frac{(N-2)!}{(N-2k-2)!((k-1)!)^2} p^{k-1} q^{k-1} (1-p-q)^{N-2k-2} \frac{dp}{du} \frac{dq}{du} \;, \\
f_{N,x}(u) &=& \frac{(N-2)!}{(N-k-2)!(k-1)!} p^{k-1} (1-p)^{N-k-1} \frac{dp}{du} \;, \\
f_{N,y}(v) &=& \frac{(N-2)!}{(N-k-2)!(k-1)!} q^{k-1} (1-q)^{N-k-1} \frac{dq}{dv}\;,
\end{eqnarray}
where $p = P(x, r_N(u)) = \int_{t \in B(x, r_N(u))} f(t) dt$ and $q = P(y,r_N(v))$ for short. Since $f(x) \leq C_a$ almost everywhere, we have
\begin{eqnarray}
p \leq Vol(B(x,r_N(u))) \left(\, \sup_{t \in B(x,r_N(u))}  f(t) \,\right) = c_d r_N(u)^d \cdot C_a \leq \frac{2u C_a e^{\psi(k)}}{N}.
\end{eqnarray}
\begin{eqnarray}
\frac{dp}{du} &=& \frac{dp}{d r_N(u)} \frac{d r_N(u)}{du} \leq S(B(x,r_N(u))) \left(\, \sup_{t \in B(x,r_N(u))} f(t) \,\right) \frac{r_N(u)}{ud} \,\notag\\
&\leq& d c_d r_N(u)^{d-1} \cdot C_a \cdot \frac{r_N(u)}{ud} = \frac{c_d r_N(u)^d C_a}{u} \leq \frac{2 C_a e^{\psi(k)}}{N}.
\end{eqnarray}
Denote $C_5 = 2 C_a e^{\psi(k)}$ for short, then $p \leq C_5 u/N$ and $\frac{dp}{du} \leq C_5/N$. Similarly, $q \leq C_5 v/N$ and $\frac{dq}{dv} \leq C_5/N$. Then we can upper bound the difference of $|f_{N,x,y}(u,v) - f_{N,x}(u) f_{N,y}(v)|$ by:
\begin{eqnarray}
&&\big|\, f_{N,x,y}(u,v) - f_{N,x}(u) f_{N,y}(v) \,\big| \,\notag\\
&=& \frac{1}{((k-1)!)^2} p^{k-1} q^{k-1} \big|\, \frac{(N-2)!}{(N-2k-2)!} (1-p-q)^{N-2k-2} - (\frac{(N-2)!}{(N-k-2)!})^2 (1-p)^{N-k-1}(1-q)^{N-k-1} \,\big| \frac{dp}{du} \frac{dq}{du} \,\notag\\
&\leq& \frac{1}{((k-1)!)^2} (\frac{C_5 u}{N})^{k-1} (\frac{C_5 v}{N})^{k-1} (\frac{C_5}{N})^2 \big|\, \frac{(N-2)!}{(N-2k-2)!} (1-p-q)^{N-2k-2} - (\frac{(N-2)!}{(N-k-2)!})^2 (1-p)^{N-k-1}(1-q)^{N-k-1} \,\big|  \,\notag\\
&\leq& \frac{1}{((k-1)!)^2} \frac{C_5^{2k} u^{k-1} v^{k-1}}{N^{2k}} \left(\,Q_1 + Q_2 + Q_3\,\right) \;, \label{eq:Q123}
\end{eqnarray}
where
\begin{eqnarray}
Q_1 &=& \frac{(N-2)!}{(N-2k-2)!} \left(\, (1-p-q)^{N-2k-2} - (1-p-q)^{N-k-1} \,\right) \;, \\
Q_2 &=& \big|\, (\frac{(N-2)!}{(N-k-2)!})^2 - \frac{(N-2)!}{(N-2k-2)!}\,\big| \, (1-p-q)^{N-k-1} \;, \\
Q_3 &=& (\frac{(N-2)!}{(N-k-2)!})^2 \left(\, (1-p)^{N-k-1}(1-q)^{N-k-1} - (1-p-q)^{N-k-1}\,\right) \;.
\end{eqnarray}
We will bound the three terms separately. For $Q_1$, notice that $(N-2)!/(N-2k-2)! \leq N^{2k}$ and
\begin{eqnarray}
(1-p-q)^{N-k-1} - (1-p-q)^{N-2k-2} \leq 1 - (1-p-q)^{k+1} \leq (k+1)(p+q) \leq \frac{(k+1)C_5(u+v)}{N}.
\end{eqnarray}
So $Q_1 \leq (k+1)C_5(u+v) N^{2k-1}$. For $Q_2$, notice that both $(N-2)!/(N-2k-2)!$ and $(\frac{(N-2)!}{(N-k-2)!})^2$ are polynomial of $N$ with $2k$ order, moreover, the coefficient of $N^{2k}$ are both 1. So they differs at most $C_6 N^{2k-1}$, where $C_6$ is some constant relevant to $k$. $(1-p-q)^{N-k-1}$ is simply upper bounded by 1. For $Q_3$, notice that $(\frac{(N-2)!}{(N-k-2)!})^2 \leq N^{2k}$ and
\begin{eqnarray}
&&(1-p)^{N-k-1}(1-q)^{N-k-1} - (1-p-q)^{N-k-1} \,\notag\\
&=& (1-p-q+pq)^{N-k-1} - (1-p-q)^{N-k-1} \,\notag\\
&\leq& (N-k-1)pq(1-p-q+pq)^{N-k-2} \leq Npq \leq \frac{C_5^2 uv}{N}.
\end{eqnarray}
Therefore, $Q_3 \leq C_5^2 uv N^{2k-1}$. Combine the upper bounds of $Q_1, Q_2, Q_3$ into ~\eqref{eq:Q123}, we obtain:
\begin{eqnarray}
&&\big|\, f_{N,x,y}(u,v) - f_{N,x}(u) f_{N,y}(v) \,\big| \,\notag\\
&\leq& \frac{1}{((k-1)!)^2} \frac{C_5^{2k} u^{k-1} v^{k-1}}{N^{2k}} \left(\, (k+1)C_5(u+v) N^{2k-1} + C_6 N^{2k-1} + C_5^2 uv N^{2k-1}\,\right) \,\notag\\
&\leq& \frac{C_7}{N} u^{k-1} v^{k-1} (1+u+v+uv) \;,
\end{eqnarray}
for some $C_7 > 0$. Plug this in \eqref{eq:cov_xi_geqan}, we obtain:
\begin{eqnarray}
&& \int_{\|x-y\| > 2 a_N} \E \left[\, \left(\, \xi_{k,1}(X) - \E\left[\,\xi_{k,1}(X)\,\right]\,\right) \left(\, \xi_{k,2}(X) - \E\left[\,\xi_{k,2}(X)\,\right]\,\right) \,|\, X_1 = x, X_2 = y\,\right] f(x) f(y) dx dy \,\notag\\
&\leq& \int_{\|x-y\| > 2 a_N} \left(\, \int_0^{b_N} \int_0^{b_N} \left|\, \log u \log v \,\right| \big|\, f_{N,x,y}(u,v) - f_{N,x}(u) f_{N,x}(v) \,\big| du dv \,\right) f(x) f(y) dx dy \,\notag\\
&\leq& \int_{\|x-y\| > 2 a_N} \left(\, \frac{C_7}{N} \int_0^{b_N} \int_0^{b_N} \left|\, \log u \log v \,\right| \cdot u^{k-1} v^{k-1} (1+u+v+uv) du dv \,\right) f(x) f(y) dx dy \,\notag\\
&\leq& \int_{\|x-y\| > 2 a_N} \left(\, \frac{C_7}{N} (\log b_N)^2 b_N^{2k+2} \,\right) f(x) f(y) dx dy \,\notag\\
&\leq& \frac{C_7}{N} (\log b_N)^2 b_N^{2k+2} \;.
\end{eqnarray}
By substituting $b_N = (\log N)^{1+\delta}$, we obtain the desired claim.

%Minimax Lower Bound
\section{Proof of Theorem \ref{thm:minimax_lower_bound} on the minimax lower bound}
\label{sec:proof_lowerbound}

The proof is based on the standard Le Cam's method \cite{Yu97}. First we will prove the $\Omega(1/N)$ lower bound. Consider two Gaussian distributions $P = \mathcal{N}(0, I_d)$ and $Q= \mathcal{N}(0, (1+\delta)(I_d))$. The norm of Hessian matrix of $P$ and $Q$ are both bounded, so $P,Q \in \mathcal{F}_d$. Then we claim that: $H(P) = d \log(2\pi e)/2$ and $H(Q) = d \log(2\pi e)/2 + d\log(1+\delta)/2$. Applying Le Cam's method, the minimax lower bound is bounded by:
\begin{eqnarray}
\inf_{\widehat{H}_n} \sup_{f \in \mathcal{F}_d} \E \left[\, \left(\, \widehat{H}_n(X) - H(X) \,\right)^2 \,\right] &\geq& \frac12   |H(P) - H(Q)|^2 (1-\|P^N - Q^N\|_{TV}) \,\notag\\
&\geq& \frac12   \left(\, d \log(1+\delta)/2 \,\right)^2 \left(\, 1 - \sqrt{2 - 2 \left(\,1 - d_h(P,Q)^2\,\right)^N}\,\right),
\end{eqnarray}
where $d_h(P,Q) = \int (\sqrt{p(x)} - \sqrt{q(x)})^2 dx = 2 - 2\int \sqrt{p(x) q(x)} dx$ is the Hellinger distance of $P$ and $Q$, and $\|P-Q\|_{\rm TV}$ is the total variation between $P$ and $Q$. We claim that  $d_h(P, Q)^2$ is bounded as:
\begin{eqnarray}
d_h^2(P, Q) = O(d \delta^2)\;.
\label{eq:lem_hellinger}
\end{eqnarray}

Therefore, by choose $\delta = \Theta(\sqrt{1/dN})$ such that $\left(\, 1 - \sqrt{2 - 2 \left(\,1 - d_h(P,Q)^2\,\right)}\,\right) > 1/2$ the minimax lower bound is given by:
\begin{eqnarray}
\inf_{\widehat{H}_n} \sup_{f \in \mathcal{F}_d} \E \left[\, \left(\, \widehat{H}_n(X) - H(X) \,\right)^2 \,\right] &\geq& \frac{1}{4} \left(\, d \log(1+\sqrt{1/dN})/2 \,\right)^2 = \Omega(d/N).
\end{eqnarray}
%for $\mathcal{F} = \mathcal{F}_d$. or $\mathcal{F} = \mathcal{D}_{\Sigma(s,L)}$.

We are now left to prove \eqref{eq:lem_hellinger}:
%The Hellinger distance of $P$ and $Q$, where $P = \mathcal{N}(0, I_d)$ and $Q = \mathcal{N}(0, (1+\delta)I_d)$ is given by:
\begin{eqnarray}
d_h(P,Q)^2 &=& 2 - 2 \int \sqrt{p(x) q(x)} \, dx \,\notag\\
&=& 2 - 2 \int \sqrt{\frac{1}{(2\pi)^{d/2}} \exp\Big\{-\frac{x^2}{2}\Big\} \frac{1}{(2\pi)^{d/2}(1+\delta)^{d/2}} \exp\Big\{-\frac{x^2}{2(1+\delta)}\Big\}} \, dx \,\notag\\
&=& 2 - 2 \int \sqrt{\frac{1}{(2\pi)^{d}(1+\delta)^{d/2}} \exp\Big\{-(\frac{1}{2}+\frac{1}{2(1+\delta)})x^2\Big\}} \, dx \,\notag\\
&=& 2 - 2 \int \frac{1}{(2\pi)^{d/2}(1+\delta)^{d/4}} \exp\Big\{-\frac{x^2}{2 \left( \frac{2+2\delta}{2+\delta} \right) }\Big\} \, dx
\,\notag\\
&=& 2 - 2 \left(\,\frac{(2\pi)^{d/2}\left( \frac{2+2\delta}{2+\delta} \right)^{d/2}}{(2\pi)^{d/2}(1+\delta)^{d/4}} \,\right)\,\notag\\
&=& 2 - 2 \left(\, \frac{1+\delta}{(1+\delta/2)^2} \, \right)^{d/4} \,\notag\\
&\leq& 2 - 2 \left( 1 - \frac{d}{4} \cdot \frac{\delta^2/4}{(1+\delta/2)^2}\right) \leq \frac{d \delta^2}{8},
\end{eqnarray}
where we use the fact that $(1-x)^N \leq 1-Nx$ for $x \leq 1$ to obtain the inequality.

The proof of the $\Omega(N^{-16/(d+8)})$ lower bound follows closely the proof of lower bound in~\cite{krishnamurthy2014nonparametric}. We will use the following lemma, which is an extension of Le Cam's method:
\begin{lemma}
\label{lem_le_cam}
Let $H$ be a functional defined on some class of functions $\mathcal{F}$. We have $u \in \mathcal{F}$ and $v_{\lambda} \in \mathcal{F}$ for any $\lambda$ in some finite index set $\Lambda$. Define $\bar{v}^N = \frac{1}{|\Lambda|} \sum_{\lambda \in \Lambda} v_{\lambda}^N$. If we have:
\begin{enumerate}
    \item For any $v_{\lambda}$, we have $H(u) - H(v_{\lambda}) > \alpha$.
    \item $\|u^N - \bar{v}^N\|_{TV} \leq \beta$.
\end{enumerate}
Then the minimax lower bound is given by:
\begin{eqnarray}
\inf_{\widehat{H}_n} \sup_{f \in \mathcal{F}} \E \left[\, \left(\, \widehat{H}_n(X) - H(X) \,\right)^2 \,\right] \geq K \cdot \alpha^2 (1-\beta)
\end{eqnarray}
for some constant $K>0$.
\label{lem:lecam}
\end{lemma}

Now let $u$ be the uniform distribution over $[0,1]^d$. To construct the $v_{\lambda}$ functions, we partition the space $[0,1]^d$ to $m^d$ hypercubes denoted by $R_1, \dots, R_{m^d}$. Let $t_j: R_j \to [0,1]^d$ maps the small hypercube $R_j$ to $[0,1]^d$. We pick a function $g$ supported on $[0,1]^d$ such that:
\begin{enumerate}
    \item $\int_{[0,1]^d} g(x) dx = 0$
    \item $\int_{[0,1]^d} g^2(x) dx = 1$
    \item $g$ belongs to the smoothness class $\mathcal{F}_d$.
\end{enumerate}
We define $u(x)$ to be the uniform distribution on $[0,1]^d$ and $v_{\lambda}$ by adding an appropriately chosen perturbation:
\begin{eqnarray}
v_{\lambda}(x) = u(x) + m^{-\gamma} \sum_{j=1}^{m^d} \lambda_j \mathbb{I}\{x \in R_j\} g(t_j(x))
\end{eqnarray}
for any $\lambda \in \Lambda = \{\pm 1\}^{m^d}$. Here we need $\gamma \geq 2$ to make sure that $v_{\lambda} \in \mathcal{F}_d$.
%and $\gamma \geq s$ to make sure that $v_{\lambda} \in \Sigma(s, L)$.
We claim the following:

\begin{lemma}
\begin{eqnarray}
H(u) - H(v_{\lambda}) \geq \frac{1}{3}m^{-2\gamma}, \forall \lambda \in \{\pm 1\}^{m^d}.
\end{eqnarray}
\label{lem_difference_h}
\end{lemma}

\begin{lemma}
\begin{eqnarray}
\| u^N - \frac{1}{|\Lambda|} v_{\lambda^N} \|_{TV}^2 \leq O(N^2 m^{-d-4\gamma}).
\end{eqnarray}
\label{lem_TV}
\end{lemma}

Therefore, let $m = \Theta(N^{2/(d+4\gamma)})$ such that $\| u^N - \frac{1}{|\Lambda|} v_{\lambda^N} \|_{TV} \leq 1/2$, and by applying Lemma~\ref{lem_le_cam}, we know that:
\begin{eqnarray}
\inf_{\widehat{H}_N} \sup_{f \in \mathcal{F}_d} \E \left[\, \left(\, \widehat{H}_N(X) - H(X) \,\right)^2 \,\right] \geq \Omega ( N^{-8\gamma/(d+4\gamma)} ).
\end{eqnarray}
We obtain the minimax lower bound of $N^{-16/(d+8)}$ by plugging in $\gamma = 2$.
%For $\mathcal{F} = \Sigma(s, L)$, we obtain a minimax lower bound of $N^{-8s/(d+4s)}$ by plugging in $\gamma = s$.

\subsection{Proof of Lemma~\ref{lem_difference_h}}

It is obvious that the entropy of uniform distribution is highest. So $H(u) > H(v_{\lambda})$. Their difference is given by:
\begin{eqnarray}
H(u) - H(v_{\lambda}) &=& - \int_{[0,1]^d} u(x) \log u(x) dx + \int_{[0,1]^d} v_{\lambda}(x) \log v_{\lambda}(x) dx \,\notag\\
&=& \sum_{j=1}^{m^d} \int_{R_j} v_{\lambda}(x) \log v_{\lambda}(x) dx \,\notag\\
&=& \sum_{j=1}^{m^d} \int_{R_j} \left(\, u(x) + m^{-\gamma} \lambda_j g(t_j(x))\right) \log \left( u(x) + m^{-\gamma} \lambda_j g(t_j(x)) dx \,\right) \,\notag\\
&\geq& \sum_{j=1}^{m^d} \int_{R_j} \left(\, m^{-\gamma} \lambda_j g(t_j(x)) + \frac{1}{3} (m^{-\gamma} \lambda_j g(t_j(x)))^2 dx \,\right) \,\notag\\
&=& m^{-\gamma} \sum_{j=1}^{m^d} \lambda_j \int_{R_j} g(t_j(x)) dx + \frac{1}{3} m^{-2\gamma} \sum_{j=1}^{m^d} \lambda^2_j \int_{R_j} g^2(t_j(x)) dx \,\notag\\
&=&\frac{1}{3} m^{-2\gamma} \sum_{j=1}^{m^d} m^{-d} = \frac{1}{3}m^{-2\gamma}.
\end{eqnarray}
Here the inequality comes from the fact that $x \log x \geq (x-1) + \frac{1}{3}(x-1)^2$ for $x \in (0.5,1.5)$.

\subsection{Proof of Lemma~\ref{lem_TV}}

The proof uses the fact that $\|p(x) - q(x)\|_{TV}^2 \leq \E_P [(\frac{q(x)}{p(x)})^2] - 1$, which comes immediately from Cauchy-Schwarz inequality. So
\begin{eqnarray}
\E_{u^N}\Big[\Big(\frac{\bar{v}^N(x)}{u(x)}\Big)^2\Big] &=& \E_{u^N} \Big[\frac{1}{|\Lambda|^2} \sum_{\lambda, \mu \in \Lambda} \frac{v_{\lambda}^N(x) v_{\mu}^N(x)}{(u^N(x))^2} \Big] \,\notag\\
&=& \frac{1}{|\Lambda|^2} \sum_{\lambda, \mu \in \Lambda} \E_{u^N} \Big[\frac{v_{\lambda}^N(x) v_{\mu}^N(x)}{(u^N(x))^2}\Big] \,\notag\\
&=& \frac{1}{|\Lambda|^2} \sum_{\lambda, \mu \in \Lambda}  \left(\, \E_u \Big[\frac{v_{\lambda}(x) v_{\mu}(x)}{(u(x))^2}\Big] \,\right)^N \,\notag\\
&=& \frac{1}{|\Lambda|^2} \sum_{\lambda, \mu \in \Lambda} \left(\, \sum_{j=1}^{m^d} \int_{R_j} (1 + m^{-\gamma} \lambda_j g(t_j(x))) (1 + m^{-\gamma} \mu_j g(t_j(x))) dx\right)^N \,\notag\\
&=& \frac{1}{|\Lambda|^2} \sum_{\lambda, \mu \in \Lambda} \left(\, \sum_{j=1}^{m^d} \int_{R_j} (1 + m^{-\gamma} (\lambda_j+\mu_j) g(t_j(x)) + m^{-2\gamma} \lambda_j \mu_j g^2(t_j(x))) dx \,\right)^N \,\notag\\
&=& \frac{1}{|\Lambda|^2} \sum_{\lambda, \mu \in \Lambda} \left(\, 1 + m^{-\gamma} \sum_{j=1}^{m^d} (\lambda_j + \mu_j) \int_{R_j} g(t_j(x)) dx + m^{-2\gamma} \sum_{j=1}^{m^d} \lambda_j \mu_j \int_{R_j} g^2(t_j(x)) dx \,\right)^N \,\notag\\
&=& \frac{1}{|\Lambda|^2} \sum_{\lambda, \mu \in \Lambda} \left(\, 1 + m^{-d-2\gamma} \sum_{j=1}^{m^d} \lambda_j \mu_j  \,\right)^N \,\notag\\
&\leq& \frac{1}{|\Lambda|^2} \sum_{\lambda, \mu \in \Lambda} \exp\left\{\, N m^{-d-2\gamma} \sum_{j=1}^{m^d} \lambda_j \mu_j\,\right\} \,\notag\\
&\leq& \frac{1}{|\Lambda|^2} \sum_{\lambda, \mu \in \Lambda} \left(\, 1 + N m^{-d-2\gamma} \sum_{j=1}^{m^d} \lambda_j \mu_j + n^2 m^{-2d-4\gamma} (\sum_{j=1}^{m^d} \lambda_j \mu_j)^2 \,\right) \,\notag\\
&=& 1 + N m^{-d-2\gamma} \sum_{j=1}^{m^d} \sum_{\lambda_j, \mu_j \in \{\pm 1\}} \lambda_j \mu_j + N^2 m^{-2d-4\gamma} \sum_{j=1}^{m^d} \sum_{k=1}^{m^d} \sum_{\lambda_j, \lambda_k, \mu_j, \mu_k \in \{\pm 1\}} \lambda_j \mu_j \lambda_k \mu_k \,\notag\\
&=& 1 + N^2 m^{-2d-4\gamma} \sum_{j=1}^{m^d} \sum_{\lambda_j, \mu_j  \in \{\pm 1\}} \lambda^2_j \mu^2_j \,\notag\\
&=& 1 + 4 N^2 m^{-d-4\gamma}
\end{eqnarray}
where the first inequality comes from the fact that $1+x \leq e^x$ and the second comes from $e^x \leq 1 + x + x^2$ for $x \leq 2$. Therefore, we have $\| u^n - \frac{1}{|\lambda|} v_{\lambda^n} \|_{TV}^2 \leq O(N^2 m^{-d-4\gamma})$.

\section{Proof of Theorem \ref{thm:consistent_KSG} on the consistency of KSG estimator}
\label{sec:proof_KSG_consistency}

Note that
\begin{eqnarray*}
    \hI_{KSG}(X;Y) &=& \hH_{KSG}(X) + \hH_{KSG}(Y) - \hH_{KL,\infty}(X,Y)\;, \\
    \hI_{BI-KSG}(X;Y) &=& \hH_{BI-KSG}(X) + \hH_{BI-KSG}(Y) - \hH_{KL,2}(X,Y) \;,
\end{eqnarray*}
where
\begin{eqnarray}
    \hH_{KL,\infty}(X,Y) &\equiv& -\psi(k) + \log N + \log{c_{d_x,\infty} c_{d_y,\infty}} + (d_x+d_y) \log \rho_{k,i,\infty} \;, \label{def:hHxy_KSG}\\
    \hH_{KSG}(X) &\equiv& -\frac{1}{N}\sum_{i=1}^N \psi(n_{x,i,\infty} + 1) + \log N + \log c_{d_x,\infty} + d_x \log \rho_{k,i,\infty}\;, \label{def:hHx_KSG} \\
    \hH_{KSG}(Y) &\equiv& -\frac{1}{N}\sum_{i=1}^N \psi(n_{y,i,\infty} + 1) + \log N + \log c_{d_y,\infty} + d_y \log \rho_{k,i,\infty}\;, \label{def:hHy_KSG}
\end{eqnarray}
and
\begin{eqnarray}
    \hH_{KL,2}(X,Y) &\equiv& -\psi(k) + \log N + \log{c_{d_x+d_y,2}} + (d_x+d_y) \log \rho_{k,i,2} \;, \label{def:hHxy}\\
    \hH_{BI-KSG}(X) &\equiv& -\frac{1}{N}\sum_{i=1}^N \log {n_{x,i,2}} + \log N + \log c_{d_x,2} + d_x \log \rho_{k,i,2}\;, \label{def:hHx}\\
    \hH_{BI-KSG}(Y) &\equiv& -\frac{1}{N}\sum_{i=1}^N \log {n_{y,i,2}} + \log N + \log c_{d_y,2} + d_y \log \rho_{k,i,2}.  \label{def:hHy}
\end{eqnarray}

We prove the following technical lemma that shows the convergence of the marginal entropy estimate~\eqref{def:hHx_KSG} and~\eqref{def:hHx} . The convergence of~\eqref{def:hHy_KSG} and~\eqref{def:hHy} is immediate by interchanging $X$ and $Y$. The convergence in probability of the joint entropy estimate~\eqref{def:hHxy_KSG} and ~\eqref{def:hHxy} are known from~\cite{Kra04}. This proves the desired claim.
%$ \lim_{N\to\infty} \Pr\left(\left|\hH_{k,N}(X,Y) - H(X,Y)\right|\varepsilon\right)=0 $T.

\begin{lemma}
    \label{lem:Hx}
    Under the hypotheses of Theorem \ref{thm:consistent_KSG},
    the estimated marginal entropy converges to the true entropy, i.e. for all $\varepsilon > 0$
    \begin{eqnarray}
    \lim_{N \to \infty} \Pr\left(\, \left|\hH_{KSG}(X) - H(X)\right| > \varepsilon\,\right) = 0 \;,
    \end{eqnarray}
    \begin{eqnarray}
    \lim_{N \to \infty} \Pr\left(\, \left|\hH_{BI-KSG}(X) - H(X)\right| > \varepsilon\,\right) = 0.
    \end{eqnarray}
\end{lemma}

\subsection{Proof of Lemma \ref{lem:Hx}}
Define
\begin{eqnarray}
\widehat{f}^{KSG}_X(X_i) &\equiv& \frac{\exp\{\psi(n_{x,i,\infty} + 1)\}}{ N c_{d_x,\infty}\rho_{k,i,\infty}^{d_x}} \;,
\end{eqnarray}
and
\begin{eqnarray}
\widehat{f}^{BI-KSG}_X(X_i) &\equiv& \frac{n_{x,i,2}}{N c_{d_x,2}\rho_{k,i,2}^{d_x}} \;,
\end{eqnarray}
such that $\widehat{H}_{KSG}(X) = -\frac{1}{N} \sum_{i=1}^N \log \widehat{f}_X^{KSG}(X_i)$ and $\widehat{H}_{BI-KSG}(X) = -\frac{1}{N} \sum_{i=1}^N \log \widehat{f}_X^{KSG}(X_i)$. From now on we will skip the subscript KSG or BI-KSG and the subscript $2$ or $\infty$ if the formula holds for both. We will specify it whenever necessary. Now we write $|\widehat{H}(X) - H(X)|$ as:
\begin{eqnarray}
    && \left| \widehat{H}(X) - H(X) \right| \notag \\
    & = & \left|-\frac{1}{N}\sum_{i=1}^N \log \widehat{f}_X(X_i) - \left(-\int f_X(x) \log f_X(x) dx\right)\right| \notag\\
    &\leq & \left|\frac{1}{N} \sum_{i=1}^N \log f_X(X_i) - \int f_X(x) \log f_X(x) dx\right| +\frac{1}{N} \sum_{i=1}^N \left|\log \widehat{f}_X(X_i) - \log f_X(X_i)\right|.
\end{eqnarray}
The first term is the error from the empirical mean. Notice that $\log f_X(X_i)$ are i.i.d.\ random variables, satisfying
\begin{eqnarray}
    \E \left|\,\log f_X(X_i)\,\right| = \int f_X(x) |\log f_X(x)| dx < +\infty
\end{eqnarray}
where the mean is given by:
\begin{eqnarray}
    \E \left(\,\log f_X(X_i)\,\right) = \int f_X(x) \log f_X(x) dx.
\end{eqnarray}
Therefore, by weak law of large numbers, we have:
\begin{eqnarray}
    \lim_{N \to \infty}\Pr\left(\,\left|\frac{1}{N} \sum_{i=1}^N \log f_X(X_i) - \int f_X(x) \log f_X(x) dx\right| > \varepsilon\,\right) = 0
\end{eqnarray}
for any $\varepsilon > 0$.
\\

The second term comes from density estimation. We denote $Z = (X,Y)$ and $f(z) = f(x,y)$ for short, then for any fixed $\varepsilon > 0$, we obtain:
\begin{eqnarray}
    && \Pr\left(\,\frac{1}{N} \sum_{i=1}^N \left|\log \widehat{f}_X(X_i) - \log f_X(X_i)\right| > \varepsilon\,\right) \notag\\
    & \leq & \Pr\left(\,\bigcup_{i=1}^N \Big\{ \left|\log \widehat{f}_X(X_i) - \log f_X(X_i)\right| > \varepsilon \Big\} \,\right) \notag \\
    & \leq & N \cdot \Pr\left(\,\left|\log \widehat{f}_X(X_i) - \log f_X(X_i)\right| > \varepsilon\,\right) \notag\\
    & = & N \int \underbrace{\Pr\left(\,\left|\log \widehat{f}_X(X_i) - \log f_X(X_i)\right| > \varepsilon\big|Z_i = z\,\right)}_{= I_1(z) + I_2(z) + I_3(z)} f(z) dz
\end{eqnarray}
where
\begin{eqnarray}
     && I_1(z) = \Pr\left(\,\rho_{k,i} > \log{N} (N f(z)c_{d_x+d_y})^{-\frac{1}{d_x+d_y}}\big|Z_i = z\,\right) \\
     && I_2(z) = \Pr\left(\,\rho_{k,i} < (\log{N})^2 (N f_X(x) c_{d_x})^{-\frac{1}{d_x}}\big|Z_i = z = (x,y)\,\right) \\
     && I_3(z) = \int_{r=(\log{N})^2 (N f_X(x) c_{d_x})^{-\frac{1}{d_x}}}^{\log{N} (N f(z)c_{d_x+d_y})^{-\frac{1}{d_x+d_y}}} \Pr\left(\,\left|\log \widehat{f}_X(X_i) - \log f_X(X_i)\right| > \varepsilon\big| \rho_{k,i} = r, Z_i = z \,\right) f_{\rho_{k,i}}(r) dr
\end{eqnarray}
where $f_{\rho_{k,i}}(r)$ is the pdf of $\rho_{k,i}$ given $Z_i = z$. We will consider the three terms separately, and show that each is bounded by $o(N^{-1})$.

\textit{$I_1$:} Let $B_Z(z, r) = \{Z: \|Z-z\| < r\}$ be the $(d_x+d_y)$-dimensional ball centered at $z$ with radius $r$. Since the Hessian matrix of $H(f)$ exists and $\|H(f)\|_2 < C$ almost everywhere, then for sufficiently small $r$, there exists $z'$ such that
\begin{eqnarray}
&& \Pr\left(\,u \in B_Z(z,r)\,\right) = \int_{\|u-z\| \leq r} f(u) du \notag \\
& = & \int_{\|u-z\| \leq r} f(z) + (u-z)^T \nabla f(z) + (u-z)^T H_f(z') (u-z)  du \notag \\
& \in & \left[\, f(z)c_{d_x+d_y} r^{d_x+d_y}(1 - C r^2)), f(z)c_{d_x+d_y} r^{d_x+d_y}(1 + C r^2))\,\right].
\end{eqnarray}
Then for sufficiently large $N$,
\begin{eqnarray}
    p_1 & = & \Pr\left(\,u \in B_Z(z,\log{N} (N f(z)c_{d_x+d_y})^{-\frac{1}{d_x+d_y}})\,\right) \notag \\
    & \geq & f(z) c_{d_x+d_y} \left(\, \log{N} (N f(z)c_{d_x+d_y})^{-\frac{1}{d_x+d_y}} \,\right)^{d_x+d_y} \left(\, 1-C (\log{N} (N f(z)c_{d_x+d_y})^{-\frac{1}{d_x+d_y}})^2 \,\right) \notag \\
    & \geq & \frac{(\log{N})^{d_x+d_y}}{2N}
\end{eqnarray}
Therefore, $I_1(z)$ is upper bounded by:
\begin{eqnarray}
I_1(z) & = & \Pr\left(\,\rho_{k,i} > \log{N} (N f(z)c_{d_x+d_y})^{-\frac{1}{d_x+d_y}}\big| Z_i = z\,\right) \notag \\
& = & \sum_{m=0}^{k-1} {N \choose m} p_1^m (1-p_1)^{N-1-m} \notag \\
&\leq & \sum_{m=0}^{k-1} N^m (1-p_1)^{N-1-m} \notag \\
&\leq & k N^{k-1} (1-\frac{(\log{N})^{d_x+d_y}}{2N})^{N-k-1} \notag \\
&\leq & k N^{k-1} \exp\{-\frac{(\log{N})^{d_x+d_y}(N-k-1)}{2N}\} \notag \\
&\leq & k N^{k-1}\exp\{-\frac{(\log{N})^{d_x+d_y}}{4}\}
\end{eqnarray}
for any $d_x, d_y \geq 1$.
\\

\textbf{$I_2$:} For sufficiently large $N$, we have
\begin{eqnarray}
    p_2 & = &\Pr\left(\,u \in B_Z(z,(\log{N})^2 (N f_X(x) c_{d_x})^{-\frac{1}{d_x}})\,\right) \notag \\
    & \leq & f(z) c_{d_x+d_y} \left(\, (\log{N})^2 (N f_X(x)c_{d_x})^{-\frac{1}{d_x}} \,\right)^{d_x+d_y} \left(\, 1+C (\log{N} (N f_X(x)c_{d_x})^{-\frac{1}{d_x}})^2 \,\right) \notag \\
    & \leq & \frac{2f(z) c_{d_x+d_y}}{(f(x)c_{d_x})^{\frac{d_x+d_y}{d_x}}} (\log{N})^{2(d_x+d_y)} N^{-\frac{d_x+d_y}{d_x}} \notag \\
    & \leq & 2f_{Y|X}(y|x) \frac{c_{d_x+d_y}}{c_{d_x}} (\log{N})^{2(d_x+d_y)} N^{-\frac{d_x+d_y}{d_x}} \notag \\
    & \leq & 2C_e \frac{c_{d_x+d_y}}{c_{d_x}} (\log{N})^{2(d_x+d_y)} N^{-\frac{d_x+d_y}{d_x}}.
\end{eqnarray}

$I_2$ is upper bounded by:
\begin{eqnarray}
I_2(z) &= & \Pr\left(\,\rho_{k,i} < \log{N} (N f_X(x) c_{d_x})^{-\frac{1}{d_x}}\big| Z_i = z\,\right) \notag \\
& = &\sum_{m=k}^{N-1} {N-1 \choose m} p_2^m (1-p_2)^{N-1-m} \notag \\
&\leq &\sum_{m=k}^{N-1} N^m p_2^m \notag \\
&\leq &\sum_{m=k}^{N-1} (2C_e \frac{c_{d_x+d_y}}{c_{d_x}} (\log{N})^{2(d_x+d_y)} N^{-\frac{d_y}{d_x}})^m \notag \\
&\leq &(4C_e \frac{c_{d_x+d_y}}{c_{d_x}})^k (\log{N})^{2K(d_x+d_y)} N^{-\frac{kd_y}{d_x}},
\end{eqnarray}
for any $d_x, d_y \geq 1$ and $k \geq 1$.
\\

\textbf{$I_3$:} Now we will consider KSG and BI-KSG separately. Also we need to specify whether we are considering $\ell_2$ or $\ell_{\infty}$ norm. For KSG, given that $Z_i = z = (x,y)$ and $\rho_{k,i,\infty} = r$, we have:

\begin{eqnarray}
    && \Pr\left(\,\left|\log \widehat{f}^{KSG}_X(X_i) - \log f_X(X_i)\right| > \varepsilon\big| \rho_{k,i,\infty} = r, Z_i = z\,\right) \notag \\
    &=&\Pr\left(\,\left|\psi(n_{x,i,\infty}+1) - \log N - \log c_{d_x,\infty} - d_x \log \rho_{k,i,\infty} - \log f_X(x)\right| > \varepsilon\big| \rho_{k,i,\infty} = r, Z_i = z\,\right). \notag \\
\end{eqnarray}
Notice that for any integer $x \geq 2$, we have $\log(x-1) < \psi(x) < \log(x)$. Therefore
\begin{eqnarray}
&&\Pr\left(\psi(n_{x,i,\infty}+1) - \log N - \log c_{d_x,\infty} - d_x \log \rho_{k,i,\infty} - \log f_X(x) < -\varepsilon\big| \rho_{k,i,\infty} = r, Z_i = z\,\right) \notag\\
&\leq& \Pr\left(\log n_{x,i,\infty} - \log N - \log c_{d_x,\infty} - d_x \log \rho_{k,i,\infty} - \log f_X(x) < -\varepsilon\big| \rho_{k,i,\infty} = r, Z_i = z\,\right) \notag\\
&=& \Pr\left(n_{x,i,\infty} < N c_{d_x,\infty}r^{d_x}f_X(x)e^{-\varepsilon} \big| \rho_{k,i,\infty} = r, Z_i = z\,\right).
\end{eqnarray}
In the other direction,
\begin{eqnarray}
&&\Pr\left(\psi(n_{x,i,\infty}+1) - \log N - \log c_{d_x,\infty} - d_x \log \rho_{k,i,\infty} - \log f_X(x) > \varepsilon\big| \rho_{k,i,\infty} = r, Z_i = z\,\right) \notag\\
&\leq& \Pr\left(\log(n_{x,i,\infty}+1) - \log N - \log c_{d_x,\infty} - d_x \log \rho_{k,i,\infty} - \log f_X(x) > \varepsilon\big| \rho_{k,i,\infty} = r, Z_i = z\,\right) \notag\\
&=& \Pr\left(n_{x,i,\infty} > Nc_{d_x,\infty}r^{d_x}f_X(x)e^{\varepsilon}-1 \big| \rho_{k,i,\infty} = r, Z_i = z\,\right).
\end{eqnarray}

For BI-KSG, we have:
\begin{eqnarray}
    && \Pr\left(\,\left|\log \widehat{f}^{BI-KSG}_X(X_i) - \log f_X(X_i)\right| > \varepsilon\big| \rho_{k,i,2} = r, Z_i = z\,\right) \notag \\
    &=&\Pr\left(\,\left|\log n_{x,i,2} - \log N - \log c_{d_x,2} - d_x \log \rho_{k,i,2} - \log f_X(x)\right| > \varepsilon\big| \rho_{k,i,2} = r, Z_i = z\,\right) \notag \\
    &=& \Pr\left(\,\left|\log n_{x,i,2} - \log N c_{d_x,2}r^{d_x}f_X(x)\right| > \varepsilon\big| \rho_{k,i,2} = r, Z_i = z\,\right) \notag \\
    &=& \Pr\left(\, n_{x,i,2} > N c_{d_x,2}r^{d_x}f_X(x)e^{\varepsilon} \big| \rho_{k,i,2} = r, Z_i = z\,\right) \notag\\
    &+& \Pr\left(\, n_{x,i,2} < N c_{d_x,2}r^{d_x}f_X(x)e^{-\varepsilon}\big| \rho_{k,i,2} = r, Z_i = z\,\right) .
\end{eqnarray}

Combine them together, we have:
\begin{eqnarray}
    && \Pr\left(\,\left|\log \widehat{f}_X(X_i) - \log f_X(X_i)\right| > \varepsilon\big| \rho_{k,i} = r, Z_i = z\,\right) \notag \\
    &\leq& \Pr\left(n_{x,i} < N c_{d_x}r^{d_x}f_X(x)e^{-\varepsilon} \big| \rho_{k,i} = r, Z_i = z\,\right) \label{eq:lb1_k} \\
    &+& \Pr\left(n_{x,i} > N c_{d_x}r^{d_x}f_X(x)e^{\varepsilon}-1 \big| \rho_{k,i} = r, Z_i = z\,\right).\label{eq:lb2_k}
\end{eqnarray}

holds for both KSG and BI-KSG estimates. Recall that in Theorem~\ref{thm_bino_tr}, given that $\rho_{k,i} = r$ and $Z_i = z$, $n_{x,i}-k$ is distributed as $\sum_{l=k+1}^{N-1} U_l$, where $U_l$ are i.i.d Bernoulli random variables with mean $p$ satisfying
\begin{eqnarray}
    r^{-d_x} \left|\, p - f_X(x) c_{d_x} r^{d_x} \,\right| \leq C_1 (r^2 + r^{d_y}).
\end{eqnarray}
For small enough $r$ such that $C_1 (r^2 + r^{d_y}) \leq \varepsilon/2$, we obtain
\begin{eqnarray}
    &&\Pr\left(\,n_{x,i} > (N-1)c_{d_x}r^{d_x}f_X(x)e^{\varepsilon} - 1 \big| \rho_{k,i} = r, Z_i = z\,\right) \notag\\
    &=& \Pr\left(\, \sum_{l=k+1}^{N-1} U_l > (N-1)c_{d_x}r^{d_x}f_X(x)e^{\varepsilon} - k - 1\right) \notag\\
    &=& \Pr\left(\, \sum_{l=k+1}^{N-1} U_l - (N-k-1) \E[U_l] > (N-1)c_{d_x}r^{d_x}f_X(x)e^{\varepsilon} - k - 1 - (N-k-1) \E[U_l]\right) \label{eq:lb_nxi}\;,
\end{eqnarray}
and the right-hand side in the probability is lower bounded by
\begin{eqnarray}
&& N c_{d_x}r^{d_x}f_X(x)e^{\varepsilon} - k - 1 - (N-k-1)\E[U_l] \,\notag\\
&\geq& N c_{d_x}r^{d_x}f_X(x)e^{\varepsilon} - k - 1 - (N-k-1)f_X(x) c_{d_x} r^{d_x} (1+\varepsilon/2) \,\notag\\
&\geq& (N-k-1)c_{d_x}r^{d_x}f_X(x)(e^{\varepsilon}-1-\varepsilon/2) - k - 1\,\notag\\
&\geq& (N-k-1)c_{d_x}r^{d_x}f_X(x)\varepsilon/4
\end{eqnarray}
for sufficiently large $N$ such that $ (N-k-1)c_{d_x}r^{d_x}f_X(x)(e^{\varepsilon}-1-\varepsilon/4) > k+1$. Since $U_l$ is Bernoulli, we have $\E[U_l^2] = \E[U_l]$. Now applying Bernstein's inequality,~\eqref{eq:lb_nxi} is upper bounded by:
\begin{eqnarray}
&& \Pr\left(\, \sum_{l=k+1}^{N-1} U_l - (N-k-1)\E[U_l] > (N-1)c_{d_x}r^{d_x}f_X(x)e^{\varepsilon} - k - (N-k-1)\E[U_l]\right) \,\notag\\
&\leq& \Pr\left(\, \sum_{l=k+1}^{N-1} U_l - (N-k-1)\E[U_l] > (N-k-1)c_{d_x}r^{d_x}f_X(x)\varepsilon/4\right) \,\notag\\
&\leq& \exp \Big\{-\frac{((N-k-1)c_{d_x}r^{d_x}f_X(x)\varepsilon/4)^2}{2\left(\, (N-k-1) \E[U_l^2] + \frac{1}{3} ((N-k-1)c_{d_x}r^{d_x}f_X(x)\varepsilon/4)\,\right)}\Big\} \,\notag\\
&\leq& \exp\Big\{-\frac{((N-k-1)c_{d_x}r^{d_x}f_X(x)\varepsilon/4)^2}{2\left(\, (N-k-1) c_{d_x}r^{d_x}f_X(x)(1+\varepsilon/2) + \frac{1}{3} ((N-k-1)c_{d_x}r^{d_x}f_X(x)\varepsilon/4)\,\right)}\Big\} \,\notag\\
&=& \exp \Big\{-\frac{\varepsilon^2}{32(1+7\varepsilon/12)}(N-k-1) c_{d_x}r^{d_x}f_X(x)\Big\}.
\end{eqnarray}
Similarly, the tail bound on the other way is given by:
\begin{eqnarray}
    &&\Pr\left(\,n_{x,i} < N c_{d_x}r^{d_x}f_X(x)e^{-\varepsilon}  \big| \rho_{k,i} = r, Z_i = z\,\right) \notag\\
    &=& \Pr\left(\, \sum_{l=k+1}^{N-1} U_l < N c_{d_x}r^{d_x}f_X(x)e^{-\varepsilon} - k \right) \notag\\
    &=& \Pr\left(\, \sum_{l=k+1}^{N-1} U_l - (N-k-1) \E[U_l] < N c_{d_x}r^{d_x}f_X(x)e^{-\varepsilon} - k - (N-k-1) \E[U_l]\right) \label{eq:ub_nxi}\;,
\end{eqnarray}
and the right hand side in the probability is upper bounded by
\begin{eqnarray}
&& N c_{d_x}r^{d_x}f_X(x)e^{-\varepsilon} - k - (N-k-1)\E[U_l] \,\notag\\
&\leq& N c_{d_x}r^{d_x}f_X(x)e^{-\varepsilon} - k - (N-k-1)f_X(x) c_{d_x} r^{d_x} (1-\varepsilon/2) \,\notag\\
&\leq& (N-k-1)c_{d_x}r^{d_x}f_X(x)e^{-\varepsilon} - (N-k-1)f_X(x) c_{d_x} r^{d_x} (1-\varepsilon/2)\,\notag\\
&=& (N-k-1)c_{d_x}r^{d_x}f_X(x) \left(\, e^{-\varepsilon} - 1 + \varepsilon/2 \,\right)\,\notag\\
&\leq& -(N-k-1)c_{d_x}r^{d_x}f_X(x)\varepsilon/4
\end{eqnarray}
for sufficiently small $r$ such that $ (k+1)c_{d_x}r^{d_x}f_X(x) e^{-\varepsilon} < k$ and sufficiently small $\varepsilon$ such that $e^{-\varepsilon} - 1 + \varepsilon/2 \leq -\varepsilon/4$. Similarly, by applying Bernstein's inequality,~\eqref{eq:ub_nxi} is upper bounded by:
\begin{eqnarray}
&& \Pr\left(\, \sum_{l=k+1}^{N-1} U_l - (N-k-1)\E[U_l] < N c_{d_x}r^{d_x}f_X(x)e^{-\varepsilon} - k - (N-k-1)\E[U_l]\right) \,\notag\\
&\leq& \Pr\left(\, \sum_{l=k+1}^{N-1} U_l - (N-k-1)\E[U_l] > -(N-k-1)c_{d_x}r^{d_x}f_X(x)\varepsilon/4\right) \,\notag\\
&\leq& \exp\Big\{-\frac{((N-k-1)c_{d_x}r^{d_x}f_X(x)\varepsilon/4)^2}{2\left(\, (N-k-1) \E[U_l^2] + \frac{1}{3} ((N-k-1)c_{d_x}r^{d_x}f_X(x)\varepsilon/4)\,\right)}\Big\} \,\notag\\
&\leq& \exp\Big\{-\frac{((N-k-1)c_{d_x}r^{d_x}f_X(x)\varepsilon/4)^2}{2\left(\, (N-k-1) c_{d_x}r^{d_x}f_X(x)(1+\varepsilon/2) + \frac{1}{3} ((N-k-1)c_{d_x}r^{d_x}f_X(x)\varepsilon/4)\,\right)}\Big\} \,\notag\\
&=& \exp\Big\{-\frac{\varepsilon^2}{32(1+7\varepsilon/12)}(N-k-1) c_{d_x}r^{d_x}f_X(x)\Big\}.
\end{eqnarray}

Therefore, $I_3(z)$ is upper bounded by:
\begin{eqnarray}
    I_3(z) &=& \int_{r=(\log{N})^2 (N f_X(x) c_{d_x})^{-\frac{1}{d_x}}}^{\log{N} (N f(z)c_{d_x+d_y})^{-\frac{1}{d_x+d_y}}} \Pr\left(\,\left|\log \widehat{f}_X(X_i) - \log f_X(X_i)\right| > \varepsilon\big| \rho_{k,i} = r, Z_i = z\,\right) f_{\rho_{k,i}}(r) dr \notag \\
    &\leq& \int_{r=(\log{N})^2 (N f_X(x) c_{d_x})^{-\frac{1}{d_x}}}^{\log{N} (N f(z)c_{d_x+d_y})^{-\frac{1}{d_x+d_y}}} 2\exp\Big\{-\frac{\varepsilon^2}{32(1+7\varepsilon/12)}(N-k-1) c_{d_x}r^{d_x}f_X(x)\Big\} f_{\rho_{k,i}}(r) dr \notag \\
    &\leq& 2\exp\Big\{-\frac{\varepsilon^2}{64} N c_{d_x}f_X(x)((\log{N})^2 (N f_X(x) c_{d_x})^{-\frac{1}{d_x}})^{d_x}\Big\}  \notag \\
    &\leq& 2\exp\Big\{-\frac{\varepsilon^2}{64} (\log{N})^{2d_x}\Big\}
\end{eqnarray}
for sufficiently large $N$ such that $(N-k-1)/(1+\frac{7}{12}\varepsilon) > N/2$ and any $d_x \geq 1$. The upper bounds of $I_1(z)$, $I_2(z)$ and $I_3(z)$ are all independent of $z$. Therefore, combine the upper bounds of $I_1(z)$, $I_2(z)$ and $I_3(z)$, we obtain
\begin{eqnarray}
    &&\Pr\left(\,\frac{1}{N} \sum_{i=1}^N \left|\log \widehat{f}_X(X_i) - \log f_X(X_i)\right| > \varepsilon\,\right) \notag\\
    &\leq & N \int (I_1(z) + I_2(z) + I_3(z)) f(z) dz \notag \\
    &= & k N^k \exp\Big\{-\frac{(\log{N})^{d_x+d_y}}{4}\Big\} + \Big(4C' \frac{c_{d_x+d_y}}{c_{d_x}}\Big)^k (\log{N})^{2k(d_x+d_y)} N^{1-\frac{k\,d_y}{d_x}} + 2N\exp\Big\{-\frac{\varepsilon^2}{64} (\log{N})^{2d_x}\Big\}.\nonumber
\end{eqnarray}
If $k > d_y/d_x$ as per our assumption, each of the three terms goes to 0 as $N \to \infty$.

Therefore
\begin{eqnarray}
\lim_{N \to \infty} \Pr\left(\,\frac{1}{N} \sum_{i=1}^N \left|\log \widehat{f}_X(X_i) - \log f_X(X_i)\right| > \varepsilon\,\right) = 0
\end{eqnarray}
Therefore, by combining the convergence of error from sampling and error from density estimation, we obtain that $\widehat{H}(X)$ converges to $H(X)$ in probability.

\section{Proof of Theorem \ref{thm:bias_BI_KSG} on the bias of KSG estimator}
\label{sec:proof_KSG_bias}

We will introduce some notations first. Let $Z = (X,Y)$, $f(x) = f(x,y)$ and $d = d_x + d_y$ for short. Let $B(z,r)$ denote the $d$-dimensional ball centered at $z$ with radius $r$, $B_X(x,r)$ denote the $d_x$-dimensional ball (on $X$ space) centered at $x$ with radius $r$. $P(z,r)$ denotes the probability mass inside $B(z,r)$, i.e., $P(z,r) = \int_{B(z,r)} f(t) dt$. Similarly, $P_X(x,r) = \int_{B_X(x,r)} f_X(t) dt$ denotes the probability mass inside $B_X(z,r)$. Now note that if $\rho_{k,i,\cdot} \leq a_N$, we can write $\iota_{k,i,2}$ and $\iota_{k,i,\infty}$ as:
\begin{eqnarray*}
    \iota_{k,i,\infty} &=& \xi_{k,i,\infty}(X) + \xi_{k,i,\infty}(Y) - \xi_{k,i,\infty}(Z) \\
    \iota_{k,i,2} &=& \xi_{k,i,2}(X) + \xi_{k,i,2}(Y) - \xi_{k,i,2}(Z) \;,
\end{eqnarray*}
where
\begin{eqnarray}
    \xi_{k,i,\infty}(Z) &\equiv& -\psi(k) + \log N + \log{c_{d_x,\infty} c_{d_y,\infty}} + d\, \log \rho_{k,i,\infty} \;, \label{def:hHz_ksg_tr}\\
    \xi_{k,i,\infty}(X) &\equiv& -\psi(n_{x,i,\infty}+1) + \log N + \log c_{d_x,\infty} + d_x \log \rho_{k,i,\infty}\;, \label{def:hHx_ksg_tr}\\
    \xi_{k,i,\infty}(Y) &\equiv& -\psi(n_{y,i,\infty}+1) + \log N + \log c_{d_y,\infty} + d_y \log \rho_{k,i,\infty}.  \label{def:hHy_ksg_tr}
\end{eqnarray}
and
\begin{eqnarray}
    \xi_{k,i,2}(Z) &\equiv& -\psi(k) + \log N + \log{c_{d,2}} + d\, \log \rho_{k,i,2} \;, \label{def:hHz_tr}\\
    \xi_{k,i,2}(X) &\equiv& -\log(n_{x,i,2}) + \log N + \log c_{d_x,2} + d_x \log \rho_{k,i,2}\;, \label{def:hHx_tr}\\
    \xi_{k,i,2}(Y) &\equiv& -\log(n_{y,i,2}) + \log N + \log c_{d_y,2} + d_y \log \rho_{k,i,2}.  \label{def:hHy_tr}
\end{eqnarray}
If $\rho_{k,i,\cdot} > a_N$, just define $\xi_{k,i,\cdot}(X) = \xi_{k,i,\cdot}(Y) = \xi_{k,i,\cdot}(Z) = 0$. Similar as the proof of Theorem \ref{thm:consistent_KSG}, we drop the superscript KSG or BI-KSG and subscript $2$ and $\infty$ for statements that holds for both. Since $\iota_{k,i}$'s  are identically distributed, we have $\E [\widehat{I}(X;Y)] = \E [\iota_{k,1}]$. By triangular inequality, the bias of $\widehat{I}(X;Y)$ can be written as:
\begin{eqnarray}
&& \E \left[\hI(X;Y) \right] - I(X;Y) \,\notag\\
&=& \E \left[\iota_{k,1}\right] - I(X;Y)  \,\notag\\
%&=& \left| \, \E\left[\left(\iota_{k,1}-I(X;Y)\right)\,\mathbb{I}\{\rho_{k,1}\leq a_N\} \right]   - \E\left[ I(X;Y)\,\mathbb{I}\{\rho_{k,1}> a_N\}\right]\, \right| \\
&\leq& \left|\, \E \left[\xi_{k,1}(X)\right] - H(X) \,\right| + \left|\, \E \left[\xi_{k,1}\right] - H(Y) \,\right| + \left|\, \E \left[\xi_{k,1}(Z)\right] - H(Z) \,\right| \,\notag\\
&\leq& \left|\, \E \left[\, \left(\, \xi_{k,1}(X) - H(X)\,\right) \cdot \mathbb{I}\{\rho_{k,1} \leq a_N\} \, \right] \,\right| + \left|\, \E \left[\, \left(\, \xi_{k,1}(X) - H(X)\,\right) \cdot \mathbb{I}\{\rho_{k,1} > a_N\} \, \right] \,\right| \,\notag\\
&& +\, \left|\, \E \left[\, \left(\, \xi_{k,1}(Y) - H(Y)\,\right) \cdot \mathbb{I}\{\rho_{k,1} \leq a_N\} \, \right] \,\right| + \left|\, \E \left[\, \left(\, \xi_{k,1}(Y) - H(Y)\,\right) \cdot \mathbb{I}\{\rho_{k,1} > a_N\} \, \right] \,\right| \,\notag\\
&& +\, \left|\, \E \left[\, \xi_{k,1}(Z) \cdot \mathbb{I}\{\rho_{k,1} \leq a_N\}  - H(Z) \, \right] \,\right| + \left|\, \E \left[\,\xi_{k,1}(Z) \cdot \mathbb{I}\{\rho_{k,1} > a_N\} \, \right] \,\right| \,\notag\\
&=& \left|\, \E \left[\, \left(\, \xi_{k,1}(X) - H(X)\,\right) \cdot \mathbb{I}\{\rho_{k,1} \leq a_N\} \, \right] \,\right| + \left|\, \E \left[\, \left(\, \xi_{k,1}(Y) - H(Y)\,\right) \cdot \mathbb{I}\{\rho_{k,1} \leq a_N\} \, \right] \,\right| \,\notag\\
&&+\, \left|\, \E \left[\, \xi_{k,1}(Z) \cdot \mathbb{I}\{\rho_{k,1} \leq a_N\}  - H(Z) \, \right] \,\right| + \left(\, |H(X)|+|H(Y)| \,\right) \Pr\left(\, \rho_{k,i} > a_N \,\right).  \label{eq:decomposition}
\end{eqnarray}

The probability that $\rho_{k,i} > a_N$ is bounded by the following lemma:

\begin{lemma}
    \label{lem:rho>a_N}
    Under the Assumption  \ref{assumption_rate_mi}.$(c)$ and $(d)$,
    %hypotheses of Theorem \ref{thm:convergence_BI_KSG},
    we have:
    \begin{eqnarray}
        \Pr (\rho_{k,i} > a_N) \leq C \left(\, N^{k-1} \exp\{-C (\log N)^{1+\delta}\} + \left(\, \frac{(\log N)^{1+\delta}}{N} \,\right)^{1/d} \,\right).
    \end{eqnarray}
\end{lemma}

Note $N^{k-1} \exp\{-C (\log N)^{1+\delta}\}$ decays faster than $1/N^c$ for any constant $c$.

Now we consider the bias of $\xi_{k,i}(Z)$, $\xi_{k,i}(X)$ and $\xi_{k,i}(Y)$ when $\rho_{k,i} \leq a_N$. $\xi_{k,1}(Z)$ is local $d$-dimensional Kozachenko-Leonenko entropy estimator~\cite{KL87} . Therefore, by Theorem~\ref{thm:bias_KL}, we obtain:

\begin{eqnarray}
    \E \left[\, \xi_{k,1}(Z) \cdot \mathbb{I}\{\rho_{k,1} \leq a_N\}  - H(Z) \, \right] &\leq& O \left(\, \frac{\left(\, \log N \,\right)^{(1+\delta)(1+1/d)}}{N^{1/d}}\,\right) \;,
\end{eqnarray}

The following lemma establishes the convergence rate for marginal entropy estimator $\xi_{k,1}(X)$.

\begin{lemma}
    \label{lem:Hx_tr}
    Under the Assumption \ref{assumption_rate_mi}.$(c)-(e)$, %hypotheses of Theorem \ref{thm:convergence_BI_KSG},
    the bias of marginal entropy estimator $\xi_{k,1}(X)$ is given by:
    \begin{eqnarray}
       \E \left[\, \left(\, \xi_{k,1}(X) - H(X)\,\right) \cdot \mathbb{I}\{\rho_{k,1} \leq a_N\} \, \right] = O \left(\, \frac{\left(\, \log N \,\right)^{(1+\delta)(1+1/d)}}{N^{1/d}} \,\right).
        \label{eq:convergence_rate_Hx}
    \end{eqnarray}
    for $k \geq d_x/d_y$.
\end{lemma}

 Convergence rate of $\xi_{k,1}(Y)$ is immediate by exchanging $X$ and $Y$ and $k \geq d_x/d_y$. Combining Theorem~\ref{thm:bias_KL}, Lemma~\ref{lem:Hx_tr} and Lemma~\ref{lem:rho>a_N}, we obtain the desired statement.

%Error of rho_{k,i} > a_N
\subsection{Proof of Lemma \ref{lem:rho>a_N}}

For $Z_1 = z$,   the $k$NN distance is larger than $a_N$, i.e. $\rho_{k,1} > a_N$ when at most $k-1$ samples are in $B(z,a_N)$, which gives
\begin{eqnarray}
\Pr\left(\,\rho_{k,1} > a_N \, \big|\, Z_1 = z\,\right) = \sum_{m=0}^{k-1} {N-1 \choose m} P(z,a_N)^m \left(\, 1-P(z,a_N) \,\right)^{N-1-m}.
\end{eqnarray}
Similar to the proof of Theorem~\ref{thm:bias_KL}, we divide the support into two parts as follows,
\begin{eqnarray}
S_1 &=& \{z: \|H_f(z')\| < C_d, \forall z' \in B(z,a_N)\} \,\notag\\
S_2 &=& \{z: \|H_f(z')\| \geq C_d, \textrm{ for some } z' \in B(z,a_N)\} = S_1^C
\end{eqnarray}
We have shown that $\int_{S_2} f(z) dz \leq 2 C_a a_N C_g$ from the proof of Theorem~\ref{thm:bias_KL}. For $z \in S_1$, since $f$ is twice continuously differentiable in $B(z,a_N)$ and $a_N$ vanishes as $N$ grows, $f(z) Vol(B(z,a_N))$ approaches $P(z,a_N)$. Precisely, by Lemma~\ref{lemma_p_xr}, for sufficiently large $N$, we have $P(z,a_N) \geq f(z) c_d a_N^d - C_d a_N^{d+2}$. This provide the following upper bound:
\begin{eqnarray}
\Pr\left(\,\rho_{k,1} > a_N \, \big|\, Z_1 = z \in S_1\,\right) & = & \sum_{m=0}^{k-1} {N-1 \choose m} P(z,a_N)^m \left(\, 1-P(z,a_N) \,\right)^{N-1-m} \,\notag \\
&\leq & \sum_{m=0}^{k-1} N^m \left(\, 1-P(z,a_N) \,\right)^{N-1-m} \,\notag \\
&\leq & k N^{k-1} (1- P(z,a_N))^{N-k-1} \,\notag \\
&\leq & k N^{k-1} \exp\{-(N-k-1)P(z,a_N)\} \,\notag \\
&\leq & k N^{k-1} \exp\{-(N-k-1)f(z) c_d a_N^d + (N-k-1)C_d a_N^{d+2}\} \,\notag \\
&\leq & k N^{k-1} \exp\{-C f(z) (\log(N))^{1+\delta}\} \exp\{\log(N)^{(1+\delta)(1+2/d)}/N^{2/d}\} \,\notag\\
&\leq& k e N^{k-1} \exp\{-C f(z) (\log(N))^{1+\delta}\}.
\end{eqnarray}

The last inequality comes from the fact that $\log(N)^{(1+\delta)(1+2/d)}/N^{2/d} < 1$ for sufficiently large $N$ and $s \geq 1$. For $z \in S_2$, we just use the trivial bound $\Pr\left(\, \rho_{k,1} > a_N \,\big|\, Z_1 = z \in S_2\,\right) \leq 1$. Taking the expectation over $Z_1$,
\begin{eqnarray}
\Pr\left(\,\rho_{k,1} > a_N \, \right) &=& \int_{S_1} f(z) \Pr\left(\,\rho_{k,i} > a_N \, \big|\, Z_i = z\,\right) dz + \int_{S_2} f(z) \Pr\left(\,\rho_{k,i} > a_N \, \big|\, Z_i = z\,\right) dz \,\notag\\
&\leq& k e N^{k-1} \int_{S_1} f(z) \exp\{-C f(z) (\log(N))^{1+\delta}\} dz + \int_{S_2} f(z) dz\,\notag\\
&\leq& k e C_c N^{k-1} \exp\{-C C_0 (\log(N))^{1+\delta}\} + 2 C_a a_N C_g,
\end{eqnarray}
where the last inequality comes from Assumption~\ref{assumption_rate_mi}.$(c)$. We complete the proof by plugging in $a_N = ((\log N)^{1+\delta}/N)^{1/d}$.

% Convergence rate of marginal entropy estimation
\subsection{Proof of Lemma \ref{lem:Hx_tr}}
Define $r_N = (\log N)^2 N^{-1/d_x}$, we can split the bias of $\xi_{k,i}(X)$ into two parts:
\begin{eqnarray}
    && \left|\, \E \left[\, \left(\, \xi_{k,1}(X) - H(X) \,\right) \cdot \mathbb{I}\{\rho_{k,1} \leq a_N\} \, \right]  \,\right| \,\notag\\
    &\leq & \left|\, \E \left[\, \left(\, \xi_{k,1}(X) - H(X) \,\right) \cdot \mathbb{I}\{\rho_{k,1} < r_N\} \, \right]  \,\right| + \left|\, \E \left[\, \left(\, \xi_{k,1}(X) - H(X) \,\right) \cdot \mathbb{I}\{r_N \leq \rho_{k,1} \leq a_N\} \, \right]  \,\right| \;,\label{eq:bias_xi}
\end{eqnarray}
If $\rho_{k,1} < r_N$, recall that $\xi_{k,1}(X) = - h(n_{x,1}) + \log c_{d_x} + \log N + d_x \log \rho_{k,1}$, where $h(x) = \log(x)$ or $\psi(x+1)$. Notice that $k < n_{x,1} < N$, so $0 \leq h(n_{x,1}) \leq 2 \log N$. Therefore, we can bound the first term of~\eqref{eq:bias_xi} by:
\begin{eqnarray}
    && \E \left[\, \left(\, \xi_{k,1}(X) - H(X) \,\right) \cdot \mathbb{I}\{\rho_{k,1} < r_N\} \, \right] \,\notag\\
    &\leq& \E \left[\, \left(\, \log N + \log c_{d_x} + d_x \log \rho_{k,1} - H(X) \,\right) \cdot \mathbb{I}\{\rho_{k,1} < r_N\} \, \right] \,\notag\\
    &\leq& \left(\, \log N + \log c_{d_x} - H(X) \,\right) \Pr \left(\, \rho_{k,1} < r_N \,\right) + d_x \int_0^{r_N} \log r f_{\rho_{k,1}}(r) dr \;,
\end{eqnarray}
where $f_{\rho_{k,1}}(r)$ is the pdf of $\rho_{k,1}$. Similarly, it can be lower bounded by:
\begin{eqnarray}
    && \E \left[\, \left(\, \xi_{k,1}(X) - H(X) \,\right) \cdot \mathbb{I}\{\rho_{k,1} < r_N\} \, \right] \,\notag\\
    &\geq& \E \left[\, \left(\, -\log N + \log c_{d_x} + d_x \log \rho_{k,1} - H(X) \,\right) \cdot \mathbb{I}\{\rho_{k,1} < r_N\} \, \right] \,\notag\\
    &\geq& \left(\, -\log N + \log c_{d_x} - H(X) \,\right) \Pr \left(\, \rho_{k,1} < r_N \,\right) + d_x \int_0^{r_N} \log r f_{\rho_{k,1}}(r) dr \;,
\end{eqnarray}
Therefore, we obtain:
\begin{eqnarray}
     &&\left|\, \E \left[\, \left(\, \xi_{k,1}(X) - H(X) \,\right) \cdot \mathbb{I}\{\rho_{k,1} < r_N\} \, \right]  \,\right| \,\notag\\
     &\leq& \left(\, \log N + \left|\, \log c_{d_x} - H(X) \,\right| \,\right) \Pr \left(\, \rho_{k,1} < r_N \,\right) + d_x \int_0^{r_N} |\log r| f_{\rho_{k,1}}(r) dr  \label{eq:bias_r_N}\;,
\end{eqnarray}

Now we will given an upper bound on the probability $\Pr \left(\, \rho_{k,1} < r \,\right)$ for any $r \leq r_N$. Given that $Z_1 = z$, Let $p_r$ be the probability inside the $\ell_p$ ball centered at $Z_1 = z = (x,y)$ with radius $r$. For sufficiently large $N$, we have
\begin{eqnarray}
    p_r & = &\Pr\left(\,u \in B_Z(z,r)\,\right) \leq \left(\, \sup_{t \in B_Z(z,r)} f(t) \,\right) c_{d} r^{d} \leq C_a c_{d} r^{d}
\end{eqnarray}

Therefore,$\Pr \left(\, \rho_{k,1} < r \,|\, Z_1 = z \,\right)$ is upper bounded by:
\begin{eqnarray}
\Pr \left(\, \rho_{k,1} < r \,|\, Z_1 = z \,\right) \notag & = & \sum_{m=k}^{N-1} {N-1 \choose m} p_r^m (1-p_r)^{N-1-m} \leq \sum_{m=k}^{N-1} N^m p_r^m \,\notag\\
&\leq& \sum_{m=k}^{N-1} (N C_a c_d r^d)^m \leq 2 (N C_a c_d r^d)^k\;, \label{eq:ub_pdf_rho}
\end{eqnarray}
Recall that $r \leq r_N = (\log N)^2 N^{-1/d_x}$, so for sufficiently large $N$, we have $N C_a c_d r^d \leq 1/2$, which gives us the last inequality. Notice that this probability is independent of $z$, therefore, we have $\Pr \left(\, \rho_{k,1} < r \,\right) \leq 2(N C_a c_d r^d)^k$. Plugging in $r_N = (\log N)^2 N^{-1/d_x}$, we obtain:
\begin{eqnarray}
    \Pr \left(\, \rho_{k,1} < r_N \,\right) \leq 2(N C_a c_d (\log N)^{2d} N^{-d/(d_x)})^k = 2 C_a^k c_d^k (\log N)^{2kd} N^{-kd_y/d_x}
\end{eqnarray}
Let $F_{\rho_{k,1}}(r)$ be the CDF of $\rho_{k,1}$ and $F_0(r) = 2(NC_a c_d r^d)^k$ be the upper bound for $F_{\rho_{k,1}}(r)$. Then using integration by parts, the integral $\int_0^{r_N} |\log r| f_{\rho_{k,1}}(r) dr$ can be bounded by:
\begin{eqnarray}
\int_0^{r_N} |\log r| f_{\rho_{k,1}}(r) dr &=& \int_0^{r_N} (-\log r) d F_{\rho_{k,i}}(r) \,\notag\\
&=& -\log (r_N) F_{\rho_{k,i}}(r_N) + \lim_{r \to 0} \left(\, \log (r) F_{\rho_{k,i}(r)} \,\right) - \int_0^{r_N} (-\frac{F_{\rho_{k,i}}(r)}{r}) dr \,\notag\\
&\leq& - \log (r_N) F_0(r_N) + \int_0^{r_N} \frac{F_0(r)}{r} dr \,\notag\\
&=& - 2 \log (r_N) (N C_a c_d r_N^d)^k + \int_0^{r_N} \frac{2(N C_a c_d r^d)^k}{r} dr \,\notag\\
&=& - 2 \log (r_N) (N C_a c_d r_N^d)^k + \frac{2}{kd}(N C_a c_d r_N^d)^k\,\notag\\
&=& \frac{2}{kd} (N C_a c_d)^k r_N^{kd} (1 - kd \log (r_N)) \,\notag\\
&=& \frac{2}{kd} (N C_a c_d)^k (\log N)^{2kd} N^{-\frac{kd}{d_x}} (1 - kd (-\frac{1}{d_x} \log N + 2 \log \log N)) \,\notag\\
&=& \frac{2(C_a c_d)^k}{kd} (\log N)^{2kd} (1 + \frac{kd}{d_x} \log N) N^{-\frac{kd_y}{d_x}}
\end{eqnarray}
If $k \geq d_x/d_y$, then there exists some constant $C$ such that $\Pr \left(\, \rho_{k,1} < r_N \,\right) \leq C (\log N)^{2kd}/N$ and $\int_0^{r_N} |\log r| f_{\rho_{k,1}}(r) dr \leq C (\log N)^{2kd+1}/N$. Therefore, plug it in~\eqref{eq:bias_r_N}, we have:
\begin{eqnarray}
&&\left|\, \E \left[\, \left(\, \xi_{k,1}(X) - H(X) \,\right) \cdot \mathbb{I}\{\rho_{k,1} \leq r_N\} \, \right]  \,\right| \,\notag\\
&\leq& \left(\, \log N + \left|\, \log c_{d_x} - H(X) \,\right| \,\right) C \frac{(\log N)^{2kd}}{N} + d_x C \frac{(\log N)^{2kd+1}}{N} \,\notag\\
&\leq& C(1+d_x) \frac{(\log N)^{2kd+1}}{N} + C \left|\, \log c_{d_x} - H(X) \,\right| \frac{(\log N)^{2kd}}{N} \leq N^{-d_y/d}
\label{eq:smallrbound}
\end{eqnarray}
for sufficiently large $N$.

Now we consider the second term of~\eqref{eq:bias_xi}. Recall that
\begin{eqnarray}
\xi_{k,1,2}(X) &\equiv& \log \left(\, \frac{c_{d_x,2} N \rho_{k,1,2}^{d_x}}{n_{x,1,2}} \, \right) \\
\xi_{k,1,\infty}(X) &\equiv& \log \left(\, \frac{c_{d_x,\infty} N \rho_{k,1,\infty}^{d_x}}{\exp\{\psi(n_{x,1,\infty}+1)\}} \,\right) \;,
\end{eqnarray}
Given that $r_N \leq \rho_{k,1,2} \leq a_N$, the bias of $\xi_{k,1,2}(X)$ is upper bounded by:
\begin{eqnarray}
&& \left|\, \E \left[\, \left(\, \xi_{k,1,2}(X) - H(X) \,\right) \cdot \mathbb{I}\{r_N \leq \rho_{k,1,2} \leq a_N\} \, \right]  \,\right| \,\notag\\
&=& \left| \E_{Z,\rho_{k,1,2}} \left[ \, \E_{n_{x,1,2}} \left[\, \left(\, \xi_{k,1,2}(X) + \int f_X(x) \log f_X(x) dx \,\right) \cdot \mathbb{I}\{ r_N \leq \rho_{k,1,2} \leq a_N\} \,\big|   \, Z, \rho_{k,1,2}\,\right] \, \right] \, \right| \,\notag\\
&\leq& \int \left(\,  \int_{r_N}^{a_N} \left|\,  \log \left(\, f_X(x)c_{d_x,2} N r^{d_x} \,\right) \, -
\E \big[\,\log(n_{x,1,2}) | \rho_{k,1,2} = r, Z_1 = z\,\big]\,\right| f_{\rho_{k,1,2}}(r) dr \,\right) f(z) dz \;,
\end{eqnarray}
where we applied the Jensen's inequality.
By noticing that $\log(x) < \psi(x+1) < \log(x+1)$ for any integer $x \geq 2$, we have $|\psi(x+1)-y| \leq \max_{\theta \in \{0,1\}} |\log(x+\theta) - y|$. So the bias of $\xi_{k,1,\infty}$ is upper bounded by:
\begin{eqnarray}
&& \E \left[\, \left(\, \xi_{k,1,\infty}(X) - H(X) \,\right) \cdot \mathbb{I}\{r_N\leq\rho_{k,1,\infty} \leq a_N\} \, \right]  \,\notag\\
&=& \left| \E_{Z,\rho_{k,1,\infty}} \left[ \, \E_{n_{x,1,\infty}} \left[\, \left(\, \xi_{k,1,\infty}(X) + \int f_X(x) \log f_X(x) dx \,\right) \cdot \mathbb{I}\{r_N\leq\rho_{k,1,\infty} \leq a_N\} \,\big|   \, Z, \rho_{k,1,\infty}\,\right] \, \right] \, \right| \,\notag\\
&\leq& \int \left(\,  \int_{r_N}^{a_N} \left|\, \E \big[\,\psi(n_{x,1,\infty}+1) | \rho_{k,1,\infty} = r, Z_1 = z\,\big] - \log \left(\, f_X(x)c_{d_x,\infty} N r^{d_x} \,\right) \,\right| f_{\rho_{k,1,\infty}}(r) dr \,\right) f(z) dz \,\notag\\
&\leq& \int \left(\,  \int_{r_N}^{a_N} \left|\, \max_{\theta \in \{0,1\}} \E \big[ \,\log(n_{x,1,\infty}+\theta) | \rho_{k,1,\infty} = r, Z_1 = z\,\big] - \log \left(\, f_X(x)c_{d_x,\infty} N r^{d_x} \,\right) \,\right| f_{\rho_{k,1,\infty}}(r) dr \,\right) f(z) dz\;,
\end{eqnarray}

Combine the arguments for KSG and BI-KSG, we obtain:
\begin{eqnarray}
&& \E \left[\, \left(\, \xi_{k,1}(X) - H(X) \,\right) \cdot \mathbb{I}\{r_N\leq \rho_{k,1} \leq a_N\} \, \right]  \,\notag\\
&\leq& \int \left(\,  \int_{r_N}^{a_N} \left|\, \max_{\theta \in \{0,1\}} \E (\,\log(n_{x,1}+\theta) | \rho_{k,1} = r, Z_1 = z\,) - \log \left(\, f_X(x)c_{d_x} N r^{d_x} \,\right) \,\right| f_{\rho_{k,1}}(r) dr \,\right) f(z) dz\;, \label{eq:bias_ksg_biksg}
\end{eqnarray}

\begin{comment}
For a smooth distribution $f_{X}(x)$, the local uniform approximation of the pdf provides a sufficiently accurate estimate of the actual pdf. We use this fact, made precise in Lemma \ref{lemma_p_xr}, to prove
the following key technical lemma, which  characterizes the Binomial distribution of $n_{x,i}$ conditioned on $Z_i = z$ and $\rho_{k,i} = r$. We show that
the deviation of the mean of the binomial from that of the uniform approximation is small even after rescaling by $r^{-d_x}$.
Hence, for sufficiently large $N$, $r$ is sufficiently small such that the rescaled deviation can be made arbitrarily small.
\begin{lemma}~\label{lemma_bino_tr}
Under the assumption \ref{assumption_rate_ent}.(c), and given $Z_1 = z = (x,y)$ and $\rho_{k,1} = r < r_N$ for some deterministic sequence of $r_N$ such that $\lim_{N \to \infty} r_N = 0$, the number of neighbors $n_{x,1}-k$ is distributed as $\sum_{l=k+1}^{N-1} U_l$, where $U_l$ are i.i.d Bernoulli random variables with mean $p$, and there exists a positive constant $C_1$ such that
\begin{eqnarray}
r^{-d_x} \left|\, p - f_X(x) c_{d_x} r^{d_x} \,\right| &\leq& C_1 \, \left(\, r^{2} + r^{d_y} \,\right) \label{eq:lemma_bino_tr}
\end{eqnarray}
for sufficiently large $N$.
\end{lemma}
\end{comment}

From now on we drop the subscript $2$ or $\infty$. Now similar as the proof of ~\ref{thm:bias_KL}, we divide the support of $X$ into two parts:
\begin{eqnarray}
S_1^{(X)} &=& \{x : \|H_{f_x}(x)\| < C_d, \forall x' \in B_X(x,a_N)\} \,\notag\\
S_2^{(X)} &=& \{x: \|H_f(x)\| \geq C_d, \textrm{ for some } x' \in B_X(x,a_N)\} = S_1^C
\end{eqnarray}
where the Lebesgue measure of $S_2^{(X)}$ is upper bounded by $2 C_h a_N$ for sufficiently small $a_N$.
Therefore, we rewrite~\eqref{eq:bias_ksg_biksg} as:
\begin{eqnarray}
&& \E \left[\, \left(\, \xi_{k,1}(X) - H(X) \,\right) \cdot \mathbb{I}\{r_N\leq \rho_{k,1} \leq a_N\} \, \right]  \,\notag\\
&\leq& \int_{S_1} \left(\,  \int_{r_N}^{a_N} \left|\, \max_{\theta \in \{0,1\}} \E (\,\log(n_{x,1}+\theta) | \rho_{k,1} = r, Z_1 = z\,) - \log \left(\, f_X(x)c_{d_x} N r^{d_x} \,\right) \,\right| f_{\rho_{k,1}}(r) dr \,\right) f(z) dz\,\notag\\
&&+ \int_{S_2} \left(\,  \int_{r_N}^{a_N} \left|\, \max_{\theta \in \{0,1\}} \E (\,\log(n_{x,1}+\theta) | \rho_{k,1} = r, Z_1 = z\,) - \log \left(\, f_X(x)c_{d_x} N r^{d_x} \,\right) \,\right| f_{\rho_{k,1}}(r) dr \,\right) f(z) dz \;,
\end{eqnarray}

Recall that in Theorem~\ref{thm_bino_tr}, given that $\rho_{k,i} = r$ and $Z_i = z$, $n_{x,i}-k$ is distributed as $\sum_{l=k+1}^{N-1} U_l$, where $U_l$ are i.i.d Bernoulli random variables with mean $p$ satisfying
\begin{eqnarray}
    r^{-d_x} \left|\, p - f_X(x) c_{d_x} r^{d_x} \,\right| \leq C_1 (r^2 + r^{d_y}).
    \label{eq:bound_p}
\end{eqnarray}
if $x \in S_1^{(X)}$. For $x \in S_2^{(X)}$, the Bernoulli property still holds, but the mean $p$ is simply bounded by
\begin{eqnarray}
r^{-d_x} | p - f_X(x) c_{d_x} r^{d_x}| \leq r^{-d_x} f_X(x) c_{d_x} r^{d_x} \leq C_a c_{d_x}
\end{eqnarray}

From now on, we will focus on $x \in S_1^{(X)}$. For $x \in S_2^{(X)}$, the analyses also hold if we replace $C_1(r^2 + r^{d_y})$ by $C_a c_{d_x}$ everywhere. We will skip that for simplicity.
For $r > r_N = (\log N)^2 N^{-1/d_x}$, we know that $p \geq f_X(x) c_{d_x} r^{d_x}/2 = f_X(x) c_{d_x} (\log N)^{2d_x}/(2N)$ for sufficiently large $N$. Therefore, for any $\theta \in \{0,1\}$, using the Taylor expansion of a logarithm, we obtain:
\begin{eqnarray}
\E \left[\,\log(n_{x,1} + \theta) \,|\, \rho_{k,1} = r, X_1 = x\,\right] &=& \log \left(\, p(N-k-1)+k+\theta \,\right) - \frac{1-p}{2p(N-k-1)} + O\left(\frac{1}{p^2(N-k-1)^2}\right)\;.
\end{eqnarray}
For sufficiently large $N$, this gives

\begin{eqnarray}
&&\left|\, \E (\,\log(n_{x,1}+\theta) \,|\, \rho_{k,1} = r, X_1 = x\,) - \log \left(\, f_X(x)c_{d_x} N r^{d_x} \,\right) \,\right|\,\notag\\
&\leq& \left|\, \log \left(\, p(N-k-1)+k+\theta \,\right) - \log \left(\, f_X(x)c_{d_x} N r^{d_x} \,\right)\,\right| + \frac{1-p}{2p(N-k-1)} + \frac{C_2}{p^2(N-k-1)^2}\,\notag\\
&\leq& \left|\log (pN) - \log \left(\, f_X(x)c_{d_x} N r^{d_x} \,\right)\,\right| + \left|\,\log (pN) - \log \left(\, pN+k(1-p)+\theta-p \,\right)\,\right| \,\notag\\
&&+\, \frac{1-p}{2p(N-k-1)} + \frac{C_2}{p^2(N-k-1)^2} \,\notag\\
&\leq& \left|\log (pN) - \log \left(\, f_X(x)c_{d_x}Nr^{d_x} \,\right)\,\right| + \frac{C_3}{pN} \label{eq:err1} \;,
\end{eqnarray}
For sufficiently large $N$ we have sufficiently small $r$ such that, from Theorem \ref{thm_bino_tr}, we get $p > f_X(x) c_{d_x} r^{d_x}/2$. Therefore the first term in~\eqref{eq:err1} is bounded by:
\begin{eqnarray}
\left|\log (pN) - \log \left(\, f_X(x)c_{d_x} N r^{d_x} \,\right)\,\right| \,
&\leq& \left|\, p - f_X(x) c_{d_x}r^{d_x}\,\right| \, \left(\frac{1}{2p} + \frac{1}{2f_X(x)c_{d_x}r^{d_x}} \right) \,\notag\\
&\leq& C_1 \left(\, r^{d_x+2} + r^{d_x+d_y} \,\right) \frac{3}{2f_X(x)c_{d_x}r^{d_x}} \,\notag\\
&\leq& \frac{3C_1 (r^2 + r^{d_y})}{2 c_{d_x} f_X(x)} \;,
\end{eqnarray}
where we used the fact that $\log x -\log y \leq |x-y| (1/(2x) + 1/(2y))$ for any positive $x$ and $y$ and
the upper bound on $|p-f_X(x) c_{d_x}r^{d_x}|$ from \eqref{eq:bound_p}.
The second term in~\eqref{eq:err1} is bounded by $2C_3/(f_X(x) r^{d_x} N)$, which gives,
for $C_4 = \max\{3C_1/2c_{d_x}, 2C_3\}$,
\begin{eqnarray}
\left|\, \E (\,\log(n_{x,1}+\theta) | \rho_{k,1} = r, X_1 = x\,) - \log \left(\, f_X(x)c_{d_x} N r^{d_x} \,\right) \,\right| \leq \frac{C_4}{f_X(x)} \,\left(\,\frac{1}{r^{d_x} N} + r^2 + r^{d_y}\,\right) \label{eq:e_log_xi}\;.
\end{eqnarray}
To integrate with respect to $\rho_{k,1} = r$, note that $\rho_{k,1}$ is simply the $k^{\rm th}$ order statistic of $N-1$ i.i.d.\ random variables $\big\{\, \|Z_2 - z\|, \|Z_3 - z\|, \dots, \|Z_N - z\| \,\big\}$. The corresponding pdf satisfies  ~\cite{david1970order}:
\begin{eqnarray}
f_{\rho^{(N-1)}_{k,1}}(r) = \frac{N-1}{k-1} f_{\rho^{(N-2)}_{k-1,1}}(r) P(z,r). \label{eq:pdf_rho_k}
\end{eqnarray}
For any $\theta \in \{0,1\}$, we have
\begin{eqnarray}
&&\int_{0}^{a_N} \left|\, \E (\,\log(n_{x,i}+\theta) | \rho_{k,i} = r, Z_i = z\,) - \log \left(\, f_X(x)c_{d_x} N r^{d_x} \,\right) \,\right| f_{\rho^{(N-1)}_{k,i}}(r) dr \,\notag\\
&\leq& C_4 \, \int_{0}^{a_N} \frac{1}{f_X(x)} \left(\, \frac{1}{r^{d_x} N} + r^2 + r^{d_y} \,\right) f_{\rho^{(N-1)}_{k,i}}(r) dr \,\notag\\
&=& C_4 \int_0^{a_N} \frac{(N-1)P(z,r)}{(k-1)f_X(x)} \left(\, \frac{1}{r^{d_x} N} + r^2 + r^{d_y} \,\right) \, f_{\rho^{(N-2)}_{k-1,1}}(r) dr \,\notag\\
&\leq& C_4 \max_{r \leq a_N} \frac{N P(z,r)}{(k-1)f_X(x)} \left(\, \frac{1}{r^{d_x} N} + r^2 + r^{d_y} \,\right)\;. \label{eq:pdf_rho_k1}
\end{eqnarray}
By Lemma \ref{lemma_p_xr}, $|P(z,r) - f(z) c_d r^d| \leq C r^{d+2}$. Therefore, for sufficiently small $a_N$, we have $P(z,r) < 2 f(z) c_d r^d$ for all $r \leq a_N$. Then we have:
\begin{eqnarray}
\max_{r \leq a_N} \frac{NP(z,r)}{(k-1)f_X(x)} \left(\, \frac{1}{r^{d_x} N} + r^2 + r^{d_y} \,\right)
&\leq& \max_{r \leq a_N} \frac{2f(z)c_d r^d N}{(k-1)f_X(x)} \left(\, \frac{1}{r^{d_x} N} + r^2 + r^{d_y} \,\right) \,\notag\\
& = & \max_{r \leq a_N} \frac{2 c_d f_{Y|X}(y|x)}{k-1} \left(\, r^{d_y} + Nr^{d+2} + Nr^{d+d_y}\,\right) \,\notag\\
&\leq& C_5 \left(\, a_N^{d_y} + N a_N^{d+2} + N a_N^{d+d_y} \,\right). \label{eq:pdf_rho_k2}
\end{eqnarray}

Since $f_{Y|X}(y|x)$ is upper bounded by $C_e$, here $C_5$ is given by $C_5 = 2c_d C_e/(k-1)$. The above upper bound holds for $x \in S_1^{(X)}$, while for $x \in S_2^{(X)}$, we have an upper bound of $C_6 (a_N^{d_y} +N a_N^d)$ for some $C_6 > 0$. Now averaging over $z$, we get:
\begin{eqnarray}
&&\E \left[\, \left(\, \xi_{k,1}(X) - H(X) \,\right) \cdot \mathbb{I}\{r_N\leq \rho_{k,1} \leq a_N\} \, \right] \,\notag\\
&\leq& C_4 C_5 \int_{S_1} f(z) \left(\, a_N^{d_y} + Na_N^{d+2} + N a_N^{d+d_y} \,\right) dz +C_4 C_6 \int_{S_2} f(z) \left(\, a_N^{d_y} +N a_N^d \,\right)\,\notag\\
&\leq &C_4 C_5 \, \left(\, a_N^{d_y} + N a_N^{d+2} + N a_N^{d+d_y} \,\right) + C_a C_4 C_6 m(S_2) \left(\, a_N^{d_y} +N a_N^d \,\right)\;.
\end{eqnarray}
here the Lebesgue measure of $S_2$ is upper bounded by $2 C_g a_N$ by Assumption~\ref{assumption_rate_mi}.$(h)$.
Together with Equation  \eqref{eq:smallrbound} and by the choice of $a_N$ in Equation~\eqref{eq:a_N}, the proof is completed.

\section{Proof of Theorem \ref{thm:variance_BI_KSG} on the variance of KSG estimator}
\label{sec:proof_KSG_variance}

Similar as the proof of Theorem~\ref{thm:bias_BI_KSG},  we can write $\iota_{k,i,2}$ and $\iota_{k,i,\infty}$ as:
\begin{eqnarray*}
    \iota_{k,i,\infty} &=& \xi_{k,i,\infty}(X) + \xi_{k,i,\infty}(Y) - \xi_{k,i,\infty}(Z) \;, \\
    \iota_{k,i,2} &=& \xi_{k,i,2}(X) + \xi_{k,i,2}(Y) - \xi_{k,i,2}(Z) \;,
\end{eqnarray*}
where $\xi_{k,i,\infty}$ and $\xi_{k,i,2}$ are defined through~\eqref{def:hHz_ksg_tr} -~\eqref{def:hHy_tr}. Similar as the proof of Theorem~\ref{thm:consistent_KSG} and~\ref{thm:bias_BI_KSG}, we drop the superscript KSG or BI-KSG and subscript $2$ and $\infty$ for statements that holds for both. Consider
\begin{eqnarray}
\widehat{H}(X) &=& \frac{1}{N} \sum_{i=1}^N \xi_{k,i}(X) \;,\quad \widehat{H}(Y) = \frac{1}{N} \sum_{i=1}^N \xi_{k,i}(Y) \;.
\end{eqnarray}
Then $\widehat{I}(X;Y)$ can be rewritten as $\widehat{I}(X;Y) = \widehat{H}(X) + \widehat{H}(Y) - \widehat{H}_{tKL}(Z)$, where $\widehat{H}_{tKL}(Z) = \frac{1}{N} \sum_{i=1}^N \xi_{k,i}(Z)$ is the truncated KL entropy estimator. By Cauchy-Schwarz inequality, we obtain:
\begin{eqnarray}
\V \left[\, \widehat{I}(X;Y) \,\right] &\leq& 3 \left(\, \V \left[\, \widehat{H}_{tKL}(Z) \,\right] + \V \left[\, \widehat{H}(X) \,\right] + \V \left[\, \widehat{H}(Y) \,\right] \,\right)\;.
\end{eqnarray}
From Theorem~\ref{thm:variance_KL}, we know that
\begin{eqnarray}
\V \left[\, \widehat{H}_{tKL}(Z) \,\right] = O \left(\, \frac{\left(\, \log \log N \right)^2 \left(\, \log N \,\right)^{(2k+2)(1+\delta)}}{N}\,\right) \;,
\end{eqnarray}
so we only need to give an upper bound for $\V [\widehat{H}(X)]$ and $\V [\widehat{H}(Y)]$, which use the adaptive choice of $n_{x,i}$ and $n_{y,i}$. The following lemma gives an upper bound for $\V [\widehat{H}(X)]$,

\begin{lemma}
    \label{lem:var_hHx}
    Under the Assumption  \ref{assumption_rate_mi}
    %hypotheses of Theorem \ref{thm:convergence_BI_KSG},
    we have:
    \begin{eqnarray}
        \V \left[\, \widehat{H}(X) \,\right] = O \left(\, \frac{ \left(\, \log N \,\right)^{3+\delta}}{N}\,\right)\;.
    \end{eqnarray}
\end{lemma}

Similarly, we have $\V \left[\, \widehat{H}(Y) \,\right] = O \left(\, \frac{ \left(\, \log N \,\right)^{3+\delta}}{N}\,\right)$. Together with Theorem~\ref{thm:variance_KL}, we obtain the desired statement.

\subsection{Proof of Lemma~\ref{lem:var_hHx}}

Recall that $\widehat{H}(X) = \frac{1}{N} \sum_{i=1}^N \xi_{k,i}(X)$, where $\xi_{k,i}(X)$ are identically distributed, we can rewrite the variance of $\widehat{H}(X)$ as
\begin{eqnarray}
\V \left[\, \widehat{H}(X) \,\right] &=& \V \left[\, \frac{1}{N} \sum_{i=1}^N \xi_{k,i}(X) \,\right]  \,\notag\\
&=& \frac{1}{N^2} \left(\, \sum_{i=1}^N \V \left[\, \xi_{k,i}(X) \,\right] + \sum_{i=1}^N \sum_{j \neq i} \Cov \left[\, \xi_{k,i}(X), \xi_{k,j}(X) \,\right] \,\right)\,\notag\\
&\leq& \frac{1}{N} \V \left[\, \xi_{k,1}(X) \,\right] + \Cov \left[\, \xi_{k,1}(X), \xi_{k,2}(X) \,\right] \;.\label{eq:dec_var_hHx}
\end{eqnarray}

We will consider the variance term and covariance term separately. The following lemma gives an upper bound for $\V \left[\, \xi_{k,1}(X) \,\right]$.

\begin{lemma}
    \label{lem:var_xi}
    Under the Assumption  \ref{assumption_rate_mi}
    %hypotheses of Theorem \ref{thm:convergence_BI_KSG},
    we have:
    \begin{eqnarray}
        \V \left[\, \xi_{k,1}(X) \,\right] = O \left(\, \left(\, \log N \,\right)^{2}\,\right)\;.
    \end{eqnarray}
\end{lemma}

The covariance term is upper bounded by the following lemma.

\begin{lemma}
    \label{lem:cov_xi}
    Under the Assumption  \ref{assumption_rate_mi}
    %hypotheses of Theorem \ref{thm:convergence_BI_KSG},
    we have:
    \begin{eqnarray}
        \Big|\, \Cov \left[\, \xi_{k,1}(X), \xi_{k,2}(X) \,\right] \,\Big|= O \left(\, \frac{\left(\, \log N \,\right)^{3+\delta}}{N}\,\right)\;.
    \end{eqnarray}
\end{lemma}

Combine Lemma~\ref{lem:var_xi} and Lemma~\ref{lem:cov_xi}, we complete the proof.

\subsection{Proof of Lemma~\ref{lem:var_xi}}

Recall that $\xi_{k,1}(X) = - h(n_{x,1}) + \log \left(\, N c_{d_x} \rho_{k,1}^{d_x} \,\right)$, where $h(x) = \log(x)$ or $\psi(x+1)$. Therefore, by Cauchy-Schwarz inequality, we have:
\begin{eqnarray}
\V \left[\, \xi_{k,1}(X) \,\right] &\leq& 2 \left(\, \V \left[\, h(n_{x,1}) \,\right] + \V \left[\, \log \left(\, N c_{d_x} \rho_{k,1}^{d_x} \,\right) \,\right] \,\right) \;.
\end{eqnarray}
Notice that $k < n_{x,1} < N$, so $0 \leq h(n_{x,1}) \leq 2 \log N$. Therefore, $\V \left[\, h(n_{x,1}) \,\right] \leq \E \left[\, \left(\, h(n_{x,i}) \,\right)^2 \,\right] \leq 4 ( \log N )^2$. For $\V \left[\, \log \left(\, N c_{d_x} \rho_{k,1}^{d_x} \,\right) \,\right]$, recall that in~\eqref{eq:ub_pdf_rho} we have shown that $\Pr ( \rho_{k,1} < r ) \leq 2(NC_a c_{d_x} r^{d_x})^k$, therefore, the CDF of $N c_{d_x} \rho_{k,1}^{d_x}$ is upper bounded by $F_{N c_{d_x} \rho_{k,1}^{d_x}}(t) \leq 2(C_a t)^k$. Moreover, since we truncated $\rho_{k,1}$ by $a_N$, so $N c_{d_x} \rho_{k,1}^{d_x} \leq N c_{d_x} a_N^{d_x}$. So the variance is upper bounded by
\begin{eqnarray}
&&\V \left[\, \log \left(\, N c_{d_x} \rho_{k,1}^{d_x} \,\right) \,\right] \leq \E \left[\, \left(\, \log \left(\, N c_{d_x} \rho_{k,1}^{d_x} \,\right) \,\right)^2 \,\right] \,\notag\\
&=& \int_0^{N c_{d_x} a_N^{d_x}} \left(\, \log t \,\right)^2 f_{N c_{d_x} \rho_{k,1}^{d_x}}(t) dt  \,\notag\\
&=& \int_0^{1} \left(\, \log t \,\right)^2 f_{N c_{d_x} \rho_{k,1}^{d_x}}(t) dt + \int_1^{N c_d a_N^{d_x}} \left(\, \log t \,\right)^2 f_{N c_{d_x} \rho_{k,1}^{d_x}}(t) dt \,\notag\\
&=& - \int_0^1 \frac{2 \log t \, F_{N c_d \rho_{k,1}^{d_x}}(t)}{t} dt + \int_1^{N c_d a_N^{d_x}} \left(\, \log t \,\right)^2 f_{N c_{d_x} \rho_{k,1}^{d_x}}(t) dt \,\notag\\
&\leq& -\int_0^1 \frac{4 \log t \,(C_a t)^k}{t} dt + \left(\, \log \left(\, N c_d a_N^{d_x} \,\right) \,\right)^2 \,\notag\\
&=& \frac{4 C_a^k}{k^2} + \left(\, \log \left(\, N c_d a_N^{d_x} \,\right) \,\right)^2 \;.
\end{eqnarray}
By plugging in $a_N = \left(\, \left(\, \log N \,\right)^{1+\delta} / N \,\right)^{1/(d_x+d_y)}$, we obtain that $\V \left[\, \log \left(\, N c_{d_x} \rho_{k,1}^{d_x} \,\right) \,\right] \leq C_1 \left(\, \log N \,\right)^2$ for some $C_1 > 0$. Therefore, we have $\V \left[\, \xi_{k,1}(X) \,\right] = O\left(\, \left(\, \log N \,\right)^2 \,\right)$.

\subsection{Proof of Lemma~\ref{lem:cov_xi}}

First, we decompose the covariance using law of total covariance as
\begin{eqnarray}
&&\Cov \left[\, \xi_{k,1}(X), \xi_{k,2}(X) \,\right]  \,\notag\\
&=& \Cov \left[\, \E \left[\, \xi_{k,1}(X) \,|\, Z_1, Z_2, \rho_{k,1}, \rho_{k,2}\,\right] \,,\, \E \left[\, \xi_{k,2}(X) \,|\, Z_1, Z_2, \rho_{k,1}, \rho_{k,2}\,\right] \,\right] \,\label{eq:cov_2}\\
&+& \E_{Z_1, Z_2, \rho_{k,1}, \rho_{k,2}} \left[\, \Cov \left[\, \xi_{k,1}(X), \xi_{k,2}(X) \,|\, Z_1, Z_2, \rho_{k,1}, \rho_{k,2} \,\right] \,\right] \;,\label{eq:cov_1}
\end{eqnarray}

For~\eqref{eq:cov_2}, we consider two cases.

(1) $\|Z_1 - Z_2\| > 2a_N$, then the two balls $B(Z_1, \rho_{k,1})$ and $B(Z_2, \rho_{k,2})$ are disjoint. Recall that in Theorem~\ref{thm_bino_tr}, we have shown that given $Z_1 = z$ and $\rho_{k,1} = r$, $n_{x,i} - k$ is distributed as $\sum_{l=k+1}^{N-1} U_l$, where $U_l$ are i.i.d. Bernoulli random variable with mean $p$ which only depends on $Z_1$ and $\rho_{k,1}$.Therefore, $\E \left[\, \xi_{k,1}(X) \,|\, Z_1, Z_2, \rho_{k,1}, \rho_{k,2}\,\right] = \E \left[\, \xi_{k,1}(X) \,|\, Z_1, \rho_{k,1} \,\right]$ only depends on $Z_1$ and $\rho_{k,1}$ i.e., only depends on $Z_1$ and its $k$-nearest neighbors. Analogously, $\E \left[\, \xi_{k,2}(X) \,|\, Z_2, \rho_{k,2} \,\right]$ only depends on $Z_2$ and its $k$-nearest neighbors. Since $B(Z_1, \rho_{k,1})$ and $B(Z_2, \rho_{k,2})$ are disjoint, so the two conditional expectations are independent, therefore, have a zero covariance.

(2) $\|Z_1 - Z_2\| \leq 2a_N$. In this case, the covariance is upper bounded by:
\begin{eqnarray}
&&\Big|\, \Cov \left[\, \E \left[\, \xi_{k,1}(X) \,|\, Z_1, Z_2, \rho_{k,1}, \rho_{k,2}\,\right] \,,\, \E \left[\, \xi_{k,2}(X) \,|\, Z_1, Z_2, \rho_{k,1}, \rho_{k,2}\,\right] \,\right] \,\Big|\,\notag\\
&\leq& \sqrt{ \V \left[\, \E \left[\, \xi_{k,1}(X) \,|\, Z_1, Z_2, \rho_{k,1}, \rho_{k,2}\,\right] \,\right] \, \V \left[\, \E \left[\, \xi_{k,2}(X) \,|\, Z_1, Z_2, \rho_{k,1}, \rho_{k,2}\,\right] \,\right]} \,\notag\\
&\leq& \sqrt{\V \left[\, \xi_{k,1}(X) \,\right]\, \V \left[\, \xi_{k,2}(X) \,\right]} \;,
\end{eqnarray}
where we use Cauchy-Schwarz for the first inequality and the fact that conditioning reduces variance for the second inequality. Recall that in Lemma~\ref{lem:var_xi} we have proved that $\V \left[\, \xi_{k,1}(X) \,\right] = O\left(\, (\log N)^2 \,\right)$ and $\xi_{k,2}$ is identically distributed as $\xi_{k,1}(X)$, so the covariance is $O \left(\, (\log N)^2 \,\right)$ in this case.
This case happens with probability
\begin{eqnarray}
\Pr \left[\, \|Z_1 - Z_2\| \leq 2a_N \,\right] &=& \int_x \left(\, \int_{y \in B(x,2a_N)} f(y) \,\right) f(x) dx \,\notag\\
&\leq& \int_x \left(\, C_a c_d (2a_N)^d \,\right) f(x) dx \,\notag\\
&=& C_a c_d (2a_N)^d = 2 C_a c_d \frac{ (\log N)^{1+\delta}}{N} \;.
\end{eqnarray}
Therefore, combine the two cases, we have
\begin{eqnarray}
&&\Big|\, \Cov \left[\, \E \left[\, \xi_{k,1}(X) \,|\, Z_1, Z_2, \rho_{k,1}, \rho_{k,2}\,\right] \,,\, \E \left[\, \xi_{k,2}(X) \,|\, Z_1, Z_2, \rho_{k,1}, \rho_{k,2}\,\right] \,\right] \,\Big| \,\notag\\
&\leq& \Big|\, \Cov \left[\, \E \left[\, \xi_{k,1}(X) \,|\, Z_1, Z_2, \rho_{k,1}, \rho_{k,2}\,\right] \,,\, \E \left[\, \xi_{k,2}(X) \,|\, Z_1, Z_2, \rho_{k,1}, \rho_{k,2}\,\right] \,\Big|\, \|Z_1 - Z_2 \| \leq 2a_N \,\right] \,\Big|\,\notag\\
&&\times \, \Pr \left[\, \|Z_1 - Z_2\| \leq 2a_N \,\right] \,\notag\\
&\leq& C_1 (\log N)^2 \, \Pr \left[\, \|Z_1 - Z_2\| \leq 2a_N \,\right] \,\notag\\
&\leq& 2 C_1 C_a c_d (\log N)^{3+\delta}/N \;,
\end{eqnarray}
for some constant $C_1$.
\\

For~\eqref{eq:cov_1}, recall that $\xi_{k,i}(X) = -h(n_{x,i}) + \log N + \log c_d + d_x \log \rho_{k,i}$ for $i \in \{1,2\}$, here $h(x) = \log(x)$ or $\psi(x+1)$. So given $Z_1$, $Z_2$ and $\rho_{k,1}$, $\rho_{k,2}$, $\Cov \left[\, \xi_{k,1}(X), \xi_{k,2}(X) \,\right] = \Cov \left[\, h(n_{x,1}), h(n_{x,2}) \,\right]$ (we will drop the conditioning on $Z_1$, $Z_2$ and $\rho_{k,1}$, $\rho_{k,2}$ for simplicity). The next step is to identify the joint distribution of $n_{x,1}$ and $n_{x,2}$. Here we consider three cases.
\\

(1) $\|X_1 - X_2\| > 2a_N$, namely the two strips $S_1 = \{x: \|X_1 - x\| \leq a_N\}$ and $S_2 = \{x: \|X_2 - x\| \leq a_N\}$ are disjoint. In this case, similarly to Theorem~\ref{thm_bino_tr}, %\textcolor{blue}
{we can show that $n_{x,1}-k$ and $n_{x,2}-k$ are jointly distributed as multinomial distribution with $N-k-2$ trials and probabilities of $p_1$ and $p_2$, respectively.} Here $p_1$ and $p_2$ are determined by $Z_1$, $Z_2$ and $\rho_{k,1}$, $\rho_{k,2}$.
%Here $p_1$, $p_2$ satisfies $p_i \geq f_X(x_i) c_{d_x} \rho_{k,i}^{d_x}/2$ for $i \in \{1,2\}$.
In order to obtain the covariance of $h(n_{x,1})$ and $h(n_{x,2})$, we use Multivariate Delta Method~\cite{papanicolaou2009taylor} stated as follows:

\begin{lemma}
    \label{lem:multi_delta}
    If $\{X_i\}_{i=1}^{\infty}$ is a sequence of random vectors satisfies $\sqrt{n} \left(\, X_i - \mu \,\right) \stackrel{\mathcal{D}}{\rightarrow} \mathcal{N}(0, \Sigma)$. For a given function $g$ with continuous first partial derivatives, then we have
    \begin{eqnarray}
        \sqrt{n} \left(\, g(X_i) - g(\mu) \,\right) \stackrel{\mathcal{D}}{\rightarrow} \mathcal{N}(0, \nabla g(\mu)^T \, \Sigma \, \nabla g(\mu))\;.
    \end{eqnarray}
\end{lemma}

Since $n_{x,1}$ and $n_{x,2}$ are jointly distributed as multinomial distribution, so we have:
\begin{eqnarray}
\sqrt{N-k-2} \left(\, \left(\, \frac{n_{x,1} - k}{N-k-2} , \frac{n_{x,2} - k}{N-k-2} \,\right) - \mu \,\right) \stackrel{\mathcal{D}}{\rightarrow} \mathcal{N}(0, \Sigma) \;,
\end{eqnarray}
where $\mu = (p_1, p_2)$ and $\Sigma = \begin{pmatrix} p_1(1-p_1) & -p_1p_2 \\ -p_1p_2 & p_2(1-p_2) \end{pmatrix}$. Since $k$ is fixed, we can replace $n_{x,i} - k$ by simply $n_{x,i}$. Now plugging in $g(x_1, x_2) = \left(\, \log (x_1), \log (x_2) \,\right)$ and $\nabla g(x_1, x_2) = \begin{pmatrix} 1/x_1 & 0 \\ 0 & 1/x_2 \end{pmatrix}$, we have:
\begin{eqnarray}
\sqrt{N-k-2} \left(\, \left(\, \log(\frac{n_{x,1}}{N-k-2}) , \log (\frac{n_{x,2}}{N-k-2}) \,\right) - \log (\mu) \,\right) \stackrel{\mathcal{D}}{\rightarrow} \mathcal{N}(0, \nabla g(\mu)^T \, \Sigma \, \nabla g(\mu)) \;,
\end{eqnarray}
here $\nabla g(\mu)^T \, \Sigma \, \nabla g(\mu) =  \begin{pmatrix} (1-p_1)/p_1 & -1 \\ -1 & (1-p_2)/p_2 \end{pmatrix}$.  For a large enough $N$,
\begin{eqnarray}
\Cov \left[\, \log (n_{x,1}), \log (n_{x,2}) \,\right] = \Cov \left[\, \log (\frac{n_{x,1}}{N-k-2}), \log (\frac{n_{x,2}}{N-k-2}) \,\right] = -\frac{1}{N-k-2}
\end{eqnarray}

%\textcolor{red}
{(If $h(x) = \psi(x+1)$, similarly, we can prove that $\Big|\, \Cov \left[\, \psi (n_{x,1}+1), \psi (n_{x,2}+1) \,\right] \,\Big| \leq \frac{2}{N-k-2}$).} Therefore, in this case, $|\, \Cov \left[\, h(n_{x,1}), h(n_{x,2}) \,\right] \,| \leq 3/N$ for sufficiently large $N$.
\\

(2) $\|X_1 - X_2\| \leq 2a_N$ but $\|Z_1 - Z_2\| > 2a_N$, namely the two balls $B_1 = \{z: \|Z_1 - z\| \leq a_N\}$ and $B_2 = \{z: \|Z_2 - z\| \leq a_N\}$ are disjoint, but the two strips $S_1$ and $S_2$ are not. In this case, we can write $n_{x,1} = k + m_1 + m_2$ and $n_{x,2} = k + m_2 + m_3$, here
\begin{itemize}
\item $m_1$ is the number of samples in $R_1 = S_1 \setminus (S_2 \bigcup B_1)$.
\item $m_2$ is the number of samples in $R_2 = (S_1 \setminus B_1) \bigcap (S_2 \setminus B_2)$.
\item $m_3$ is the number of samples in $R_3 = S_2 \setminus (S_1 \bigcup B_2)$.
\end{itemize}

\begin{figure}
	\begin{center}
	\includegraphics[width=0.7\textwidth]{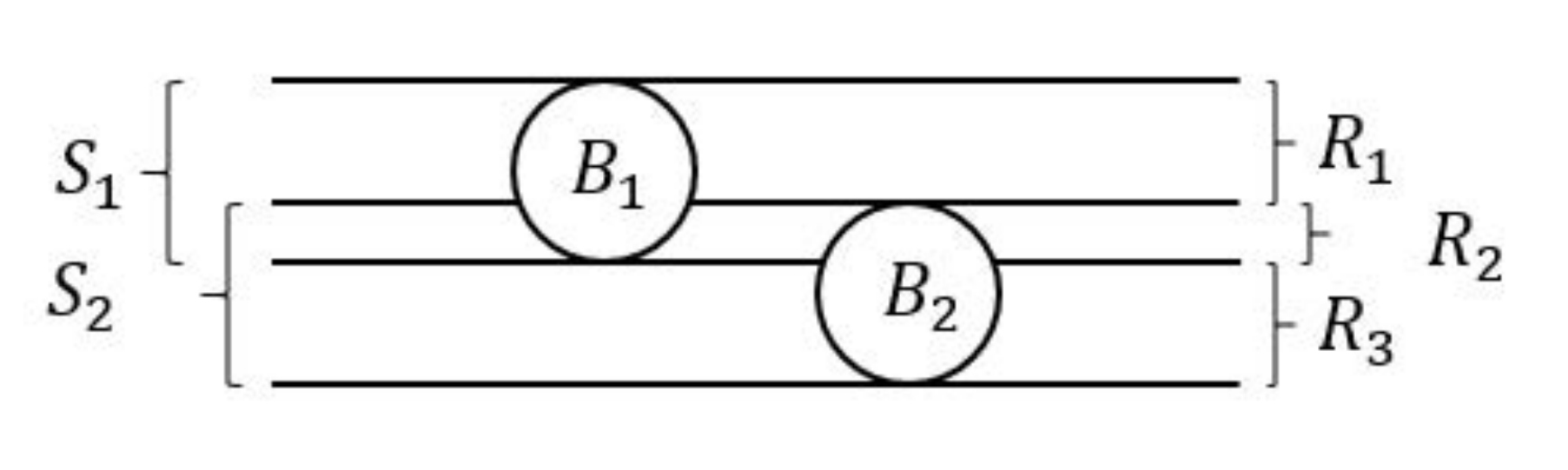}
	\end{center}
	\caption{Regions $R_1$, $R_2$ and $R_3$.}
	\label{fig:regions}
\end{figure}

Figure.~ref{fig:regions} illustrates the positions of regions $R_1$, $R_2$ and $R_3$. Similarly to Theorem~\ref{thm_bino_tr}, we can show that $m_1$, $m_2$ and $m_3$ are jointly distributed as multinomial distribution with $N-k-2$ trials and probabilities of $p_1$, $p_2$ and $p_3$, respectively. The probabilities are determined by $Z_1$, $Z_2$ and $\rho_{k,1}$, $\rho_{k,2}$. Analogously as case 1, here we have:
\begin{eqnarray}
\sqrt{N-k-2} \left(\, \left(\, \frac{n_{x,1} - k}{N-k-2} , \frac{n_{x,2} - k}{N-k-2} \,\right) - \mu \,\right) \stackrel{\mathcal{D}}{\rightarrow} \mathcal{N}(0, \Sigma) \;,
\end{eqnarray}
here $\mu = (p_1 + p_2, p_2 + p_3)$ and $\Sigma = \begin{pmatrix} (p_1+p_2)(1-p_1-p_2) & p_2-(p_1+p_2)(p_2+p_3) \\ p_2-(p_1+p_2)(p_2+p_3) & (p_2+p_3)(1-p_2-p_3) \end{pmatrix}$. Follow the same analysis as case 1, we have:
\begin{eqnarray}
\Cov \left[\, h (n_{x,1}), h (n_{x,2}) \,\right] &=& \frac{p_2 - (p_1+p_2)(p_2+p_3)}{(N-k-2)(p_1+p_2)(p_2+p_3)} \,\notag\\
&=& \frac{p_2}{(N-k-2)(p_1+p_2)(p_2+p_3)} - \frac{1}{N-k-2} \;,\\
\Rightarrow \Big|\, \Cov \left[\, h (n_{x,1}), h (n_{x,2}) \,\right] \,\Big| &\leq& \frac{1}{(N-k-2)(p_1+p_2)} + \frac{1}{N-k-2} \,\notag\\
&\leq& \frac{2}{(N-k-2)(p_1+p_2)} \leq \frac{4}{N(p_1+p_2)} \;,
\end{eqnarray}
for sufficiently large $N$. Notice that $p_1+p_2$ is the probability in $S_1 \setminus B_1$. In Theorem~\ref{thm_bino_tr}, we have shown that $p_1 + p_2 \geq f_X(X_1) c_{d_x} \rho_{k,1}^{d_x}/2$. Therefore, $\big|\, \Cov \left[\, h (n_{x,1}), h (n_{x,2}) \,\right] \,\big| \leq 8/(f_X(X_1) N c_{d_x} \rho_{k,1}^{d_x})$.
\\

(3) $\|Z_1 - Z_2\| \leq 2a_N$, namely the two balls $B_1 = \{z: \|Z_1 - z\| \leq a_N\}$ and $B_2 = \{z: \|Z_2 - z\| \leq a_N\}$ are intersected. In this case, it is hard to identify the joint distribution of $n_{x,1}$ and $n_{x,2}$. But using Cauchy-Schwarz inequality and law of total covariance, we can upper bound the covariance by:
\begin{eqnarray}
&&\Big|\, \Cov \left[\, h (n_{x,1}), h (n_{x,2}) \,|\, Z_1, Z_2, \rho_{k,1}, \rho_{k,2}\,\right] \,\Big| \,\notag\\
&\leq& \sqrt{\V \left[\, h(n_{x,1}) \,|\, Z_1, Z_2, \rho_{k,1}, \rho_{k,2}\,\right] \, \V \left[\, h(n_{x,2}) \,|\, Z_1, Z_2, \rho_{k,1}, \rho_{k,2}\,\right]} \,\notag\\
&\leq& \sqrt{\V \left[\, h(n_{x,1}) \,\right] \, \V \left[\, h(n_{x,2}) \,\right]} \leq C_3 (\log N)^2  \;.
\end{eqnarray}
for some constant $C_3$.
\\

Now combine the three cases. By $E_1$, $E_2$ and $E_3$ we denote the event that case (1), (2) or (3) happens. So
\begin{eqnarray}
&& \Big|\, \E_{Z_1, Z_2, \rho_{k,1}, \rho_{k,2}} \left[\, \Cov \left[\, \xi_{k,1}(X), \xi_{k,2}(X) \,|\, Z_1, Z_2, \rho_{k,1}, \rho_{k,2} \,\right] \,\right] \,\Big|\,\notag\\
&=& \Big|\, \E_{Z_1, Z_2, \rho_{k,1}, \rho_{k,2}} \left[\, \Cov \left[\, h(n_{x,1}), h(n_{x,2}) \,|\, Z_1, Z_2, \rho_{k,1}, \rho_{k,2} \,\right] \,\right] \,\Big|\,\notag\\
&\leq& \Big|\, \E \left[\, \E_{\rho_{k,1}, \rho_{k,2}} \left[\, \Cov \left[\, h(n_{x,1}), h(n_{x,2}) \,|\, Z_1, Z_2, \rho_{k,1}, \rho_{k,2} \,\right] \,\right] \,\Big|\, E_1 \,\right] \,\Big| \times \Pr [E_1] \,\label{eq:cov_term1}\\
&+& \Big|\, \E \left[\, \E_{\rho_{k,1}, \rho_{k,2}} \left[\, \Cov \left[\, h(n_{x,1}), h(n_{x,2}) \,|\, Z_1, Z_2, \rho_{k,1}, \rho_{k,2} \,\right] \,\right] \,\Big|\, E_2 \,\right] \,\Big| \times \Pr [E_2] \,\label{eq:cov_term2}\\
&+& \Big|\, \E \left[\, \E_{\rho_{k,1}, \rho_{k,2}} \left[\, \Cov \left[\, h(n_{x,1}), h(n_{x,2}) \,|\, Z_1, Z_2, \rho_{k,1}, \rho_{k,2} \,\right] \,\right] \,\Big|\, E_3 \,\right] \,\Big| \times \Pr [E_3] \,\label{eq:cov_term3}
\end{eqnarray}

We will deal with the three terms separately as follows.
\begin{enumerate}
\item For~\eqref{eq:cov_term1}, we use the upper bound $\Cov \left[\, h(n_{x,1}), h(n_{x,2}) \,|\, Z_1, Z_2, \rho_{k,1}, \rho_{k,2} \,\right] \leq 3/N$ and $\Pr [E_1] \leq 1$. So~\eqref{eq:cov_term1} is upper bounded by $3/N$.

\item For~\eqref{eq:cov_term2}, the inner expectation is upper bounded by:
\begin{eqnarray}
\E_{\rho_{k,1}, \rho_{k,2}} \left[\, \Cov \left[\, h(n_{x,1}), h(n_{x,2}) \,|\, Z_1, Z_2, \rho_{k,1}, \rho_{k,2} \,\right] \,\right] \leq \E \left[\, \frac{8}{f_X(X_1) N c_{d_x} \rho_{k,1}^{d_x}} \,\right] \;,
\end{eqnarray}
Recall that the pdf of $\rho_{k,1}$ is given in~\eqref{eq:pdf_rho_k}. Following the same analysis as~\eqref{eq:pdf_rho_k1} and~\eqref{eq:pdf_rho_k2}, we obtain:
\begin{eqnarray}
\E \left[\, \frac{8}{f_X(X_1) N c_{d_x} \rho_{k,1}^{d_x}} \,\right] \leq C_4 a_N^{d_y} \;,
\end{eqnarray}
for some constant $C_4 > 0$. Moreover, the probability of $E_2$ is upper bounded by:
\begin{eqnarray}
\Pr [E_2] &\leq& \Pr\left[\, \|X_1 - X_2\| \leq 2a_N \,\right] \,\notag\\
&=& \int_x f_X(x) \left(\, \int_{y \in B_X(x,2a_N)} f_X(y) dy \,\right) dx \,\notag\\
&\leq& \int_x f_X(x) C_f c_{d_x} (2a_N)^{d_x} dx \leq C_f c_{d_x} (2a_N)^{d_x}
\end{eqnarray}
Therefore,~\eqref{eq:cov_term2} is upper bounded by $C_5 a_N^{d_x+d_y}$ for some constant $C_5 > 0$. By plugging in the choice $a_N = \left(\, (\log N)^{1+\delta}/N \,\right)^{1/(d_x+d_y)}$,~\eqref{eq:cov_term2} is upper bounded by $C_5 (\log N)^{1+\delta}/N$.

\item For~\eqref{eq:cov_term3}, the expected covariance is upper bounded by $O\left(\, (\log N)^2 \,\right)$. The probability of $E_3$ is upper bounded by:
\begin{eqnarray}
\Pr [E_3] &=& \Pr\left[\, \|Z_1 - Z_2\| \leq 2a_N \,\right] \,\notag\\
&=& \int_z f(z) \left(\, \int_{t \in B(z,2a_N)} f(t) dt \,\right) dz \,\notag\\
&\leq& \int_z f(z) C_a c_{d_x+d_y} (2a_N)^{d_x+d_y} dz \leq C_a c_{d_x+d_y} (2a_N)^{d_x+d_y}
\end{eqnarray}
By plugging in $a_N$, $\Pr[E_3] \leq C_6 (\log N)^{1+\delta}/N$ for some constant $C_6 > 0$. Therefore,~\eqref{eq:cov_term3} is upper bounded by $C_6 (\log N)^{3+\delta} / N $.
\end{enumerate}

Combine the three cases and analysis of~\eqref{eq:cov_2}, we obtain the desired statement.

\section{Proof of Theorem~\ref{thm_bino_tr}}
\label{sec:proof_bino_tr}

\begin{comment}
We restate Theorem~3 from the main text for ease of reference.
\begin{theorem}~\label{thm_bino_tr}
Given $(X_i, Y_i) = (x,y)$ that satisfies Assumption \ref{assumption_rate_ent}.(d) and $\rho_{k,i,2} = r < r_N$ for some deterministic sequence of $r_N$ such that $\lim_{N \to \infty} r_N = 0$, the number of neighbors $n_{x,i,2}-k$ is distributed as $\sum_{l=k+1}^{N-1} U_l$, where $U_l$ are i.i.d.\  Bernoulli random variables with mean $p$, and there exists a positive constant $C_1$ such that
\begin{eqnarray}
r^{-d_x} \left|\, p - f_X(x) c_{{d_x},2} r^{d_x} \,\right| &\leq& C_1 \, \left(\, r^{2} + r^{d_y} \,\right), \label{eq:thm_bino_tr}
\end{eqnarray}
for sufficiently large $N$.
\end{theorem}
\end{comment}

Given that $Z_1 = z = (x,y)$ and $\rho_{k,1} = r $, let $\{2, 3, \dots, N\} = S \cup \{j\} \cup T$ be a partition of the indices with $\left|S\right| = k-1$ and $\left|T\right| = N-k-1$.  Define an event $\mathcal{A}_{S, j, T}$ associated to the partition as:
\begin{eqnarray}
    \mathcal{A}_{S, j, T} = \big\{\, \|Z_s - z\| < \|Z_j - z\|, \forall s \in S,  \textrm{ and } \|Z_t - z\| > \|Z_j - z\|,\forall t \in T \,\big\}.
\end{eqnarray}
Since $Z_j - z$ are i.i.d.\ random variables each of the events $\mathcal{A}_{S, j, T}$ has identical probability. The number of all partitions is $\frac{(N-1)!}{(N-k-1)!(k-1)!}$ and thus  $\Pr\left(\,\mathcal{A}_{S, j, T}\,\right) = \frac{(N-k-1)!(k-1)!}{(N-1)!}$. So the cdf of $n_{x, i}$ is given by:
\begin{eqnarray}
  \Pr\left(\,n_{x,1} \leq k+m\big|\rho_{k,1} = r, Z_1 = z\,\right)
    &= &\sum_{S,j,T} \Pr\left(\,\mathcal{A}_{S, j, T}\,\right) \Pr\left(\,n_{x,1} \leq k+m\big|\mathcal{A}_{S, j, T}, \rho_{k,1} = r, Z_1 = z\,\right) \notag \\
    &= &\frac{(N-k-1)!(k-1)!}{(N-1)!} \sum_{S,j,T} \Pr\left(\,n_{x,i} \leq k+m\big|\mathcal{A}_{S, j, T}, \rho_{k,1} = r, Z_1 = z\,\right)
\end{eqnarray}

Now condition on event $\mathcal{A}_{S,j,T}$ and $\rho_{k,1} = r$, namely $Z_j$ is the $k$-nearest neighbor with distance $r$, $S$ is the set of samples with distance smaller than $r$ and $T$ is the set of samples with distance greater than $r$. Recall that $n_{x,1}$ is the number of samples with $\|X_j-x\| < r$. For any index $s \in S \cup \{j\}$, $\|X_s-x\| < r$ is satisfied. Therefore, $n_{x,1} \leq k+m$ means that there are no more than $m$ samples in $T$ with $X$-distance smaller than $r$. Let $U_l = \mathbb{I}\{\|X_l - x\| < r \big| \|Z_l - z\| > r\}. $Therefore,
\begin{eqnarray}
    &&\Pr\left(\,n_{x,1} \leq k+m\big|\mathcal{A}_{S, j, T}, \rho_{k,1} = r, Z_1 = z\,\right) \notag \\
    &=& \Pr\left(\,\sum_{t \in T} \mathbb{I}\{\|X_t - x\| < r\} \leq m\big|~\|Z_s - z\| < r, \forall s \in S, \|Z_j - z\| = r, \|Z_t - z\| > r,\forall t \in T, Z_i = z\,\right) \notag \\
    &=& \Pr\left(\,\sum_{t \in T} \mathbb{I}\{\|X_t - x\| < r\} \leq m\big|~ \|Z_t - z\| > r,\forall t \in T\,\right) = \Pr \left(\, \sum_{l=k+1}^{N-1} U_l \leq m\,\right).
\end{eqnarray}

We can drop the conditioning of $Z_s$'s for $s \not\in T$ since $Z_s$ and $X_t$ are independent. Therefore, given that $\|Z_t-z\| > r$ for all $t \in T$, the variables $\mathbb{I}\{\|X_t - x\| < r\}$ are i.i.d.\ and have the same distribution as $U_l$. We conclude:
\begin{eqnarray}
    \Pr\left(\,n_{x,1} \leq k+m\big|\rho_{k,1} = r, Z_1 = z\,\right)
    &=& \frac{(N-k-1)!(k-1)!}{(N-1)!} \sum_{S,j,T} \Pr\left(\,n_{x,i} \leq k+m\big|\mathcal{A}_{S, j, T}, \rho_{k,i} = r, Z_i = z\,\right) \notag \\
    &=& \frac{(N-k-1)!(k-1)!}{(N-1)!} \sum_{S,j,T}  \Pr \left(\, \sum_{l=k+1}^N U_l \leq m\,\right)
    =  \Pr \left(\, \sum_{l=k+1}^{N-1} U_l \leq m\,\right).
\end{eqnarray}
Thus we have shown that $n_{x,i}-k$ has the same distribution as $\sum_{l=k+1}^{N-1} U_l$ given $Z_i = z$ and $\rho_{k,i} = r$, in other words is a Binomial random variable.

Now we bound  the mean of $U_l$:
\begin{eqnarray}
    p = \E [U_l] = \Pr\left(\,\|X_l - x\| < r\big|~\|Z_l - z\| > r\,\right) = \frac{P_X(x,r) - P(z,r)}{1 - P(z,r)} .
\end{eqnarray}
By Lemma ~\ref{lemma_p_xr}, we have:
\begin{eqnarray}
\left|\, P_X(x,r) - f_X(x) c_{d_x} r^{d_x} \,\right| \leq C  r^{d_x+2}.
\end{eqnarray}
and
\begin{eqnarray}
\left|\, P(z,r) - f(z) c_{d} r^{d} \,\right| \leq  r^{d+2}.
\end{eqnarray}
Therefore, the difference of $p$ and $f_X(x)c_{d_x}r^{d_x}$ is bounded by:
\begin{eqnarray}
\left|\, p - f_X(x) c_{d_x} r^{d_x}\,\right|
&\leq& \left|\,\frac{P_X(x,r) - P(z,r)}{1 - P(z,r)}  - P_X(x,r)\,\right| + \left|\, P_X(x,r) - f_X(x) c_{d_x} r^{d_x} \,\right| \,\notag\\
&\leq& \frac{P(z,r)(1-P_X(x,r))}{1-P(z,r)} + \left|\, P_X(x,r) - f_X(x) c_{d_x} r^{d_x} \,\right| \,\notag\\
&\leq& P(z,r) + C r^{d_x+2}
\leq C \left(\, r^{d_x+2} + r^{d_x+d_y}\,\right).
\end{eqnarray}

\section{Acknowledgement}
The authors thank Sreeram Kannan for introducing the KSG estimator to them, Yihong Wu for many helpful discussions, and anonymous reviewers for their constructive feedback. 

%------------------------------------------

\bibliographystyle{plain}
\small
\bibliography{umi}

\begin{thebibliography}{10}

\bibitem{acharya2014estimating}
J.~Acharya, A.~Orlitsky, A.~T. Suresh, and H.~Tyagi.
\newblock Estimating renyi entropy of discrete distributions.
\newblock {\em arXiv preprint arXiv:1408.1000}, 2014.

\bibitem{ahmad1976nonparametric}
I.~A. Ahmad and P.~Lin.
\newblock A nonparametric estimation of the entropy for absolutely continuous
  distributions (corresp.).
\newblock {\em Information Theory, IEEE Transactions on}, 22(3):372--375, 1976.

\bibitem{battiti1994using}
R.~Battiti.
\newblock Using mutual information for selecting features in supervised neural
  net learning.
\newblock {\em Neural Networks, IEEE Transactions on}, 5(4):537--550, 1994.

\bibitem{batu2000testing}
T.~Batu, L.~Fortnow, R.~Rubinfeld, W.~D. Smith, and P.~White.
\newblock Testing that distributions are close.
\newblock In {\em Foundations of Computer Science, 2000. Proceedings. 41st
  Annual Symposium on}, pages 259--269. IEEE, 2000.

\bibitem{biau2011weighted}
G.~Biau, F.~Chazal, D.~Cohen-Steiner, L.~Devroye, and C.~Rodriguez.
\newblock A weighted k-nearest neighbor density estimate for geometric
  inference.
\newblock {\em Electronic Journal of Statistics}, 5:204--237, 2011.

\bibitem{bickel2009springer}
P~Bickel, P~Diggle, S~Fienberg, U~Gather, I~Olkin, and S~Zeger.
\newblock Springer series in statistics.
\newblock 2009.

\bibitem{birge1995estimation}
L.~Birg{\'e} and P.~Massart.
\newblock Estimation of integral functionals of a density.
\newblock {\em The Annals of Statistics}, pages 11--29, 1995.

\bibitem{chan2015multivariate}
C.~Chan, A.~Al-Bashabsheh, J.~B. Ebrahimi, T.~Kaced, and T.~Liu.
\newblock Multivariate mutual information inspired by secret-key agreement.
\newblock {\em Proceedings of the IEEE}, 103(10):1883--1913, 2015.

\bibitem{chanclustering}
C~Chan, A~Al-Bashabsheh, T~Kaced, Q~Zhou, and T~Liu.
\newblock Clustering of random variables by multivariate mutual information.
\newblock {\em submitted to IEEE Transactions on Information Theory.[Online].
  Available: http://bit. ly/1CawCYo}.

\bibitem{cover1991information}
T.~M. Cover and J.~A. Thomas.
\newblock Information theory and statistics.
\newblock {\em Elements of Information Theory}, pages 279--335, 1991.

\bibitem{dasgupta2014optimal}
S.~Dasgupta and S.~Kpotufe.
\newblock Optimal rates for k-nn density and mode estimation.
\newblock In {\em Advances in Neural Information Processing Systems}, pages
  2555--2563, 2014.

\bibitem{david1970order}
H.~A. David and H.~N. Nagaraja.
\newblock {\em Order statistics}.
\newblock Wiley Online Library, 1970.

\bibitem{eggermont1999best}
P.~PB. Eggermont and V.~N. LaRiccia.
\newblock Best asymptotic normality of the kernel density entropy estimator for
  smooth densities.
\newblock {\em Information Theory, IEEE Transactions on}, 45(4):1321--1326,
  1999.

\bibitem{fix1951discriminatory}
E.~Fix and J.~L. Hodges~Jr.
\newblock Discriminatory analysis-nonparametric discrimination: consistency
  properties.
\newblock Technical report, DTIC Document, 1951.

\bibitem{fleuret2004fast}
F.~Fleuret.
\newblock Fast binary feature selection with conditional mutual information.
\newblock {\em The Journal of Machine Learning Research}, 5:1531--1555, 2004.

\bibitem{gao2014efficient}
S.~Gao, G.~Ver~Steeg, and A.~Galstyan.
\newblock Efficient estimation of mutual information for strongly dependent
  variables.
\newblock {\em arXiv preprint arXiv:1411.2003}, 2014.

\bibitem{gao2015estimating}
S.~Gao, G~Ver~Steeg, and A.~Galstyan.
\newblock Estimating mutual information by local gaussian approximation.
\newblock {\em arXiv preprint arXiv:1508.00536}, 2015.

\bibitem{hall1993estimation}
P.~Hall and S.~C. Morton.
\newblock On the estimation of entropy.
\newblock {\em Annals of the Institute of Statistical Mathematics},
  45(1):69--88, 1993.

\bibitem{hjort1996locally}
N.~L. Hjort and M.~C. Jones.
\newblock Locally parametric nonparametric density estimation.
\newblock {\em The Annals of Statistics}, pages 1619--1647, 1996.

\bibitem{Janzing13}
D.~Janzing, D.~Balduzzi, M.~Grosse-Wentrup, and B.~Sch{\"o}lkopf.
\newblock Quantifying causal influences.
\newblock {\em The Annals of Statistics}, 41(5):2324--2358, 2013.

\bibitem{jiao2015justification}
J.~Jiao, T.~A. Courtade, K.~Venkat, and T.~Weissman.
\newblock Justification of logarithmic loss via the benefit of side
  information.
\newblock {\em Information Theory, IEEE Transactions on}, 61(10):5357--5365,
  2015.

\bibitem{jiao2015minimax}
J.~Jiao, K.~Venkat, Y.~Han, and T.~Weissman.
\newblock Minimax estimation of functionals of discrete distributions.
\newblock {\em Information Theory, IEEE Transactions on}, 61(5):2835--2885,
  2015.

\bibitem{joe1989estimation}
H.~Joe.
\newblock Estimation of entropy and other functionals of a multivariate
  density.
\newblock {\em Annals of the Institute of Statistical Mathematics},
  41(4):683--697, 1989.

\bibitem{kandasamy2015nonparametric}
K.~Kandasamy, A.~Krishnamurthy, B.~Poczos, and L.~Wasserman.
\newblock Nonparametric von mises estimators for entropies, divergences and
  mutual informations.
\newblock In {\em Advances in Neural Information Processing Systems}, pages
  397--405, 2015.

\bibitem{khan2007relative}
S.~Khan, S.~Bandyopadhyay, A.~R. Ganguly, S.~Saigal, D.~J. Erickson~III,
  V.~Protopopescu, and G.~Ostrouchov.
\newblock Relative performance of mutual information estimation methods for
  quantifying the dependence among short and noisy data.
\newblock {\em Physical Review E}, 76(2):026209, 2007.

\bibitem{kinney2014equitability}
J.~B. Kinney and G.~S. Atwal.
\newblock Equitability, mutual information, and the maximal information
  coefficient.
\newblock {\em Proceedings of the National Academy of Sciences},
  111(9):3354--3359, 2014.

\bibitem{KL87}
LF~Kozachenko and Nikolai~N Leonenko.
\newblock Sample estimate of the entropy of a random vector.
\newblock {\em Problemy Peredachi Informatsii}, 23(2):9--16, 1987.

\bibitem{kpotufe2011pruning}
S.~Kpotufe and U.~von Luxburg.
\newblock Pruning nearest neighbor cluster trees.
\newblock {\em arXiv preprint arXiv:1105.0540}, 2011.

\bibitem{Kra04}
A.~Kraskov, H.~St{\"o}gbauer, and P.~Grassberger.
\newblock Estimating mutual information.
\newblock {\em Physical review E}, 69(6):066138, 2004.

\bibitem{krishnamurthy2014nonparametric}
A.~Krishnamurthy, K.~Kandasamy, B.~Poczos, and L.~Wasserman.
\newblock Nonparametric estimation of renyi divergence and friends.
\newblock {\em arXiv preprint arXiv:1402.2966}, 2014.

\bibitem{liu2012exponential}
H.~Liu, L.~Wasserman, and J.~D. Lafferty.
\newblock Exponential concentration for mutual information estimation with
  application to forests.
\newblock In {\em Advances in Neural Information Processing Systems}, pages
  2537--2545, 2012.

\bibitem{loader1996local}
C.~R. Loader.
\newblock Local likelihood density estimation.
\newblock {\em The Annals of Statistics}, 24(4):1602--1618, 1996.

\bibitem{loftsgaarden1965nonparametric}
D.~O. Loftsgaarden and C.~P. Quesenberry.
\newblock A nonparametric estimate of a multivariate density function.
\newblock {\em The Annals of Mathematical Statistics}, 36(3):1049--1051, 1965.

\bibitem{lombardi2016nonparametric}
D.~Lombardi and S.~Pant.
\newblock Nonparametric k-nearest-neighbor entropy estimator.
\newblock {\em Physical Review E}, 93(1):013310, 2016.

\bibitem{DBLP:journals/corr/MakkuvaW16}
A.~V. Makkuva and Y.~Wu.
\newblock On additive-combinatorial affine inequalities for shannon entropy and
  differential entropy.
\newblock {\em CoRR}, abs/1601.07498, 2016.

\bibitem{moon2014ensemble}
K.~R. Moon and A.~O. Hero.
\newblock Ensemble estimation of multivariate f-divergence.
\newblock In {\em Information Theory (ISIT), 2014 IEEE International Symposium
  on}, pages 356--360. IEEE, 2014.

\bibitem{muller2012information}
A.~C. M{\"u}ller, S.~Nowozin, and C.~H. Lampert.
\newblock {\em Information theoretic clustering using minimum spanning trees}.
\newblock Springer, 2012.

\bibitem{nguyen2010estimating}
X.~Nguyen, M.~J. Wainwright, and M.~I. Jordan.
\newblock Estimating divergence functionals and the likelihood ratio by convex
  risk minimization.
\newblock {\em Information Theory, IEEE Transactions on}, 56(11):5847--5861,
  2010.

\bibitem{nilsson2007estimation}
M.~Nilsson and W.~B. Kleijn.
\newblock On the estimation of differential entropy from data located on
  embedded manifolds.
\newblock {\em Information Theory, IEEE Transactions on}, 53(7):2330--2341,
  2007.

\bibitem{mcbook}
A.~B. Owen.
\newblock {\em Monte Carlo theory, methods and examples}.
\newblock 2013.

\bibitem{pal2010estimation}
D.~P{\'a}l, B.~P{\'o}czos, and C.~Szepesv{\'a}ri.
\newblock Estimation of r{\'e}nyi entropy and mutual information based on
  generalized nearest-neighbor graphs.
\newblock In {\em Advances in Neural Information Processing Systems}, pages
  1849--1857, 2010.

\bibitem{paninski2003estimation}
L.~Paninski.
\newblock Estimation of entropy and mutual information.
\newblock {\em Neural computation}, 15(6):1191--1253, 2003.

\bibitem{paninski2004estimating}
L.~Paninski.
\newblock Estimating entropy on m bins given fewer than m samples.
\newblock {\em Information Theory, IEEE Transactions on}, 50(9):2200--2203,
  2004.

\bibitem{paninski2008undersmoothed}
L.~Paninski and M.~Yajima.
\newblock Undersmoothed kernel entropy estimators.
\newblock {\em Information Theory, IEEE Transactions on}, 54(9):4384--4388,
  2008.

\bibitem{papanicolaou2009taylor}
Alex Papanicolaou.
\newblock Taylor approximation and the delta method, 2009.

\bibitem{peng2005feature}
H.~Peng, F.~Long, and C.~Ding.
\newblock Feature selection based on mutual information criteria of
  max-dependency, max-relevance, and min-redundancy.
\newblock {\em Pattern Analysis and Machine Intelligence, IEEE Transactions
  on}, 27(8):1226--1238, 2005.

\bibitem{perez2009estimation}
F.~P{\'e}rez-Cruz.
\newblock Estimation of information theoretic measures for continuous random
  variables.
\newblock In {\em Advances in neural information processing systems}, pages
  1257--1264, 2009.

\bibitem{silverman1986density}
B.~W. Silverman.
\newblock {\em Density estimation for statistics and data analysis}, volume~26.
\newblock CRC press, 1986.

\bibitem{singh2003nearest}
H.~Singh, N.~Misra, V.~Hnizdo, A.~Fedorowicz, and E.~Demchuk.
\newblock Nearest neighbor estimates of entropy.
\newblock {\em American journal of mathematical and management sciences},
  23(3-4):301--321, 2003.

\bibitem{singh2014exponential}
S.~Singh and B.~P{\'o}czos.
\newblock Exponential concentration of a density functional estimator.
\newblock In {\em Advances in Neural Information Processing Systems}, pages
  3032--3040, 2014.

\bibitem{singh2014generalized}
S.~Singh and B.~P{\'o}czos.
\newblock Generalized exponential concentration inequality for r{\'e}nyi
  divergence estimation.
\newblock 2014.

\bibitem{sricharan2013ensemble}
K.~Sricharan, D.~Wei, and A.~O. Hero.
\newblock Ensemble estimators for multivariate entropy estimation.
\newblock {\em Information Theory, IEEE Transactions on}, 59(7):4374--4388,
  2013.

\bibitem{tsybakov1996root}
A.~B. Tsybakov and E.~C. Van~der Meulen.
\newblock Root-n consistent estimators of entropy for densities with unbounded
  support.
\newblock {\em Scandinavian Journal of Statistics}, pages 75--83, 1996.

\bibitem{turney2002thumbs}
P.~D. Turney.
\newblock Thumbs up or thumbs down?: semantic orientation applied to
  unsupervised classification of reviews.
\newblock In {\em Proceedings of the 40th annual meeting on association for
  computational linguistics}, pages 417--424. Association for Computational
  Linguistics, 2002.

\bibitem{valiant2011estimating}
G.~Valiant and P.~Valiant.
\newblock Estimating the unseen: an n/log (n)-sample estimator for entropy and
  support size, shown optimal via new clts.
\newblock In {\em Proceedings of the forty-third annual ACM symposium on Theory
  of computing}, pages 685--694. ACM, 2011.

\bibitem{van1992estimating}
B.~Van~Es.
\newblock Estimating functionals related to a density by a class of statistics
  based on spacings.
\newblock {\em Scandinavian Journal of Statistics}, pages 61--72, 1992.

\bibitem{van2005edgeworth}
M.~M. Van~Hulle.
\newblock Edgeworth approximation of multivariate differential entropy.
\newblock {\em Neural computation}, 17(9):1903--1910, 2005.

\bibitem{vasicek1976test}
O.~Vasicek.
\newblock A test for normality based on sample entropy.
\newblock {\em Journal of the Royal Statistical Society. Series B
  (Methodological)}, pages 54--59, 1976.

\bibitem{ver2014discovering}
G.~Ver~Steeg and A.~Galstyan.
\newblock Discovering structure in high-dimensional data through correlation
  explanation.
\newblock In {\em Advances in Neural Information Processing Systems}, pages
  577--585, 2014.

\bibitem{ver2014maximally}
G.~Ver~Steeg and A.~Galstyan.
\newblock Maximally informative hierarchical representations of
  high-dimensional data.
\newblock {\em stat}, 1050:27, 2014.

\bibitem{steeg2015information}
G.~Ver~Steeg and A.~Galstyan.
\newblock The information sieve.
\newblock {\em arXiv preprint arXiv:1507.02284}, 2015.

\bibitem{vincent2003locally}
Pascal Vincent, Yoshua Bengio, et~al.
\newblock Locally weighted full covariance gaussian density estimation.
\newblock Technical report, Technical report 1240, 2003.

\bibitem{wang2005divergence}
Q.~Wang, S.~R. Kulkarni, and S.~Verd{\'u}.
\newblock Divergence estimation of continuous distributions based on
  data-dependent partitions.
\newblock {\em Information Theory, IEEE Transactions on}, 51(9):3064--3074,
  2005.

\bibitem{wang2009divergence}
Q.~Wang, S.~R. Kulkarni, and S.~Verd{\'u}.
\newblock Divergence estimation for multidimensional densities
  via-nearest-neighbor distances.
\newblock {\em Information Theory, IEEE Transactions on}, 55(5):2392--2405,
  2009.

\bibitem{wells1996multi}
W.~M. Wells, P.~Viola, H.~Atsumi, S.~Nakajima, and R.~Kikinis.
\newblock Multi-modal volume registration by maximization of mutual
  information.
\newblock {\em Medical image analysis}, 1(1):35--51, 1996.

\bibitem{wu2014minimax}
Y.~Wu and P.~Yang.
\newblock Minimax rates of entropy estimation on large alphabets via best
  polynomial approximation.
\newblock {\em arXiv preprint arXiv:1407.0381}, 2014.

\bibitem{Yu97}
B.~Yu.
\newblock Assouad, fano, and le cam.
\newblock In {\em Festschrift for Lucien Le Cam}, pages 423--435. Springer,
  1997.

\bibitem{zhu2014bias}
Jie Zhu, Jean-Jacques Bellanger, Huazhong Shu, Chunfeng Yang, and R{\'e}gine
  Le~Bouquin Jeann{\`e}s.
\newblock Bias reduction in the estimation of mutual information.
\newblock {\em Physical Review E}, 90(5):052714, 2014.

\end{thebibliography}
%------------------------------------------

\end{document}